\documentclass[letterpaper]{article} 
\usepackage{aaai2026}  
\usepackage{amsfonts}
\usepackage{algpseudocode}
\usepackage[table]{xcolor}
\usepackage{booktabs}
\usepackage{amsmath}
\usepackage{cleveref}
\usepackage{multirow} 
\usepackage{amsfonts}
\usepackage{times}  
\usepackage{helvet}  
\usepackage{courier}  
\usepackage[hyphens]{url}  
\usepackage{graphicx} 
\usepackage{natbib}  
\usepackage{caption} 
\usepackage{algorithm}

\usepackage{newfloat}
\usepackage{listings}
\DeclareCaptionStyle{ruled}{labelfont=normalfont,labelsep=colon,strut=off} 
\floatstyle{ruled}
\newfloat{listing}{tb}{lst}{}
\floatname{listing}{Listing}
\title{DomainCQA: Crafting Knowledge-Intensive QA from Domain-Specific Charts}
\author{
    Yujing Lu\textsuperscript{\rm 1}\thanks{Equal contribution.},
    Ling Zhong\textsuperscript{\rm 1}\footnotemark[1],
    Jing Yang\textsuperscript{\rm 1}\footnotemark[1],
    Weiming Li\textsuperscript{\rm 1}\footnotemark[1],
    Peng Wei\textsuperscript{\rm 2},
    Yongheng Wang\textsuperscript{\rm 1},\\
    Manni Duan\textsuperscript{\rm 1}\thanks{Corresponding author.},
    Qing Zhang\textsuperscript{\rm 1}\footnotemark[2]
}
\affiliations{
    \textsuperscript{\rm 1}Zhejiang Lab, Hangzhou, China\\
    \textsuperscript{\rm 2}National Astronomical Observatories, Chinese Academy of Sciences, Beijing, China\\
    \{luyujing, zhongling, yangjing0128, liwm, wangyh, duanmanni, qing.zhang\}@zhejianglab.org,\\ weipeng01@nao.cas.cn
}

\usepackage{bibentry}
\nocopyright

\begin{document}

\maketitle

\begin{abstract}
Chart Question Answering (CQA) evaluates Multimodal Large Language Models (MLLMs) on visual understanding and reasoning over chart data. However, existing benchmarks mostly test surface-level parsing, such as reading labels and legends, while overlooking deeper scientific reasoning. We propose DomainCQA, a framework for constructing domain-specific CQA benchmarks that emphasize both visual comprehension and knowledge-intensive reasoning. It integrates complexity-aware chart selection, multitier QA generation, and expert validation. Applied to astronomy, DomainCQA yields AstroChart, a benchmark of 1,690 QA pairs over 482 charts, exposing persistent weaknesses in fine-grained perception, numerical reasoning, and domain knowledge integration across 21 MLLMs. Fine-tuning on AstroChart improves performance across fundamental and advanced tasks. Pilot QA sets in biochemistry, economics, medicine, and social science further demonstrate DomainCQA’s generality. Together, our results establish DomainCQA as a unified pipeline for constructing and augmenting domain-specific chart reasoning benchmarks.

\end{abstract}

\begin{links}
    \link{Code}{https://github.com/LingZhong01/DomainCQA}
    \link{Datasets}{https://huggingface.co/datasets/yangjing0128/AstroChart}
    \link{Extended version}{https://arxiv.org/abs/2503.19498}
\end{links}

\section{Introduction}

The success of Multimodal Large Language Models (MLLMs) has sparked growing interest in their ability to process and analyze scientific charts, which play a crucial role in conveying complex research data \cite{rekateam2024reka,openai2024gpt4}. Among various chart-related tasks, Chart Question Answering (CQA) has emerged as a fundamental challenge, requiring MLLMs to extract, interpret, and reason about chart-based information in response to natural language queries.

\begin{figure}[t]
    \centering
    \includegraphics[width=1\linewidth]{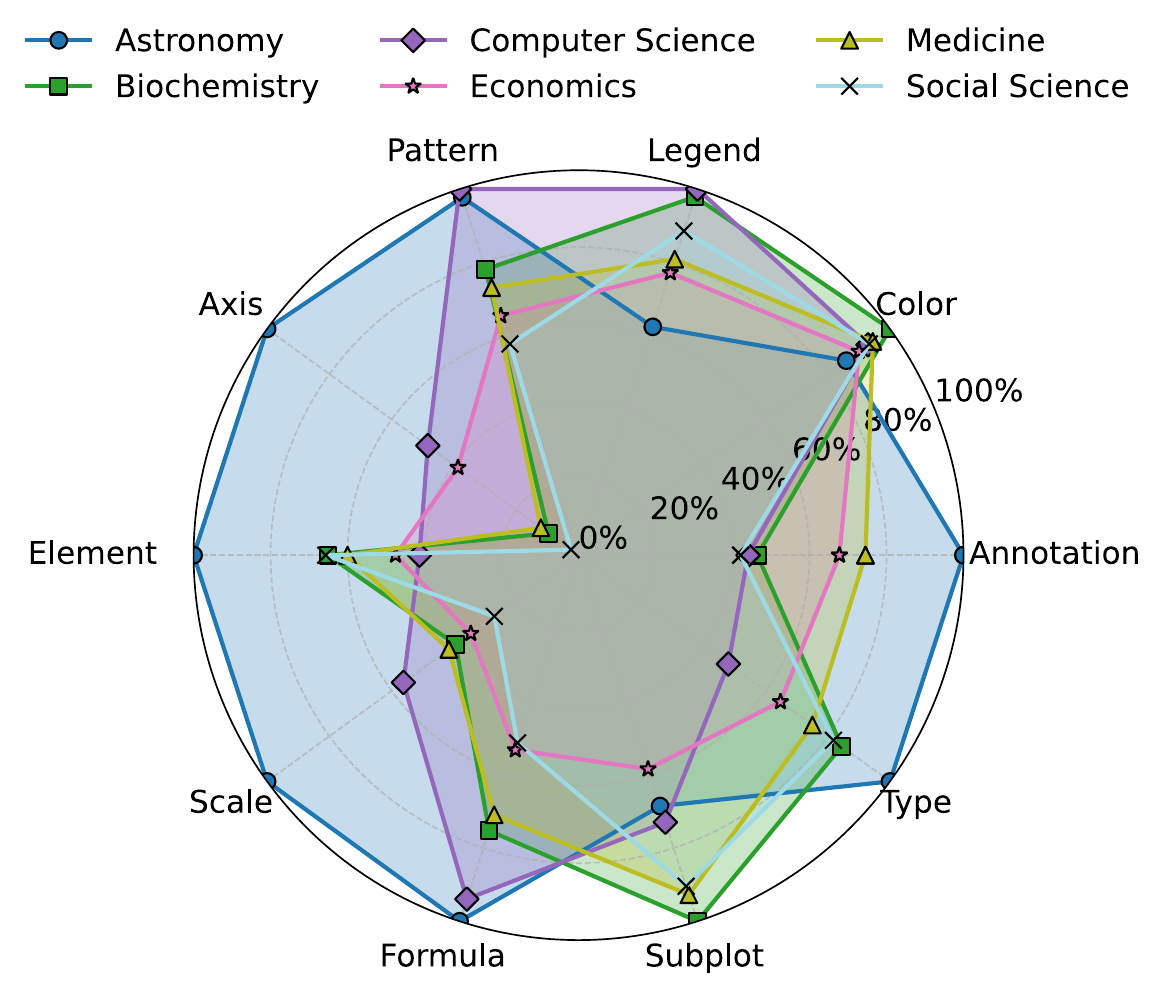}
    \caption{Radar plot of chart complexity across domains by comparing various visual design features, computed from 500 sampled charts per domain. Each axis represents a normalized design element contributing to overall chart complexity (formally defined later as the Chart Complexity Vector, or CCV). The domain-specific differences motivate our complexity-aware chart selection strategy.}
    \label{fig:ccv_domain_comparison}
\end{figure}

Although recent benchmarks in CQA, such as ChartQA~\cite{chartqa}, PlotQA~\cite{plotqa}, CharXiv~\cite{charxiv}, and SciCap~\cite{scicap}, have greatly advanced the field, all of them are deliberately \emph{knowledge-agnostic}. Their question-answer (QA) pairs probe a model’s ability to parse axes, legends and visual layouts, yet never require \emph{domain--specific reasoning}. Consequently, we still do not know whether modern MLLMs can truly integrate visual cues \textit{and} scientific knowledge.

Simply extending existing benchmark--building pipelines is inadequate for two reasons: (1) \textbf{Chart selection:} current pipelines choose charts either randomly or by ad--hoc manual curation, overlooking the fact that the mix of visual elements differs sharply from one scientific field to another, as Figure \ref{fig:ccv_domain_comparison} shows that astronomy charts emphasize annotation and formula usage, biochemistry charts lean on color and subplot, etc. In short, the charts selected in these benchmarks are \emph{not} \textit{domain‐representative}; (2) \textbf{Question design:} existing CQA datasets still focus on superficial visual cues; they rarely ask questions that demand domain knowledge, for instance, in astronomy charts correlating oscillation-frequency histograms with stellar classifications or interpreting how a redshift-magnitude scatter plot reflects cosmic expansion. In short, the questions designed in these benchmarks are \emph{not} \textit{knowledge‐intensive}.

To address the two gaps identified above, we present \textbf{DomainCQA}, a framework for building domain-specific CQA benchmarks that integrates chart selection, QA generation, and expert QA validation into one seamless process. We encode each candidate chart with a 10-dimensional \textit{Chart Complexity Vector (CCV)} and apply non-parametric Gibbs sampling to select the subset of charts used for questions that test basic understanding. For domain knowledge probing, we propose a chart abstract selector using chain-of-thought (CoT) reasoning to identify the most representative chart, along with a voting validator that enhances robustness through cross-model majority voting. Across both chart pools, we construct two tiers of QA: \textit{Fundamental QA (FQA)} and \textit{Advanced QA (AQA)}, and pass every QA pair through a multi-stage human review. The resulting benchmark is both domain-representative in its visuals and genuinely knowledge-intensive in its questions.

As a concrete application of DomainCQA, we construct \textbf{AstroChart}, the first CQA benchmark for astronomy. Leveraging our pipeline, we select $482$ representative charts and generate $1,690$ QA pairs. Of these, $1,509$ are FQA pairs that test the understanding of the chart itself, while $181$ are AQA pairs that require extra astronomical knowledge beyond the chart. Evaluating 21 state-of-the-art MLLMs on AstroChart exposes three persistent weaknesses: (i) chart reasoning – inferring trends and relationships from visual encodings; (ii) numerical computation – extracting values and performing arithmetic reliably; and (iii) domain-fact integration – combining chart evidence with astronomy-specific knowledge. Fine-tuning these models on data generated by DomainCQA yields notable gains, confirming the framework’s value for both evaluation and data creation.

Beyond astronomy, we create pilot sets in biochemistry, economics, medicine and social science, each with domain specific charts and QA pairs showing that DomainCQA generalizes well across disciplines. These results confirm that the framework effectively addresses the two key gaps: selecting representative charts and generating knowledge-intensive questions.

Our key contributions are as follows:
(1) DomainCQA, a three-phase framework for building domain-specific CQA benchmarks;
(2) CCV, a $10$-dimensional descriptor that captures domain-dependent visual traits and guides chart selection;
(3) Chart abstracts, defined as charts summarizing articles' main findings, are ideal anchors for knowledge-intensive question generation;
(4) AstroChart, the first CQA benchmark for astronomy; we evaluate 21 state-of-the-art (SOTA) MLLMs in zero-shot and fine-tuned settings to probe their domain-specific chart understanding.

\section{Related Work}

\paragraph{\textbf{MLLMs for Chart Understanding}}
Recent progress in MLLMs has substantially advanced chart understanding. Proprietary models such as GPT-4o~\cite{openai2024gpt4}, Claude 3.5~\cite{claude3}, Qwen-VL~\cite{bai2023qwenvl}, and Gemini-2.5 \cite{comanici2025gemini2_5_flash} have demonstrated strong multimodal reasoning capabilities. Meanwhile, open-source MLLMs are rapidly evolving, offering accessible and customizable alternatives. Many models primarily focus on enhancing general vision-language ability through improved alignment, stronger representations, and more efficient inference, which improves performance on chart-related tasks. Notable examples include LLaVA~\cite{liu2023llava,liu2023improvedllava,liu2024llavanext}, mPLUG-Owl~\cite{ye2023mplugowl,ye2023mplugowl2,ye2024mplugowl3longimagesequenceunderstanding}, SPHINX~\cite{liu2024sphinxxscalingdataparameters}, InternVL~\cite{chen2023internvl}, CogVLM~\cite{hong2024cogvlm2}, MiniCPM~\cite{yao2024minicpm}, and Pixtral~\cite{agrawal2024pixtral12b}. In contrast, other models are specifically fine-tuned on chart-related tasks to better support structured data understanding, such as UniChart~\cite{masry-etal-2023-unichart},  Matcha~\cite{liu-etal-2023-matcha}, ChartAssistant~\cite{meng-etal-2024-chartassistant}, and TinyChart~\cite{zhang2024tinychartefficientchartunderstanding}.

\paragraph{\textbf{Benchmarks for CQA Evaluation}}
A CQA benchmark consists of two key components: charts and corresponding QA pairs, both essential for evaluating a model’s chart comprehension capabilities \cite{huang-etal-2024-chart-survey}. Early datasets like DVQA~\cite{dvqa} and FigureQA~\cite{figureqa} utilized synthetic charts alongside templated QA pairs, whereas later efforts such as PlotQA~\cite{plotqa}, LEAF-QA~\cite{leafqa}, and LEAF-QA++\cite{leafqa++} incorporated real numerical data with synthetic visualizations. More recent benchmarks, such as ChartQA\cite{chartqa}, OpenCQA~\cite{opencqa}, and MMC-Benchmark~\cite{mmcbench}, introduced charts sourced from real-world datasets. Among these, OpenCQA pioneered open-ended CQA tasks. The growing capabilities of LLMs have enabled recent studies such as SciGraphQA\cite{scigraphqa}, ChartX\cite{chartx}, and CharXiv\cite{charxiv} to generate more diverse QA pairs. Nevertheless, existing benchmarks mainly focus on general or broad scientific domains and lack the domain-specific focus required for detailed chart interpretation.

\section{DomainCQA Framework}

\begin{figure*}[t]
    \centering
    \includegraphics[width=1\linewidth]{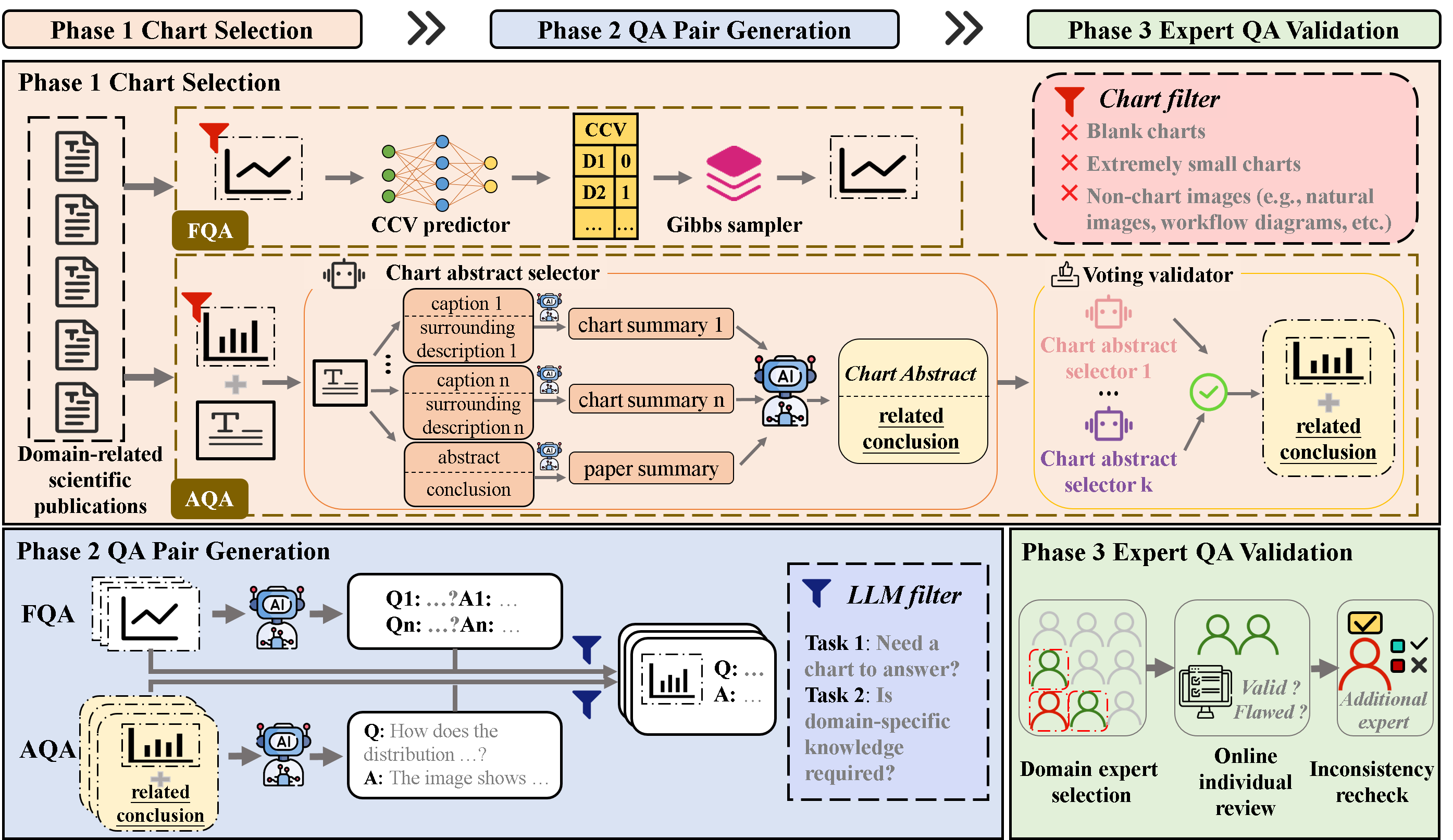}
    \caption{Overview of the DomainCQA framework for constructing domain-specific CQA benchmarks. The pipeline consists of three stages: Chart Selection, QA Pair Generation and Expert QA Validation. The resulting benchmarks support evaluation of both visual comprehension and knowledge-intensive reasoning.}
    \label{fig:flowchart}
\end{figure*} 

DomainCQA (see Figure \ref{fig:flowchart}) offers a systematic framework to build domain-specific CQA benchmarks that test both general visual understanding and specialized reasoning. It defines two types of QA tasks:
\begin{itemize}
    \item \textbf{Fundamental QA (FQA)}: testing chart comprehension via basic visual reasoning like label recognition, color differentiation, and simple comparisons.
    \item \textbf{Advanced QA (AQA)}: requiring domain knowledge beyond the chart, including interpreting specialized symbols, terms, or concepts.
\end{itemize}
Together, questions from both tasks enable the benchmark to evaluate chart understanding across a spectrum from surface-level comprehension to discipline-specific insight.

\subsection{Chart Selection}
\label{sec:chartSelection}
DomainCQA selects the charts separately for the FQA pairs and the AQA pairs, ensuring that each set is aligned with the specific evaluation requirements of its type of question. 

\paragraph{Charts for FQA} Our goal is to build an FQA-chart pool whose visual variety matches the unknown, true distribution of charts in a scientific domain.  We operate on a pre-compiled corpus of domain charts and focus on two ingredients: a Chart Complexity Vector (CCV) that embeds each chart in a ten-dimensional feature space, and a non-parametric Gibbs sampler that draws a subset whose joint CCV statistics closely match those of the corpus. 

Each CCV dimension measures a distinct aspect of visual difficulty, such as plot elements, color diversity, annotation density, and visual clutter. We train a ResNet-50 classifier on an annotated subset to predict the ten CCV attributes for the remaining charts, yielding 10-dimensional representations that capture domain-specific patterns (see Appendix A for more details on CCV).

Random sampling disregards the structured distribution of visual complexity within each domain, producing samples that do not faithfully reflect domain-specific patterns. Instead, we treat the CCV collection as an empirical distribution and perform non-parametric Gibbs sampling\cite{Casella1992ExplainingTG} to preserve marginal distributions and inter-dimensional dependencies (see Appendix B for the Gibbs sampling pseudocode).

Let $\mathcal{C} = \{\mathbf{c}^{(1)}, \dots, \mathbf{c}^{(N)}\} \subset \mathbb{R}^{10}$ be the set of CCV vectors for all candidate charts. Each $\mathbf{c}^{(n)} = (c^{(n)}_{1}, \dots, c^{(n)}_{10})$ encodes the visual, structural, and interpretive attributes of a chart. At each iteration t, we:
\begin{enumerate}
    \item Randomly choose a dimension $k_t \in \{1,\dots,10\}$;
	\item Sample a target value $\zeta \sim \hat{p}_{k_t}$, the empirical marginal distribution of dimension $k_t$;
	\item Search for a new chart $\mathbf{c}^{(t)} \in \mathcal{C}$ that best matches the current state $\mathbf{c}^{(t-1)}$ on the remaining $9$ dimensions and is closest to $\zeta$ in dimension $k_t$.
\end{enumerate}
The resulting chart subset approximates the latent domain distribution in CCV space and serves as our FQA chart pool.

\paragraph{Charts for AQA} Selecting charts for AQA requires more than visual diversity, instead it demands charts that meaningfully reflect domain knowledge. A naive approach would be to reuse charts from the FQA set and pose domain-specific questions on them. However, this often results in noisy inputs that dilute question quality. Many visually complex charts are tangentially related to the paper’s core findings, making them poor candidates for knowledge-intensive tasks.

We address this by targeting a chart that directly reflects a paper’s main scientific conclusions, commonly referred to as \textit{chart abstract}. To identify them, we design a lightweight two-stage LLM-based method: a \textit{chart abstract} selector that leverages CoT to identify the chart most relevant to a paper’s abstract and conclusion, and a voting validator, which aggregates reasoning outputs from multiple LLMs via cross-model majority voting to enhance selection reliability (see Appendix C for the pseudocode of AQA chart selection).

This approach yields semantically meaningful charts that support deeper reasoning, as demonstrated by our later experiments across five scientific domains (see Sec.~5.5). QA pairs constructed from chart abstracts consistently outperform those from our FQA method in both domain relevance and QA validity.

\subsection{QA Pair Generation}
From selected charts, we design two types of questions: FQA and AQA. FQA covers four categories of tasks: 
\textit{Visual} (recognizing graphical elements), 
\textit{Data} (retrieving and computing values), 
\textit{Inference} (inferring patterns and relations), and 
\textit{Chart Description} (summarizing the visual content). 
AQA is formulated as a knowledge-based inference (\textit{KB-Inference}) task, requiring integration of external scientific knowledge with visual content.

To ensure quality, we apply a secondary LLM-based validation filter to all generated QA pairs. This verifier checks two key criteria:
(1) whether the QA pair is grounded in the visual content of the chart, and
(2) for AQA, whether it requires domain-specific knowledge to answer (see Appendices D and E for prompt templates and validation criteria.)

\subsection{Expert QA Validation}
To ensure benchmark quality, each QA pair undergoes expert review to validate both its clarity and factual correctness. Reviewers label each item as either: \textit{Valid} (the question is well-posed and the answer is accurate); \textit{Flawed} (the question is ambiguous, misleading, or the answer is incorrect).
All QA pairs are independently assessed by domain experts. Disagreements are resolved through additional review rounds until consensus is reached.

\section{AstroChart: A Benchmark for Astronomy}
We present a complete benchmark instantiation, AstroChart, in the astronomy domain, comprising $1,690$ QA pairs grounded in $482$ charts. We also conducted partial experiments in other domains to validate key steps, chart selection and QA pair generation for AQA, as detailed in Evaluation.

\label{astrochart}

\paragraph{Chart Selection}
To construct the FQA chart portion of AstroChart, we collected figures from arXiv astronomy papers published between 2007 and 2023. A ResNet-18 classifier, trained to detect non-scientific or low-quality visuals, was used to filter out irrelevant figures. For each remaining chart, we computed its CCV and applied non-parametric Gibbs sampling to select $305$ charts whose CCV distribution approximates the overall domain distribution, ensuring a diverse and representative subset.

\begin{figure}[t]
    \centering
    \includegraphics[width=1\columnwidth]{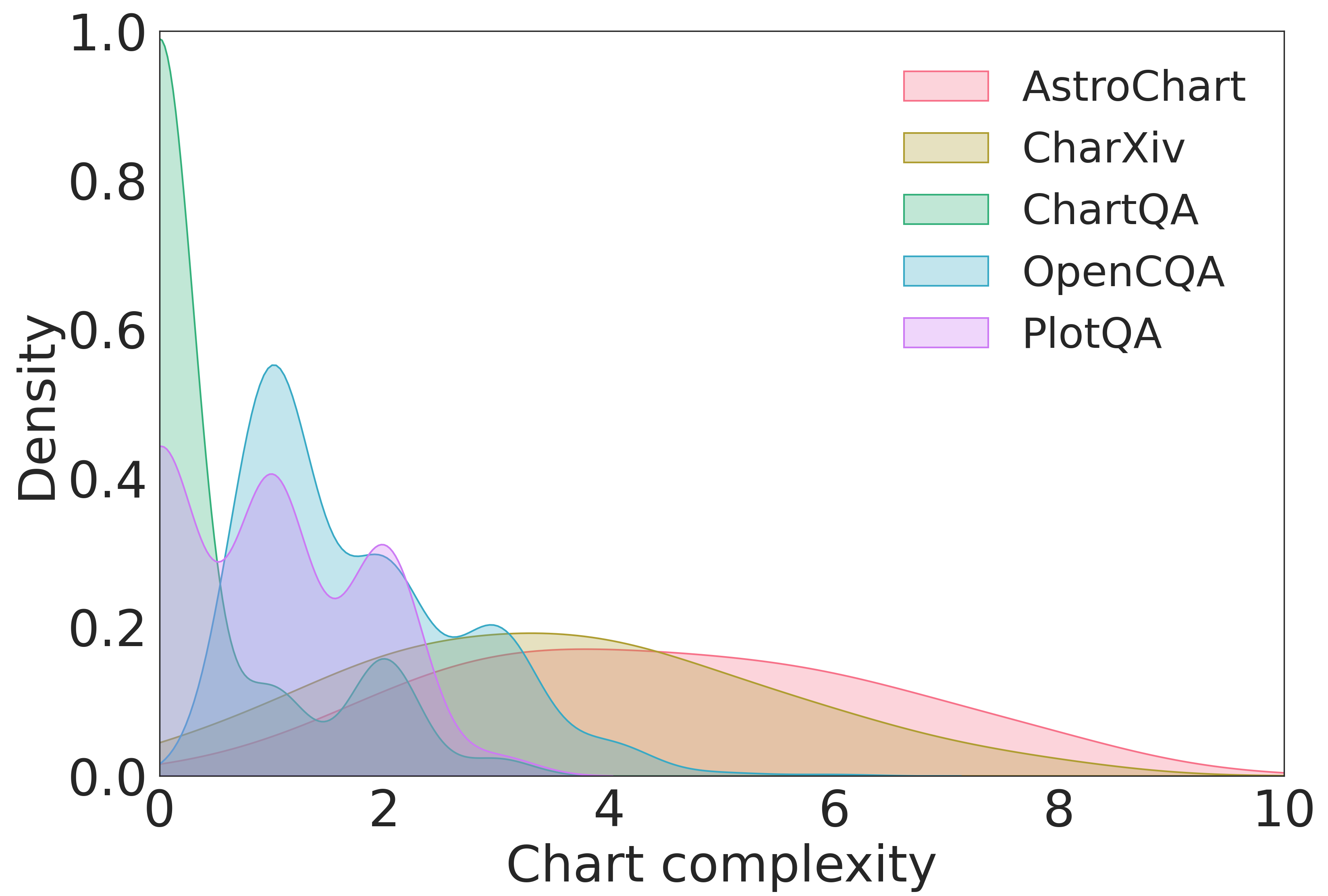} 
    \caption{Chart complexity calculated from CCVs across benchmarks, where AstroChart shows a broader and higher complexity distribution than other benchmarks, with more domain-specific charts in the 6–10 range (see Appendix A.2 for CCV score details).}
    \label{fig:ccsDist}
\end{figure}

To assess the representativeness of our selected charts, we further compared the visual complexity of AstroChart with existing CQA benchmarks. Specifically, we computed the CCV for each chart in several public datasets (CharXiv, ChartQA, OpenCQA, PlotQA), and summed the ten CCV dimensions to obtain an overall complexity score. As illustrated in Figure~\ref{fig:ccsDist}, the charts in these benchmarks are mainly clustered in the 1 to 4 range, indicating simple visual structures. In contrast, AstroChart centers around 4 to 6 and exhibits a broader spread across the complexity spectrum. This suggests that AstroChart offers richer and more varied visual content, better aligned with real-world scientific charts.

For AQA, we targeted the high-impact literature by selecting the top 1\% most-cited articles each year in the six main subfields of astronomy (See Appendix F). After applying the same filtering process, we identify \textit{chart abstracts} using a consensus-based approach from GPT-4o and Claude 3.5. This results in $178$ high-quality charts suitable for domain-specific reasoning.

In total, AstroChart includes 482 distinct charts with one in both (see Appendix G for visualizations).

\paragraph{QA Pair Generation} We employ Claude 3.5 to generate QA pairs using category-specific prompts, ensuring that each question is well aligned with its associated chart. To refine quality, GPT-4o is used to automatically filter out QA pairs that either lack a clear connection to the chart or do not require external domain knowledge for answering.

We further assessed the reliability of GPT-4o’s filtering by comparing its judgments against human annotations on $200$ randomly sampled QA pairs. Beyond achieving 96.5\% overall accuracy, GPT-4o demonstrated substantial agreement with human reviewers, with a Cohen’s Kappa \cite{cohen1960coefficient} of $0.77$, indicating strong consistency in identifying deletable items. Most discrepancies were conservative false positives, underscoring GPT-4o’s cautious filtering style and practical reliability at scale (see Appendix H for details).

The final dataset comprises $1,690$ QA pairs, including $1,509$ FQA pairs and $181$ AQA pairs, as summarized in Table~\ref{tab:astronomical_benchmark} (see Appendix I for examples).

\begin{table}[t]
  \centering
  \small
  \renewcommand{\arraystretch}{1.1}
  \begin{tabular}{lllr}
    \hline
    \textbf{Type} & \textbf{Category} & \textbf{Aspect} & \textbf{Count} \\
    \hline
    \multirow{4}{*}{FQA} 
    & \multirow{4}{*}{Visual} & Color       & 211 \\
                              &             & Style       & 133 \\
                              &             & Text        & 213 \\
                              &             & Layout      & 45  \\ \cline{2-4}
    & \multirow{3}{*}{Data}   & Point       & 130 \\
                              &             & Interval    & 102 \\
                              &             & Calculation & 84  \\ \cline{2-4}
    & Inference               &             & 289 \\
    & Chart Description                 &             & 302 \\
    \hline
    \multirow{1}{*}{AQA} 
    & KB-Inference            &             & 181 \\
    \hline
    \multicolumn{3}{l}{\textbf{Total}}      & \textbf{1690} \\
    \hline
  \end{tabular}
  \caption{Distribution of question types in AstroChart}
  \label{tab:astronomical_benchmark}
\end{table}

\paragraph{Expert QA Validation} We conducted a comprehensive verification of the entire AstroChart benchmark to ensure its accuracy and reliability. A team of eight astronomy experts independently reviewed all $1,690$ QA pairs using our custom online assessment platform, with a total annotation time exceeding $160$ hours. Each pair of QAs was evaluated by two randomly assigned reviewers and any disagreements were resolved through additional review rounds until consensus was reached (details in Appendix J). This rigorous expert validation process reinforces the credibility of AstroChart as a high-quality benchmark for evaluating MLLMs in astronomical chart understanding.

\section{Evaluation}
To assess the utility and difficulty of AstroChart, we design three experiments. First, we benchmark 21 SOTA MLLMs under a zero-shot setting to assess their capabilities across question categories. Second, we construct a training set using the same pipeline as AstroChart (excluding expert validation), fine-tune a representative model, and test its performance on both AstroChart and other benchmarks to assess generalization. Third, we compare AstroChart with CharXiv to evaluate relative difficulty.  Finally, we also verify that the DomainCQA framework can produce high-quality AQA pairs in other scientific domains.
\label{sec:evaluations}
\subsection{Setup and Metrics}
\paragraph{Zero-Shot Setup}
We evaluated 21 MLLMs, including both proprietary and open-source variants. Proprietary models were accessed via API, and open-source models were run locally on a single Nvidia A100-80GB GPU. Under the zero-shot protocol, each model received only the chart and its corresponding question, without any in-context examples or prior training. Four astronomy researchers were also invited to establish a human baseline by answering 10\% of questions from each category using the same prompts as the models to ensure fairness.

\paragraph{Fine-Tuning Setup}
To evaluate training effectiveness, we constructed a fine-tuning dataset using the same pipeline as AstroChart, omitting the final expert QA validation step. This yielded 9,857 training and 8,729 validation scientific charts, from which we generated 86,681 and 21,738 QA pairs, respectively. We fine-tuned an open-source model, MiniCPM-V2.6-8B, using the parameter-efficient LoRA \cite{hu2021loralowrankadaptationlarge} method. Training was conducsted on 8 Nvidia A100-80GB GPUs with BF16 mixed precision and DeepSpeed ZeRO-2 \cite{rajbhandari2021zero} optimization for scalability and efficiency.

\paragraph{Evaluation Metrics} We assess the accuracy of model outputs for both numerical and open-ended questions (details in Appendix K). For numerical responses, we computed relative error normalized by the axis range for retrieval tasks, and required an exact match for derivation tasks such as counting or arithmetic. For open-ended responses, an LLM judge (DeepSeek-V3) assigned scores from $0$ to $1$ based on relevance, correctness, and completeness, following \citet{liu-etal-2023-g}. To verify scoring reliability, we compared DeepSeek-V3’s scores with human annotations on $176$ samples, yielding a Pearson correlation of $0.816$, Spearman correlation of $0.817$, and MAE of $0.096$. 
ROUGE-L \cite{lin-2004-rouge}, BLEU-4 \cite{papineni-etal-2002-bleu}, and L3Score \cite{pramanick2024spiqa} show similar trends to LLM scoring (see Appendix L).

\subsection{Benchmarking 21 MLLMs on AstroChart}
We report the performance of 21 MLLMs on AstroChart across FQA and AQA categories, as shown in \cref{tab:model_evaluation_results}.
\begin{table*}[ht]
\centering
\setlength{\tabcolsep}{1.5pt} 
\scriptsize  
\renewcommand{\arraystretch}{1.25}
\begin{tabular}{lccccccccccccc}
\hline
\multicolumn{1}{c}{} 
& \multicolumn{11}{c}{\textbf{FQA}} 
& \multicolumn{1}{c}{\textbf{AQA}} 
& \multicolumn{1}{c}{} \\
\cmidrule(lr){2-12} \cmidrule(lr){13-13}
\multicolumn{1}{c}{} & \multicolumn{5}{c}{\textbf{Visual/602}}         & \multicolumn{4}{c}{\textbf{Data/316}} & \multicolumn{1}{c}{} & \multicolumn{1}{c}{} & \multicolumn{1}{c}{}                                        & \multicolumn{1}{c}{}                                   \\ \cmidrule(lr){2-6} \cmidrule(lr){7-10}
\multirow{-3}{*}{\textbf{Model}} & \multicolumn{1}{c}{\textbf{All}}     & \multicolumn{1}{c}{\textbf{Color}}   & \multicolumn{1}{c}{\textbf{Style}}   & \multicolumn{1}{c}{\textbf{Text}}    & \multicolumn{1}{c}{\textbf{Layout}}  & \multicolumn{1}{c}{\textbf{All}}     & \multicolumn{1}{c}{\textbf{Point}}   & \multicolumn{1}{c}{\textbf{Interval}} & \multicolumn{1}{c}{\textbf{Calculation}} & \multicolumn{1}{c}{\multirow{-2}{*}{\textbf{Infer./289}}} & \multicolumn{1}{c}{\multirow{-2}{*}{\textbf{Chart Desc./302}}} & \multicolumn{1}{c}{\multirow{-2}{*}{\textbf{KB-Infer./181}}} & \multicolumn{1}{c}{\multirow{-3}{*}{\textbf{All/1690}}} \\ \hline

\textbf{Human Baseline(10\% Sample)} & \cellcolor[HTML]{E7E6E6}{98.60} & 98.54 & 98.40 & 98.63 & 99.53 & \cellcolor[HTML]{E7E6E6}{96.40} & 98.62 & 93.86 & 96.50 & 91.82 & 70.00 & 39.00 & \cellcolor[HTML]{E7E6E6}{85.56} \\
\hline
\multicolumn{14}{c}{\textbf{Proprietary Multimodal Large Language Models}} \\
\hline\textbf{Gemini-2.5-Pro\cite{comanici2025gemini2_5_flash}} & \cellcolor[HTML]{E7E6E6}\textbf{88.22} & 87.37 & \textbf{87.70} & \textbf{90.23} & \textbf{84.67} & \cellcolor[HTML]{E7E6E6}\textbf{72.66} & \textbf{81.22} & \textbf{75.10} & 56.43 & 81.31 & 81.09 & \textbf{73.65} & \cellcolor[HTML]{E7E6E6}\textbf{81.30} \\
\textbf{Gemini-2.5-flash\cite{comanici2025gemini2_5_flash}} & \cellcolor[HTML]{E7E6E6}87.21 & 87.04 & 87.04 & 88.73 & 81.78 & \cellcolor[HTML]{E7E6E6}64.15 & 68.34 & 63.40 & \textbf{58.57} & \textbf{82.01} & \textbf{82.65} & 72.49 & \cellcolor[HTML]{E7E6E6}79.62 \\
\textbf{GPT-4o\cite{openai2024gpt4}} & \cellcolor[HTML]{E7E6E6}86.23 & \textbf{88.92} & 84.15 & 85.31 & \textbf{84.67} & \cellcolor[HTML]{E7E6E6}53.19 & 53.78 & 60.35 & 43.57 & 75.40 & 80.96 & 73.04 & \cellcolor[HTML]{E7E6E6}75.84 \\
\textbf{Qwen-VL-Max\cite{bai2023qwenvl}} & \cellcolor[HTML]{E7E6E6}83.13 & 87.79 & 76.78 & 83.43 & 77.78 & \cellcolor[HTML]{E7E6E6}50.96 & 52.27 & 56.55 & 42.14 & 75.16 & 76.62 & 68.23 & \cellcolor[HTML]{E7E6E6}72.99 \\
\hline
\multicolumn{14}{c}{\textbf{Open-source Multimodal Large Language Models}} \\
\hline   

\textbf{TinyChart-3B\cite{zhang2024tinychartefficientchartunderstanding}} 
    & \cellcolor[HTML]{E7E6E6}29.41 & 47.75 & 25.15 & 13.71 & 27.11 
    & \cellcolor[HTML]{E7E6E6}12.22 & 18.80 & 9.39  & 5.48  
    & 23.94 & 1.56 & 20.83 & \cellcolor[HTML]{E7E6E6}19.36 \\

\textbf{Llava1.5-7B\cite{liu2023improvedllava}} 
    & \cellcolor[HTML]{E7E6E6}31.04 & 49.39 & 27.70 & 14.34 & 33.33 
    & \cellcolor[HTML]{E7E6E6}8.47  & 8.88  & 7.98  & 8.45  
    & 42.53 & 13.94 & 45.36 & \cellcolor[HTML]{E7E6E6}27.26 \\

\textbf{Llava1.6-Mistral-7B\cite{liu2024llavanext}} 
    & \cellcolor[HTML]{E7E6E6}46.45 & 61.36 & 41.96 & 33.00 & 50.89 
    & \cellcolor[HTML]{E7E6E6}13.77 & 17.74 & 14.18 & 7.14  
    & 49.24 & 23.84 & 48.23 & \cellcolor[HTML]{E7E6E6}36.97 \\

\textbf{Qwen-VL-Chat-7B\cite{bai2023qwenvl}} 
    & \cellcolor[HTML]{E7E6E6}44.47 & 55.73 & 41.04 & 32.68 & 54.44 
    & \cellcolor[HTML]{E7E6E6}10.47 & 16.11 & 6.72  & 6.31  
    & 38.89 & 22.05 & 45.19 & \cellcolor[HTML]{E7E6E6}33.23 \\

\textbf{Janus-Pro-7B\cite{lu2024deepseekvl}}
    & \cellcolor[HTML]{E7E6E6}66.69 & 74.74 & 67.26 & 56.62 & 74.67 
    & \cellcolor[HTML]{E7E6E6}32.27 & 35.10 & 38.09 & 20.83 
    & 56.37 & 51.23 & 54.75 & \cellcolor[HTML]{E7E6E6}54.45 \\

\textbf{MiniCPM-V2.6-8B\cite{yao2024minicpm}} &  \cellcolor[HTML]{E7E6E6}70.31 & 75.92 & 61.89 & 71.92 & 61.78 & \cellcolor[HTML]{E7E6E6}33.30 & 34.87 & 43.74 & 18.21 & 55.16 & 55.60 & 54.20 & \cellcolor[HTML]{E7E6E6}56.44 \\

\textbf{InternVL3‑8B\cite{zhu2025internvl3}} 
    & \cellcolor[HTML]{E7E6E6}66.64 & 72.72 & 62.04 & 62.23 & 71.78 
    & \cellcolor[HTML]{E7E6E6}35.41 & 38.89 & 42.87 & 20.95 
    & 54.33 & 49.57 & 54.09 & \cellcolor[HTML]{E7E6E6}54.30 \\

\textbf{mPLUG-Owl2-8.2B\cite{ye2023mplugowl2}} 
    & \cellcolor[HTML]{E7E6E6}28.54 & 39.95 & 27.70 & 16.48 & 32.00 
    & \cellcolor[HTML]{E7E6E6}9.48  & 11.15 & 12.20 & 3.57  
    & 38.41 & 9.37  & 42.21 & \cellcolor[HTML]{E7E6E6}24.70 \\

\textbf{Pixtral-12B\cite{agrawal2024pixtral12b}} & \cellcolor[HTML]{E7E6E6}79.27 & 83.00 & 75.70 & 78.26 & 76.89 & \cellcolor[HTML]{E7E6E6} 51.54 & 53.63 & 60.64 & 37.26 & 71.90 & 78.74 & 69.28 & \cellcolor[HTML]{E7E6E6}71.66 \\

\textbf{Llava1.6-Vicuna-13B\cite{liu2024llavanext}} 
    & \cellcolor[HTML]{E7E6E6}49.45 & 66.43 & 44.74 & 34.84 & 50.89 
    & \cellcolor[HTML]{E7E6E6}13.23 & 17.77 & 10.40 & 9.64  
    & 44.36 & 23.44 & 50.77 & \cellcolor[HTML]{E7E6E6}37.30 \\

\textbf{SPHINX-v2-13B\cite{liu2024sphinxxscalingdataparameters}} 
    & \cellcolor[HTML]{E7E6E6}31.68 & 48.40 & 29.41 & 18.26 & 21.11 
    & \cellcolor[HTML]{E7E6E6}7.23  & 13.36 & 1.47  & 4.76  
    & 37.27 & 6.13  & 44.25 & \cellcolor[HTML]{E7E6E6}24.84 \\

\textbf{Llama4-Maverick-17B\cite{meta2025llama4}} 
    & \cellcolor[HTML]{E7E6E6}84.27 & 86.01 & 78.59 & 86.20 & \textbf{83.56} 
    & \cellcolor[HTML]{E7E6E6}56.30 & 55.14 & 58.27 & \textbf{55.71} 
    & 77.02 & 76.42 & \textbf{74.64} & \cellcolor[HTML]{E7E6E6}75.37 \\

\textbf{CogVLM2-19B\cite{hong2024cogvlm2}} 
    & \cellcolor[HTML]{E7E6E6}66.29 & 74.81 & 54.52 & 64.04 & 71.78 
    & \cellcolor[HTML]{E7E6E6}29.27 & 29.82 & 37.48 & 18.45 
    & 51.90 & 54.90 & 50.66 & \cellcolor[HTML]{E7E6E6}53.20 \\

\textbf{Gemma-3-27B\cite{team2025gemma}} 
    & \cellcolor[HTML]{E7E6E6}69.93 & 69.30 & 68.44 & 69.06 & 80.89 
    & \cellcolor[HTML]{E7E6E6}37.21 & 38.22 & 47.63 & 22.98 
    & 58.72 & 66.23 & 62.54 & \cellcolor[HTML]{E7E6E6}60.44 \\

\textbf{Llava1.6-Yi-34B\cite{liu2024llavanext}} 
    & \cellcolor[HTML]{E7E6E6}50.63 & 66.34 & 44.37 & 37.93 & 53.56 
    & \cellcolor[HTML]{E7E6E6}18.19 & 17.60 & 25.30 & 10.48 
    & 47.09 & 36.19 & 55.36 & \cellcolor[HTML]{E7E6E6}41.89 \\

\textbf{Qwen2.5-VL-72B\cite{bai2025qwen2}} 
    & \cellcolor[HTML]{E7E6E6}83.21 & 85.31 & 77.04 & 86.34 & 76.22 
    & \cellcolor[HTML]{E7E6E6}53.46 & 54.57 & 56.36 & 48.21 
    & 72.46 & 77.52 & 68.34 & \cellcolor[HTML]{E7E6E6}73.20 \\

\textbf{Pixtral-large-124B\cite{mistral2024pixtral_large}} 
    & \cellcolor[HTML]{E7E6E6}\textbf{86.11} & \textbf{86.76} & \textbf{82.59} & \textbf{88.22} & 82.44 
    & \cellcolor[HTML]{E7E6E6}\textbf{59.38} & \textbf{63.67} & \textbf{63.51} & 47.74 
    & \textbf{78.65} & \textbf{80.93} &  70.83 & \cellcolor[HTML]{E7E6E6}\textbf{77.23} \\ 
\hline
\multicolumn{14}{c}{\textbf{Fine-tuned}} \\
\hline

\textbf{MiniCPM-V2.6-8B-fine-tuned} & \cellcolor[HTML]{E7E6E6}78.15↑ & 81.08\textbf{↑} & 76.26↑ & 76.76↑ & 76.00↑ & \cellcolor[HTML]{E7E6E6}37.47↑  & 37.66↑ & 47.78↑ & 24.64↑ & 56.30↑  & 60.89↑  & 57.02↑ & \cellcolor[HTML]{E7E6E6}61.46↑ \\

\hline
\end{tabular}
\caption{Accuracy (\%) on the AstroChart benchmark. ``Infer.'' denotes Inference, and ``Chart Desc.'' denotes Chart Description, and ``KB-Infer.'' denotes KB-Inference. Bold numbers indicate the best-performing model among proprietary and open-source MLLMs, respectively (see Appendix M for model architecture details).}
\label{tab:model_evaluation_results}
\end{table*}

In FQA, models performed strongly on visual understanding tasks—top performers such as Gemini-2.5-Pro and GPT-4o achieved over 85\% accuracy across categories like color, style, and layout, indicating mature capabilities in recognizing and interpreting visual elements. In contrast, data-centric tasks, especially those involving interval comparison and numerical calculation, remained more challenging. Although leading models exceeded 60\% on interval questions, calculation accuracy typically stayed below 50\%, exposing a gap in quantitative reasoning. 

For AQA, which focuses on knowledge-based inference, performance declined further. Even top models scored below 75\%, showing the challenge of integrating chart evidence with astronomy knowledge. In the human baseline, researchers achieved only 39\%, far lower than leading VLMs, suggesting that even experts face limits beyond their subfields. These results confirm AstroChart as a valuable benchmark for assessing MLLMs’ scientific reasoning ability.

\subsection{Fine-Tuning a Representative MLLM}
To further assess AstroChart’s value as a training resource, we fine-tuned MiniCPM-V2.6-8B, the strongest performer among mid-sized open-source models, using a training set generated by the same pipeline as AstroChart (excluding expert validation). As shown in Table~\ref{tab:model_evaluation_results}, the fine-tuned model achieves consistent improvements across all FQA and AQA categories, with an overall gain of 5.02\%, confirming the effectiveness of our training data in enhancing both visual understanding and scientific reasoning.

To evaluate generalization, we tested the fine-tuned MiniCPM on three existing CQA benchmarks: CharXiv, ChartQA, and MMC-Benchmark. As shown in Table~\ref{tab:other_benchmark_result}, performance changes are minimal, i.e., some metrics slightly increase while others slightly drop. This suggests that the model has not overfitted AstroChart and that its learned reasoning skills remain largely transferable across domains.
\begin{table}[htbp]
\centering
\renewcommand{\arraystretch}{1.2}
\setlength{\tabcolsep}{2pt}
\footnotesize
\begin{tabular}{lccc}
\hline
\multicolumn{4}{c}{\cellcolor[HTML]{C0C0C0}\textbf{CharXiv}} \\ \hline
              & Descriptive & Reasoning & Overall            \\
MiniCPM-V2.6-8B  & 54.06     & 29.15   & 49.08   \\
MiniCPM-V2.6-8B--fine-tuned    & 54.58     & 29.30   & \textbf{49.52$\uparrow$} \\ \hline
\multicolumn{4}{c}{\cellcolor[HTML]{C0C0C0}\textbf{ChartQA}}          \\ \hline
              & Human       & Augmented & Overall            \\
MiniCPM-V2.6-8B  & 57.12     & 83.84   & 70.48            \\
MiniCPM-V2.6-8B-fine-tuned    & 58.24     & 80.80   & 69.52\textbf{$\downarrow$}          \\ \hline
\multicolumn{4}{c}{\cellcolor[HTML]{C0C0C0}\textbf{MMC-Benchmark}}    \\ \hline
              & MCQ         & T/F       & Overall            \\
MiniCPM-V2.6-8B & 66.46            & 77.56            & 74.40            \\
MiniCPM-V2.6-8B-fine-tuned   & 62.38            & 77.42            & 73.28\textbf{$\downarrow$}          \\ \hline
\end{tabular}
\caption{Performance of MiniCPM-V2.6-8B before and after fine-tuning on various CQA benchmarks.}
\label{tab:other_benchmark_result}
\end{table}

\subsection{Difficulty Comparison with CharXiv}
\begin{figure}[htb]
\centering
\includegraphics[width=\linewidth]{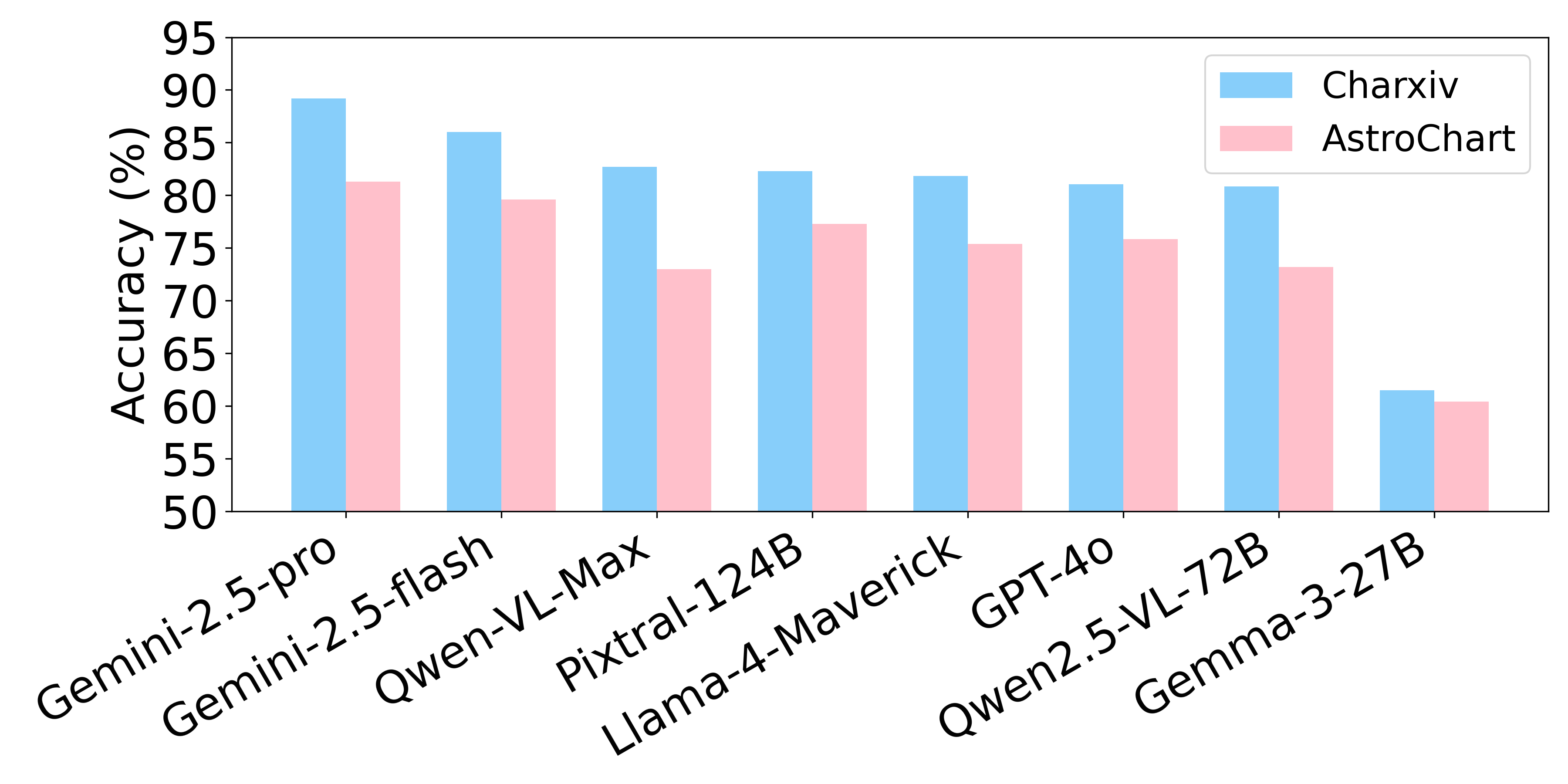}
\caption{Performance comparison of MLLMs on Charxiv and AstroChart.}
\label{fig:benchmark_comparison}
\end{figure}

Figure~\ref{fig:benchmark_comparison} compares model performance on AstroChart and CharXiv. We choose CharXiv for this comparison because, among existing benchmarks, it contains charts with the second-highest overall visual complexity after AstroChart (see Figure~\ref{fig:ccsDist}). To ensure fairness, we randomly sample 1,600 QA pairs from CharXiv to match AstroChart in size. Despite this, we observe a consistent performance drop across multiple MLLMs on AstroChart, highlighting its greater difficulty. All evaluations follow a unified metric framework for consistency.

This gap stems not only from complex visual structures but also from domain-specific questions that require deeper scientific reasoning rather than shallow visual interpretation.

\subsection{Evaluation of AQA Generation on Domains}
\label{otherdomains}

To evaluate the generalizability of DomainCQA across disciplines, we conduct a pilot study in four additional scientific domains: biochemistry, economics, medicine, and social science. While the full benchmark includes both FQA and AQA components, we focus on AQA, which selects chart abstracts and generates knowledge-intensive questions. In contrast, FQA involves domain-aware sampling and requires minimal downstream evaluation. This study examines whether AQA can reliably identify knowledge-centric charts and generate high-quality, domain-relevant QA pairs across diverse fields.

Domain experts independently assess each QA pair along two dimensions. Domain relevance is scored on a 1–5 scale, with higher scores indicating deeper and more precise use of domain-specific knowledge beyond what is directly shown in the chart. QA validity is scored as 1 (correct), 0 (cannot determine), or -1 (incorrect), based on clarity of the question and factual correctness of the answer.

As shown in Table~\ref{tab:chart_selection_effectiveness}, AQA-generated QA pairs generally receive higher scores in both relevance and validity compared to those from randomly sampled FQA charts across all domains (see Appendix O for rating criteria). This expert validation confirms the robustness and adaptability of the DomainCQA methodology, supporting its application to a broad range of scientific fields.

\begin{table}[t]
\centering
\small
\begin{tabular}{lcc}
\toprule
\textbf{Domain} & \textbf{Relevance (R/A)} & \textbf{Validity (R/A)} \\
\midrule
Astronomy        & 3.27 / 3.84  & 0.73 / 0.76 \\
Biochemistry     & 3.48 / 3.80  & 0.86 / 0.98 \\
Economics        & 3.48 / 3.78  & 0.95 / 0.98 \\
Medicine         & 3.21 / 3.77  & 0.89 / 0.96 \\
Social Science   & 3.23 / 3.46  & 0.95 / 0.93 \\
\bottomrule
\end{tabular}
\caption{Expert validation scores for QA pairs generated from randomly sampled charts from the FQA pool (R) vs. AQA-selected charts (A).}
\label{tab:chart_selection_effectiveness}
\end{table}
\subsection{Discussion}
\label{discussion}
\paragraph{Limitations Revealed by AstroChart}
AstroChart highlights key weaknesses in current MLLMs when handling scientific charts. Most models do well on visual tasks like identifying layouts or chart types, but struggle with detailed perception, especially in distinguishing similar colors or reading small labels. Their numerical reasoning is also weak, that is, models often misread axis values or return full axis ranges instead of specific intervals. On calculation tasks, such issues are made worse by OCR errors and limited math skills.

AQA evaluation reveals deeper challenges in domain understanding. Many models give vague, generic responses, confuse scientific ideas, or misuse technical terms. This shows a clear gap in vision-language alignment and the lack of embedded scientific knowledge. A major reason is that most vision-language pretraining relies on generic image–caption pairs, which fail to expose models to the structured layouts and domain-specific terminology found in scientific charts (see Appendix N for failure cases).

\paragraph{Effectiveness of DomainCQA}
Our results demonstrate the effectiveness of DomainCQA as both a benchmark construction framework and a practical training pipeline. By reusing the same generation methodology to build a fine-tuning set without targeting specific weaknesses, we cover challenging tasks like data interpretation, visual discrimination, and domain-informed inference. Fine-tuning on this dataset consistently improves performance on both FQA and AQA tasks, showing the QA pairs’ informativeness and training value. The fine-tuned model also performs well on external benchmarks such as CharXiv, ChartQA, and MMC-Benchmark, indicating it has not overfit to AstroChart and that its reasoning skills transfer across domains. Moreover, DomainCQA can be easily applied to other scientific fields, highlighting its generalizability as a domain-independent CQA construction pipeline.

\section{Conclusion \& Future Work}

\paragraph{Conclusion}
We present \textbf{DomainCQA}, a structured methodology for building domain-specific chart QA benchmarks, and demonstrate its effectiveness through \textit{AstroChart}, the first CQA benchmark for astronomy. AstroChart captures both basic chart understanding and domain-informed reasoning. Through extensive evaluation of 21 MLLMs, we reveal consistent weaknesses in chart understanding, especially when models integrate visual features with domain-specific knowledge. In addition to AstroChart, we apply DomainCQA to four scientific fields, such as biochemistry, economics, medicine, and social science, conducting pilot AQA studies with expert validation. These results confirm the generality and effectiveness of our methodology in producing high-quality, relevant, and challenging QA pairs. Furthermore, using data generated by DomainCQA for fine-tuning significantly improves MLLM performance across diverse chart reasoning tasks without overfitting, highlighting the training utility of our pipeline.

\paragraph{Future Work}
Building on our preliminary exploration across multiple scientific domains, we plan to extend DomainCQA into a broader suite of benchmarks in multiple scientific domains. Our long-term goal is to establish DomainCQA as a standard framework for chart-based scientific reasoning in real-world MLLM applications.

\section*{Ethical Statement}
This work does not involve human or animal subjects. All data used in this work are chart-based and originate from publicly available scientific publications. These materials were accessed solely for research purposes, and no proprietary, confidential, or human-related information is involved. No ethical concerns were identified in the construction of the benchmarks and experiments.

\section*{Acknowledgments}
We sincerely thank the anonymous reviewers and contributing researchers for their valuable feedback. This research was supported by the National Natural Science Foundation of China (U22A2032), the Leading Innovation and Entrepreneurship Team of Zhejiang Province of China (Grant No. 2023R01008), Zhejiang Provincial Science and Technology Plan Project (2023C01120), Key R\&D Program of Zhejiang (2024SSYS0012), and the China Manned Space Project (CMS-CSST-2025-A21).

\bibliography{main}

\clearpage
\onecolumn
\appendix
\renewcommand{\thesection}{\Alph{section}}      
\renewcommand{\thesubsection}{\Alph{section}.\arabic{subsection}}


\clearpage
\section{Appendix}
\section{A. Details of Chart Complexity Vector (CCV)}
\label{app:details_for_ccv}
\subsection{A.1. The Definition of CCV}
\label{app:definition_ccv}
To quantify the complexity of scientific charts, we introduce the Chart Complexity Vector (CCV), 
\cref{tab:definition_ccv} defines the Chart Complexity Vector (CCV), which quantifies chart complexity across ten attributes categorized into visual complexity (annotation, color, legend, pattern), data interpretation complexity (axis, element, formula, scale), and structural complexity (subplot, type). Each attribute is assigned a binary score of 0 (simple) or 1 (complex) based on specific criteria.

\begin{table}[htbp]
\centering
\resizebox{1\textwidth}{!}{
\begin{tabular}{c p{4cm} p{8cm}}
\hline
\textbf{\#} & \textbf{Attributes} & \multicolumn{1}{c}{\textbf{Complexity Definition (0 or 1)}} \\ \hline
\multicolumn{2}{l}{\textbf{Visual Complexity}} &  \\ \hline
\multirow{2}{*}{1} & \multirow{2}{=}{ Annotation Complexity} & 0: No annotations are used. \\  
& & 1: Extensive annotations such as descriptive text, arrows, or markers. \\ \hline

\multirow{2}{*}{2} & \multirow{2}{=}{ Color Complexity} & 0: Monochrome or up to two distinct colors. \\  
 & & 1: More than two distinct colors or employs a color bar. \\ \hline

\multirow{2}{*}{3} & \multirow{2}{=}{ Legend Complexity} & 0: Contains no legend or a minimal legend (no more than three categories). \\  
 & & 1: Contains a complex legend (more than three categories or intricate representations). \\ \hline

\multirow{2}{*}{4} & \multirow{2}{=}{ Pattern Complexity} & 0: No more than two distinct graphical patterns. \\  
& & 1: More than two distinct graphical patterns. \\ \hline

\multicolumn{2}{l}{\textbf{Data Interpretation Complexity}} &  \\ \hline

\multirow{2}{*}{5} & \multirow{2}{=}{ Axis Complexity} & 0: Single axis. \\  
& & 1: Multiple dependent axes (e.g., secondary y-axis, dual x-axis). \\ \hline

\multirow{2}{*}{6} & \multirow{2}{=}{ Element Complexity} & 0: Contains a single graphical element. \\  
 & & 1: Contains more than two graphical elements. \\ \hline

\multirow{2}{*}{7} & \multirow{2}{=}{ Scale Complexity} & 0: Employs a linear scale. \\  
 & & 1: Utilizes logarithmic, power-law, or mixed scales. \\ \hline

\multirow{2}{*}{8} & \multirow{2}{=}{ Formula Complexity} & 0: Does not contain any mathematical formulas. \\  
& & 1: Includes mathematical formulas. \\ \hline

\multicolumn{2}{l}{\textbf{Structural Complexity}} &  \\ \hline

\multirow{2}{*}{9} & \multirow{2}{=}{Subplot Complexity} & 0: A single-panel chart. \\  
& & 1: Multiple interrelated subplots are present. \\ \hline

\multirow{2}{*}{10} & \multirow{2}{=}{Type Complexity} & 0: Uses basic chart types (e.g., bar, line, scatter, histogram, pie) with a sparse dataset. \\  
& & 1: Uses complex chart types (e.g., heatmaps, network graphs, mixed-element charts, multiple chart combinations) and/or handles high-density data (e.g., dense bar, line, scatter, or histogram charts). \\ \hline
\end{tabular}
}
\caption{Definitions of CCV Attributes}
\label{tab:definition_ccv}
\end{table}

\clearpage
\subsection{A.2. Proportion of Complexity Aspects in AstroChart}
\label{app:proportion_ccv_astrochart}
We proposed a multi-label chart classification model aimed at predicting the 10 complexity dimensions defined by the CCV framework. The model is built upon a ResNet-50 backbone, followed by 10 parallel binary classification heads corresponding to each complexity dimension. To address significant class imbalance, we employed Focal Loss, weighted random sampling, and a range of data augmentation strategies. Training was conducted on a human-annotated, multi-domain dataset covering 6 domains, consisting of 2,474 training samples, 246 validation samples and 248 testing samples. The model achieved a Macro F1 score of 61.50\%, Macro Precision of 58.15\%, and Macro Recall of 65.95\% on the testing set.

Building upon this classifier, we further analyze the CCV complexity distribution of charts in the AstroChart benchmark. As summarized in \cref{tab:proportion_atrochart}, the distribution of simple versus complex charts across the ten CCV dimensions reveals distinct structural characteristics of astronomical visualizations. The results indicate that Color Complexity (78\%) and Type Complexity (68\%) are the most frequently observed complex attributes, suggesting a prevalence of multi-colored and structurally diverse charts. In contrast, Axis Complexity (92\%) and Element Complexity (65\%) are predominantly simple, implying that most charts use a single axis and contain limited graphical elements. These statistics provide insight into the complexity characteristics of charts in AstroChart.

\begin{table}[htbp]

\centering
\small
\resizebox{0.6\textwidth}{!}{
\begin{tabular}{cccc}
\hline
\textbf{\#} & \textbf{Attributes} & \textbf{Simple} & \textbf{Complex} \\ \hline
\multicolumn{4}{l}{\textbf{Visual Complexity}} \\ \hline
1 & Annotation Complexity & 59\% & 41\% \\
2 & Color Complexity & 22\% & 78\% \\
3 & Legend Complexity & 53\% & 47\% \\
4 & Pattern Complexity & 49\% & 51\% \\ \hline
\multicolumn{4}{l}{\textbf{Data Interpretation Complexity}} \\ \hline
5 & Axis Complexity & 92\% & 8\% \\
6 & Element Complexity & 65\% & 35\% \\
7 & Scale Complexity & 58\% & 42\% \\
8 & Formula Complexity & 63\% & 37\% \\ \hline
\multicolumn{4}{l}{\textbf{Structural Complexity}} \\ \hline
9 & Subplot Complexity & 58\% & 42\% \\
10 & Type Complexity & 32\% & 68\% \\ \hline
\end{tabular}
}
\caption{The proportion of simple/complex charts across different complexity aspects in AstroChart}
\label{tab:proportion_atrochart}
\end{table}

\clearpage
\subsection{A.3. Examples of CCV in AstroChart}
\label{app:examples_ccv}
\begin{figure}[htbp]
    \centering
    \includegraphics[width=1\linewidth]{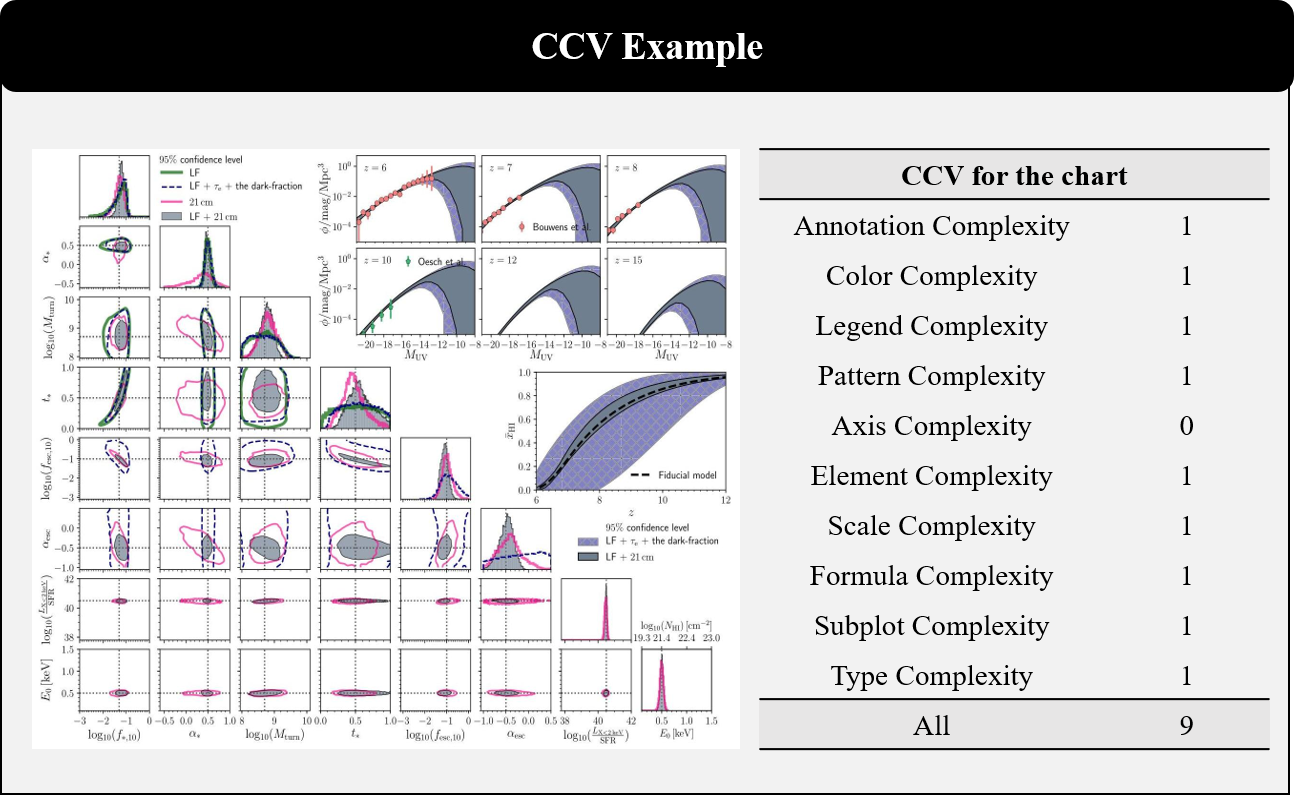}
    \caption{Example for CCV in AstroChart}
    \label{fig:ccv_1}
\end{figure}

\begin{figure}[htbp]
    \centering
    \includegraphics[width=1\linewidth]{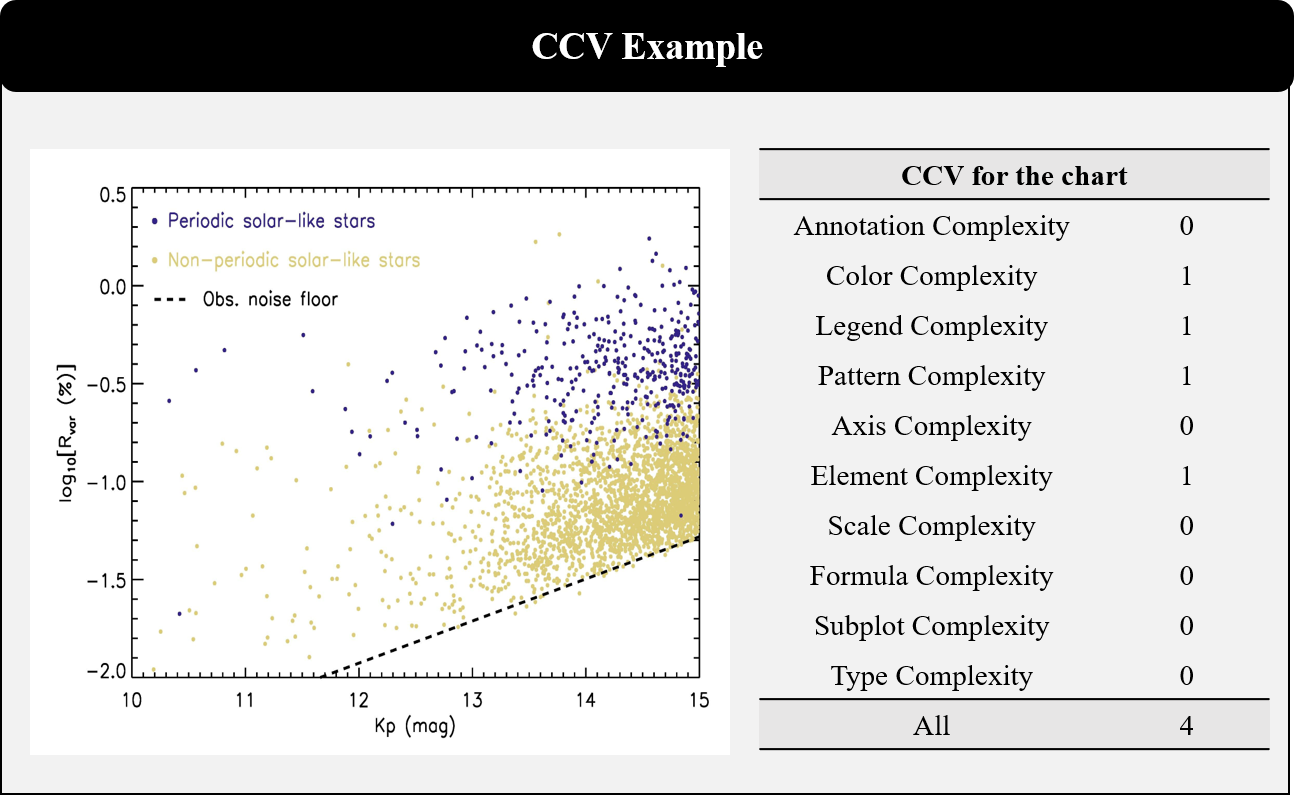}
    \caption{Example for CCV in AstroChart}
    \label{fig:ccv_2}
\end{figure}

To illustrate how CCV attributes are applied in practice, we provide concrete examples from the AstroChart dataset, 
\cref{fig:ccv_1} and \cref{fig:ccv_2} illustrate the process of computing CCV for each astronomical chart.

\clearpage
\subsection{A.4. Representative Charts by CCV Score Ranges}
\label{app:socre_example_ccv}

To better illustrate how the Chart Complexity Vector (CCV) reflects real-world chart variation, we present example charts from the AstroChart dataset corresponding to three distinct CCV score ranges: low (0–3), medium (4–7), and high (8–10). These examples demonstrate increasing levels of visual, structural, and data interpretation complexity.

\begin{figure}[htbp]
    \centering
    \includegraphics[width=0.95\linewidth]{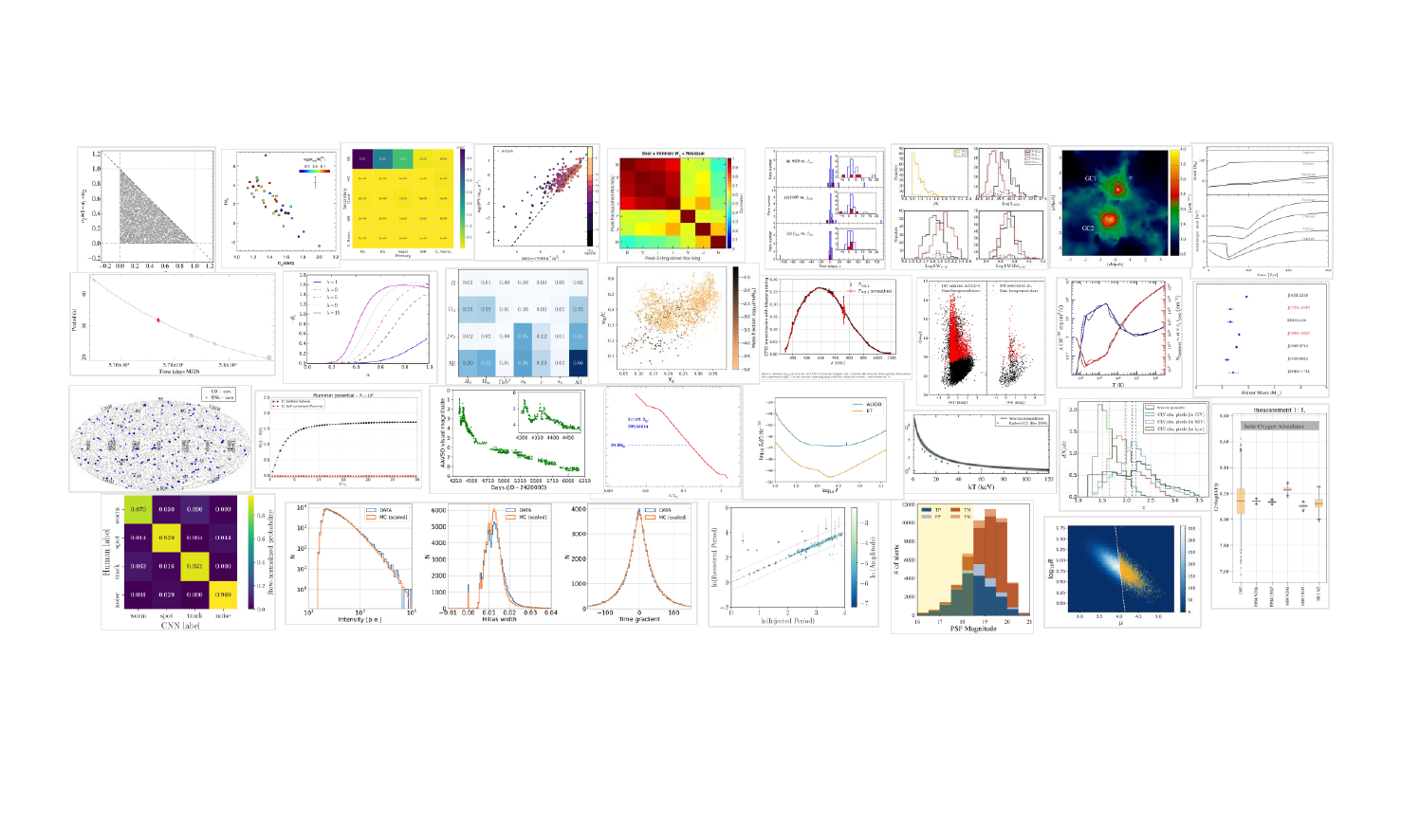}
    \caption{Example chart with low CCV score (0–3): Simple structure with minimal annotations and basic data patterns.}
    \label{fig:ccv_low}
\end{figure}

\begin{figure}[htbp]
    \centering
    \includegraphics[width=0.95\linewidth]{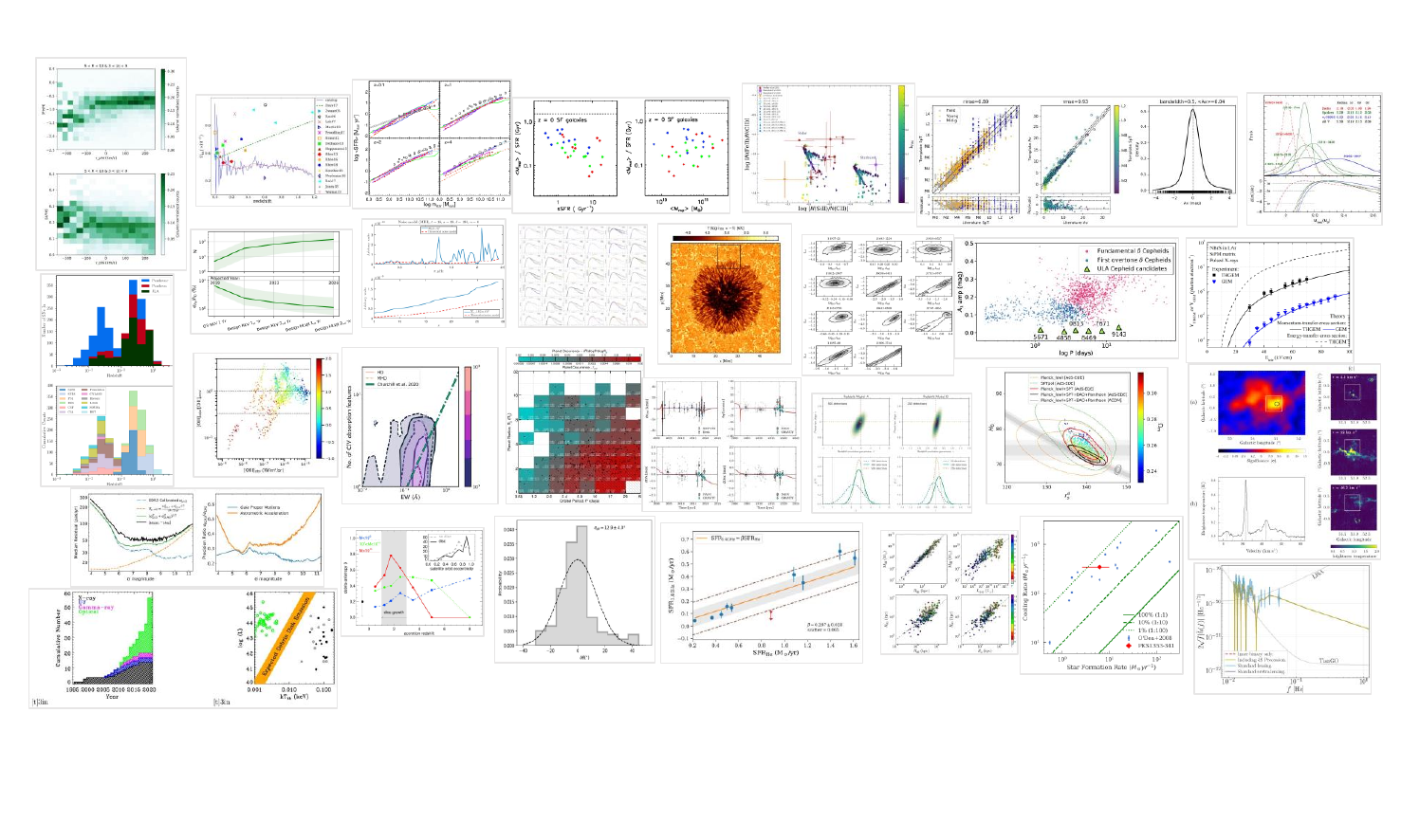}
    \caption{Example chart with medium CCV score (4–7): Moderate use of visual and structural complexity such as subplots or multiple legends.}
    \label{fig:ccv_medium}
\end{figure}

\begin{figure}[htbp]
    \centering
    \includegraphics[width=0.95\linewidth]{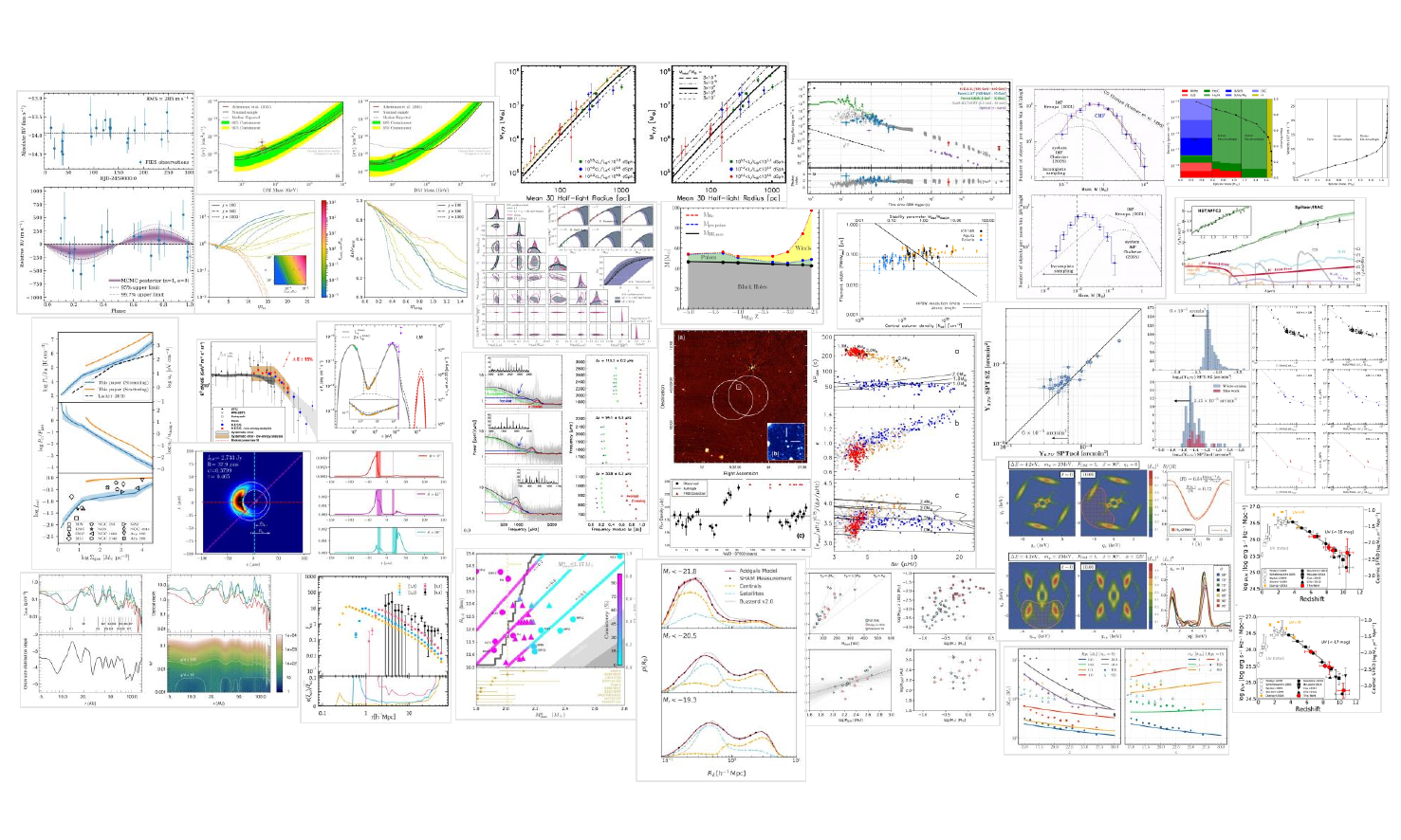}
    \caption{Example chart with high CCV score (8–10): Highly complex layout with rich annotations, multiple axes, and diverse data encodings.}
    \label{fig:ccv_high}
\end{figure}

\clearpage

\section{B. Gibbs sampling }

We employ a Gibbs sampling strategy to construct a representative set of charts for FQA generation. The detailed procedure is presented below.
\label{app:gibbs_sampling}

\begin{algorithm}[!ht]
\caption{Gibbs sampling for chart selection in FQA pairs}
\begin{algorithmic}[1]
\Require Chart dataset $D$ with $CCV(c)$ for each $c \in D$
\Ensure Selected benchmark charts $S$
\State \textbf{Initialize:} Randomly select initial $S \subset D$ of size \texttt{target\_size}
\Repeat
  \For{each chart $c^* \in S$}
    \State Select aspect $\alpha$ in $CCV$
    \State Fix other aspects, sample $v \sim P(\alpha \mid S, D)$
    \State Find $c_{\text{new}} \in D$ with $\alpha(c_{\text{new}}) = v$
    \If{$CCV(S \cup \{c_{\text{new}}\} - \{c^*\})$ is valid}
      \State Replace $c^*$ with $c_{\text{new}}$ in $S$
    \EndIf
  \EndFor
\Until{distribution stabilizes}
\State \Return $S$
\end{algorithmic}
\end{algorithm}

\clearpage
\section{C. COT\&VOT}

\label{app:cotvot}
To identify the most representative chart for generating Advanced Question-Answer (AQA) pairs, we design a CoT\&VoT-based selection framework. CoT (Chain-of-Thought) reasoning enables models to summarize chart content in a structured manner, while VoT (Voting over Thought) aggregates multiple model outputs to ensure robust selection. The algorithm below outlines how we utilize multiple LLMs to assess the alignment between each chart and the paper’s core scientific narrative (abstract and conclusion), ultimately selecting the most relevant chart via majority voting.

\begin{algorithm}[H]
    \caption{CoT and VoT for chart selection in AQA pairs}

    \label{algo:cotvot}
    \alglinenumber[1]  

    \begin{algorithmic}[1]
      \Require Paper $P$ with charts $\{C_1, ..., C_N\}$; models $\{M_1, ..., M_k\}$
      \Ensure Selected chart abstract $C^*$
      \For{each $m_j \in \{M_1, \ldots, M_k\}$}
        \State Extract abstract and conclusion, generate $P_j$
        \For{each chart $C_i$}
          \State Extract caption/description
          \State Generate summary $S_{ij}$ using $M_j$
          \State Compute relevance $R_{ij}$ with $P_j$
        \EndFor
        \State Select chart $C^*_j$ with highest $R_{ij}$
      \EndFor
      \State Identify $C^*$ by majority vote of $C^*_j$
      \Return $C^*$
    \end{algorithmic}
\end{algorithm}
\clearpage
\section{D. Prompts for question-answer pair generation in AstroChart}
\label{app:prompts_qa_generation}

We employed Claude 3.5 to generate question-answer pairs, designing distinct prompts for each category of questions. Specifically, for the two primary types: FQA pair and AQA pair.
We implemented different input configurations.
For the FQA pair, the input consisted of the chart along with its corresponding caption. For the AQA pair, the input additionally included descriptive content from the associated paper.
This differentiation was essential, as knowledge-based questions often require contextual background derived from the broader content of the paper.

\begin{figure}[htbp]
    \centering
    \includegraphics[width=\linewidth]{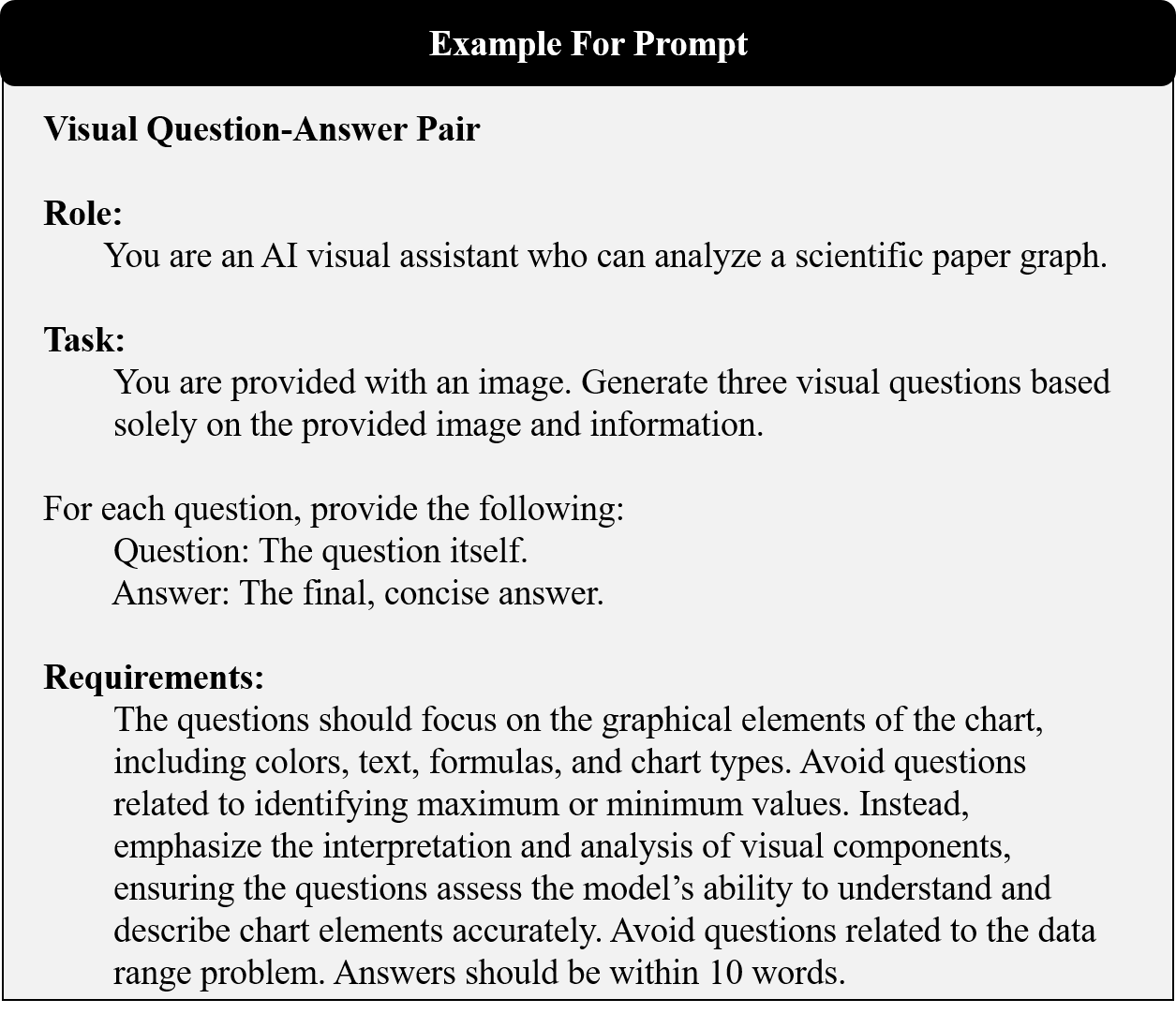}
    \caption{Prompt for visual question-answer pair generation.}
    \label{fig:visual_prompt}
\end{figure}

\begin{figure}[htbp]
    \centering
    \includegraphics[width=\linewidth]{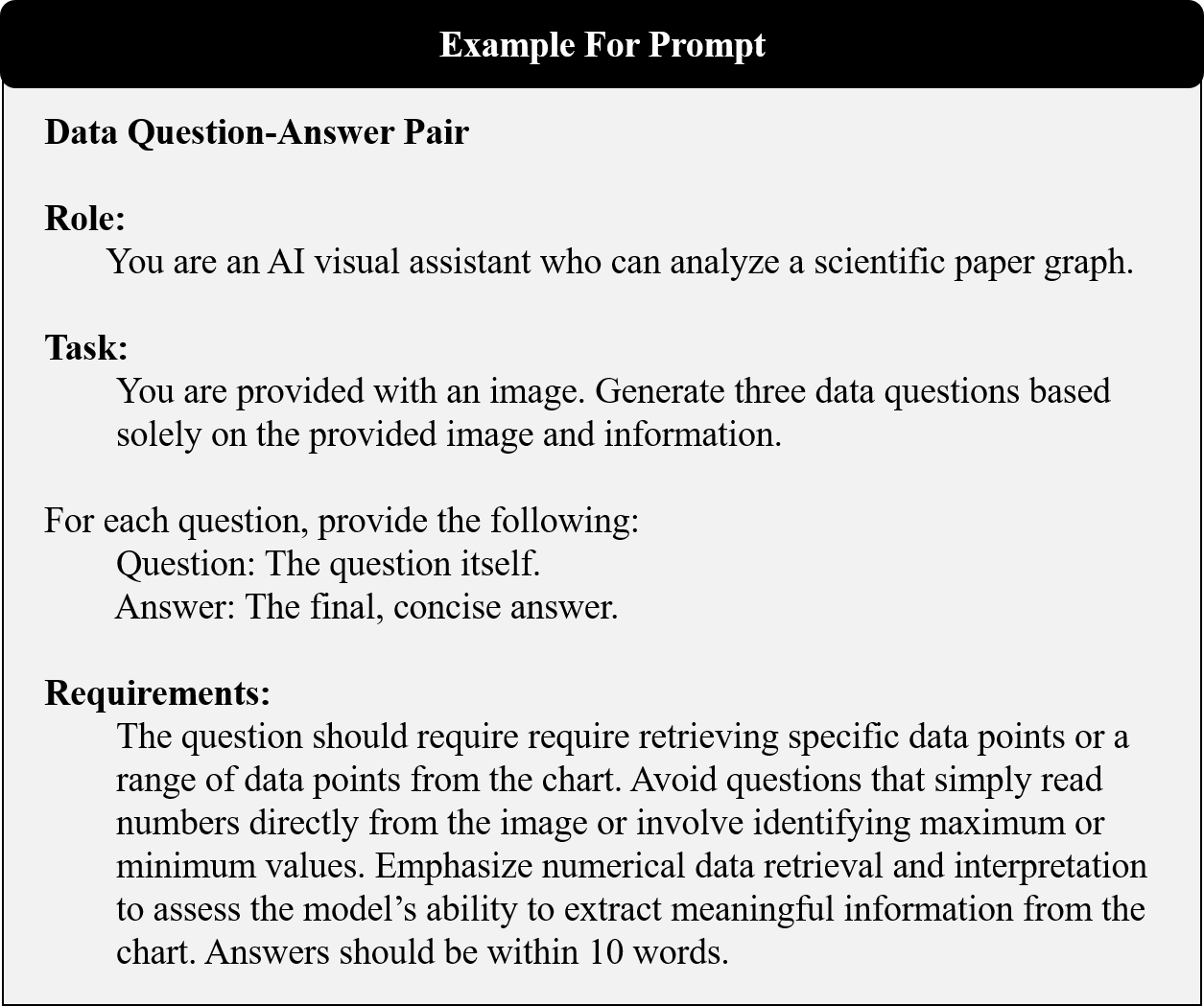}
    \caption{Prompt for data question-answer pair generation.}
    \label{fig:data_prompt}
\end{figure}

\begin{figure}[htbp]
    \centering
    \includegraphics[width=\linewidth]{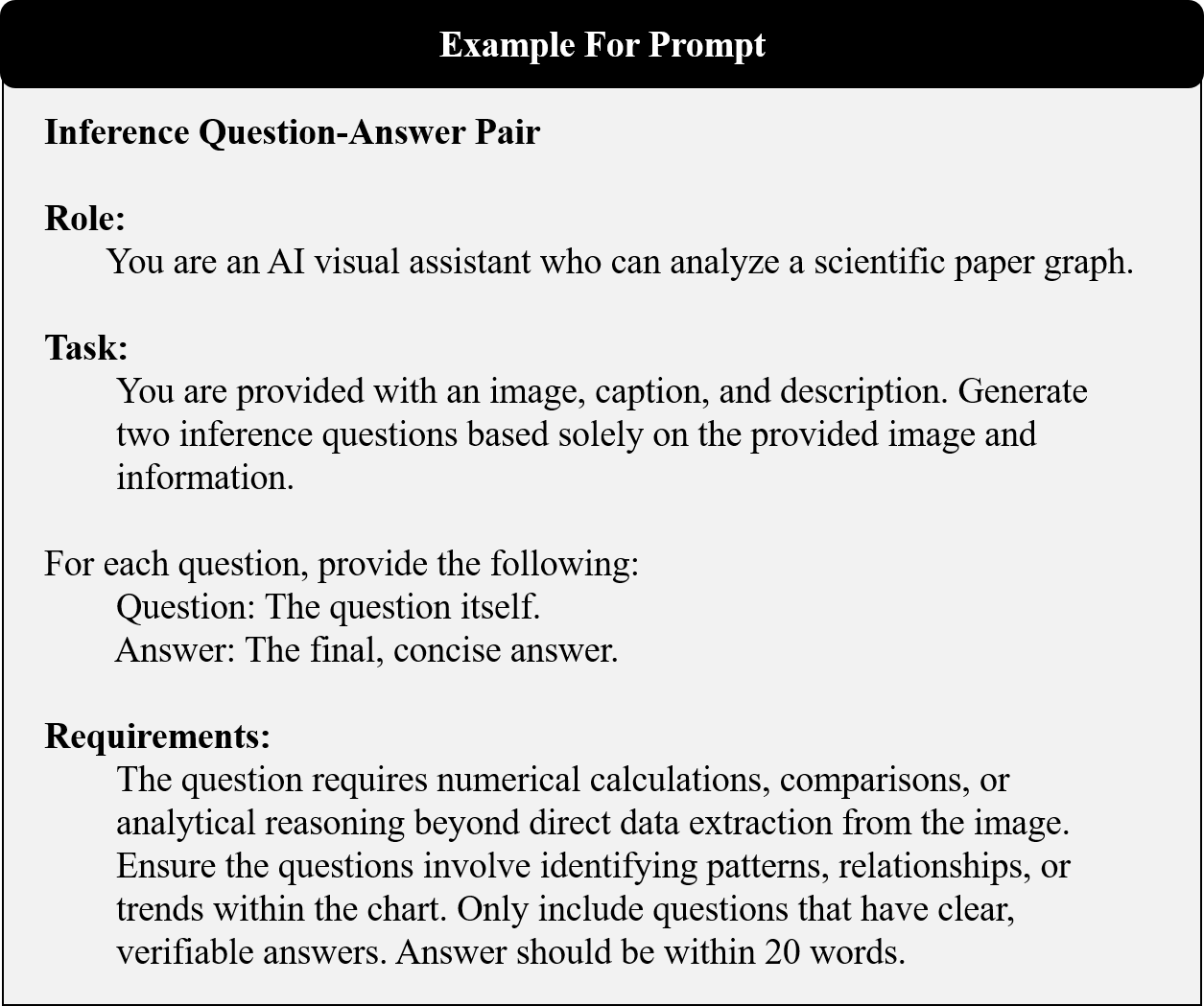}
    \caption{Prompt for inference question-answer pair generation.}
    \label{fig:inference_prompt}
\end{figure}

\begin{figure}[htbp]
    \centering
    \includegraphics[width=\linewidth]{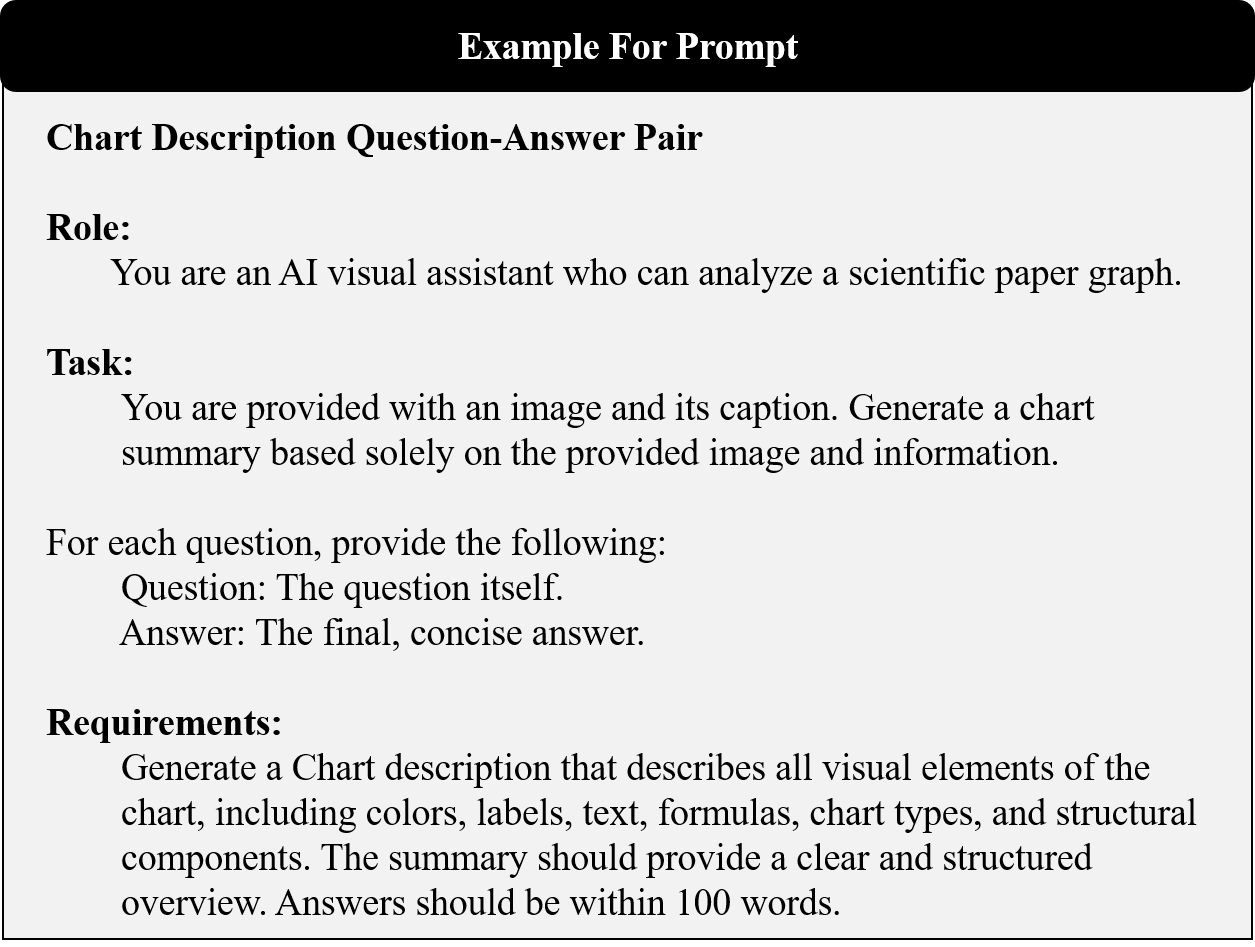}
    \caption{Prompt for chart description question-answer pair generation.}
    \label{fig:summ_prompt}
\end{figure}

\begin{figure}[htbp]
    \centering
    \includegraphics[width=\linewidth]{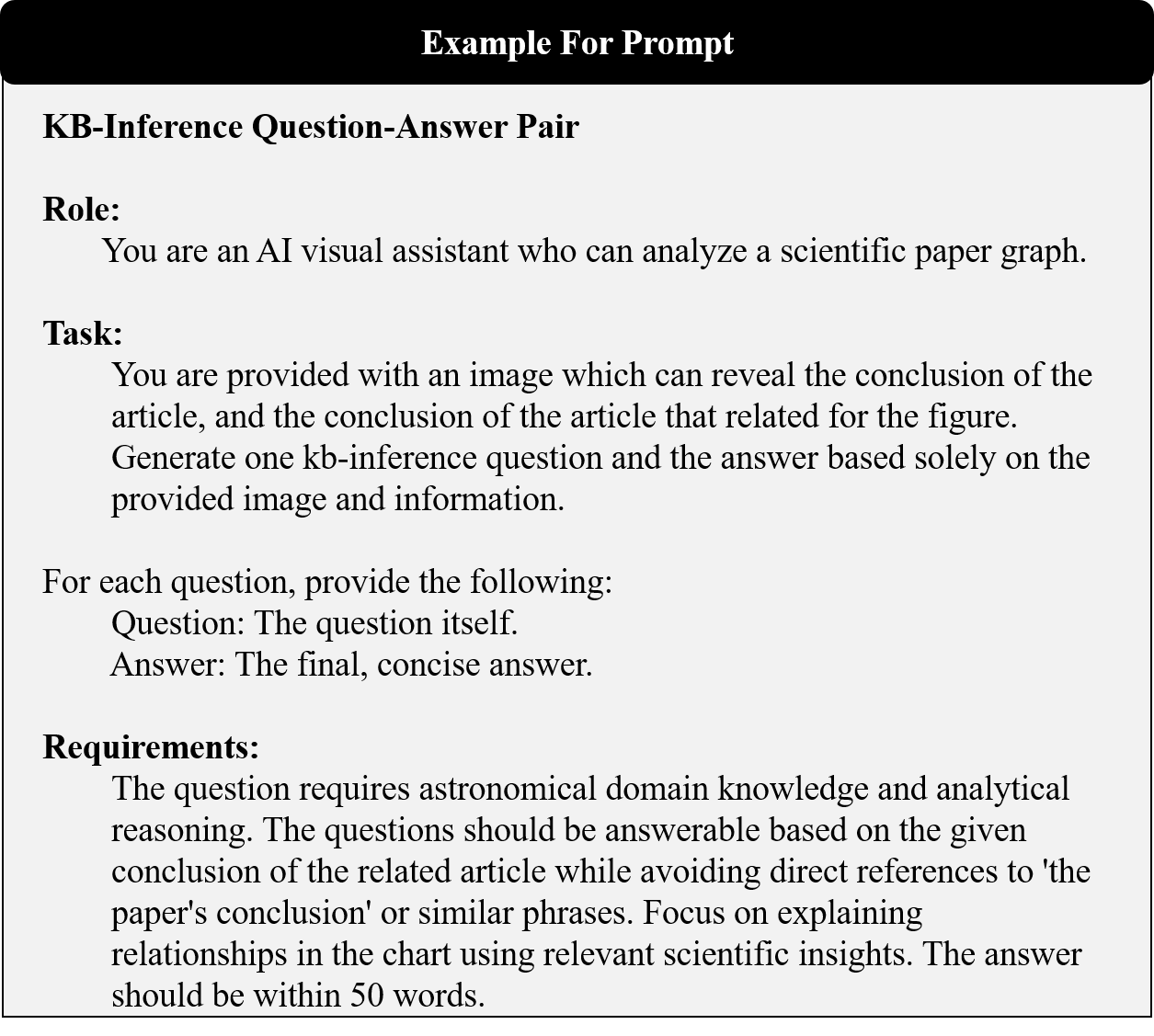}
    \caption{Prompt for KB-inference question-answer pair generation.}
    \label{fig:kb-inference_prompt}
\end{figure}

To generate different question-answer pair types, we formulated targeted prompts:
\begin{itemize}
    \item \textbf{FQA pairs:}
            \begin{itemize}
                \item \textit{Visual questions-answer pair} (\cref{fig:visual_prompt})
                The questions should focus on the graphical elements of the chart, including colors, labels, text, formulas, and chart types.
                \item \textit{Data questions-answer pair} (\cref{fig:data_prompt})
                The questions should require retrieving specific data points or a range of values from the chart.
                \item \textit{Inference questions-answer pair} (\cref{fig:inference_prompt})
                The questions should involve numerical calculations, comparisons, or analytical reasoning beyond direct data extraction from the chart.
                \item \textit{Chart Description questions-answer pair} (\cref{fig:summ_prompt})
                The task is to generate a comprehensive summary describing all visual elements of the chart, including colors, labels, texts, formulas, chart types, and structural components.
            \end{itemize}
        \item \textbf{AQA pairs:}
                \begin{itemize}
                \item \textit{KB-Inference questions-answer pair} (\cref{fig:kb-inference_prompt})
                The question requires astronomical domain knowledge and analytical reasoning, with a focus on explaining chart relationships using scientific insights, without directly referencing the article's conclusion.
            \end{itemize}    
    \end{itemize}

\clearpage
\section{E. Prompt for GPT4-o filter}

\label{app:gpt4-o filter}

To ensure the quality of generated QA pairs, we employ a GPT-4o based filtering mechanism. This appendix provides the exact prompt used to instruct GPT-4o to identify and remove low-quality QA pairs. The goal is to retain only those questions that are clearly grounded in the given chart and, in the case of AQA pairs, necessitate domain-specific knowledge to answer. This filtering step is crucial for maintaining the relevance and rigor of our benchmark.

\begin{figure}[htbp]
    \centering
    \includegraphics[width=1\linewidth]{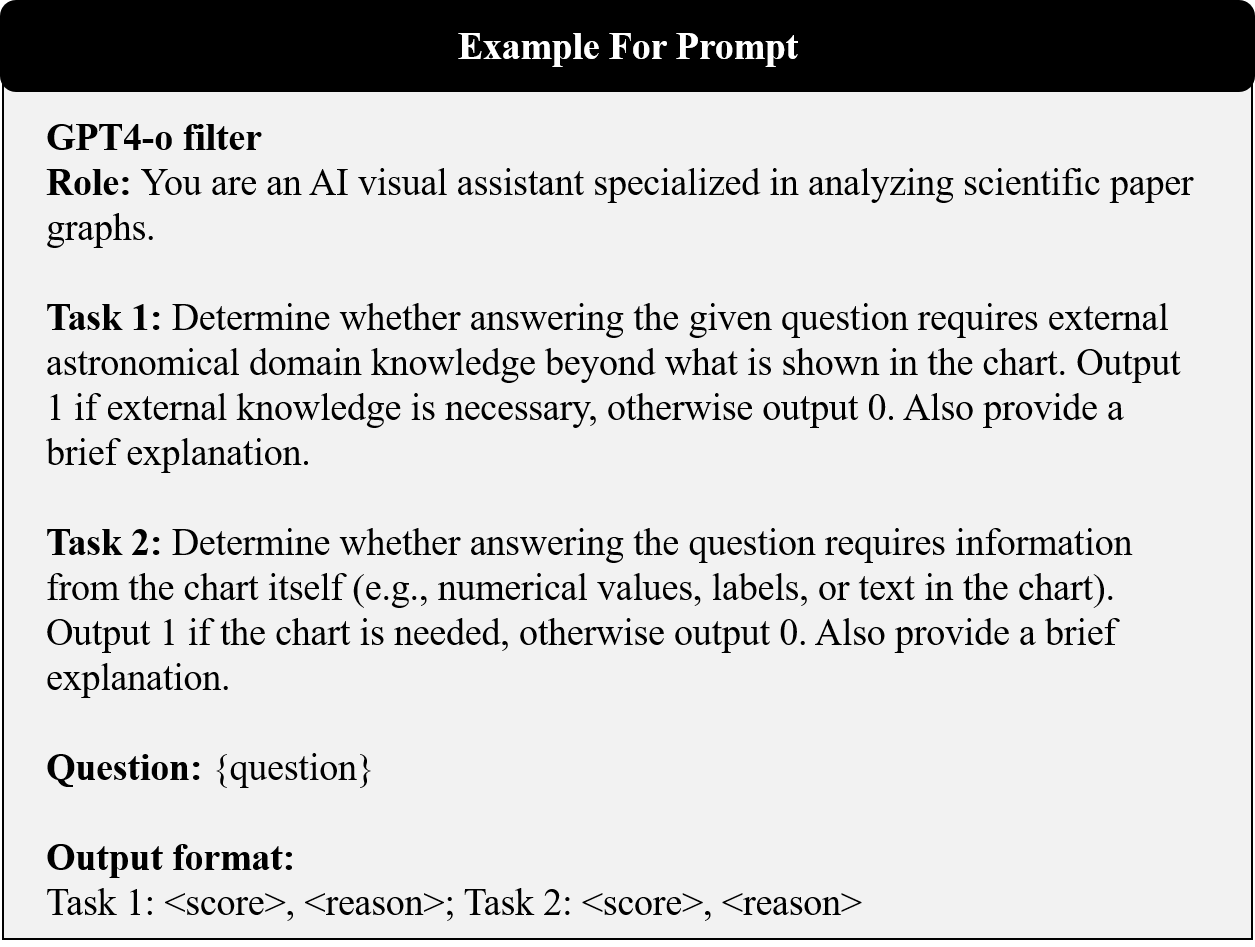}
    \caption{This is the prompt for gpt4-o filter.}
    \label{fig:gpt4-o filter}
\end{figure}

\clearpage
\section{F. Subdomains of astronomy}

\label{app:highQualityPaper}

To ensure the domain diversity and scientific rigor of AQA pairs, we identify six major subdomains within the field of astronomy: High Energy, Earth \& Planetary, Solar \& Stellar, Cosmology \& Nongalactic, Galaxies, and Instrumentation \& Methods. From each subdomain, we select the top \(1\%\) most-cited papers annually as the source material for question generation. This figure summarizes the distribution of these high-impact papers across subdomains, reflecting the relative volume of influential literature in each area.

\begin{figure}[htbp]
    \centering
    \includegraphics[width=1\linewidth]{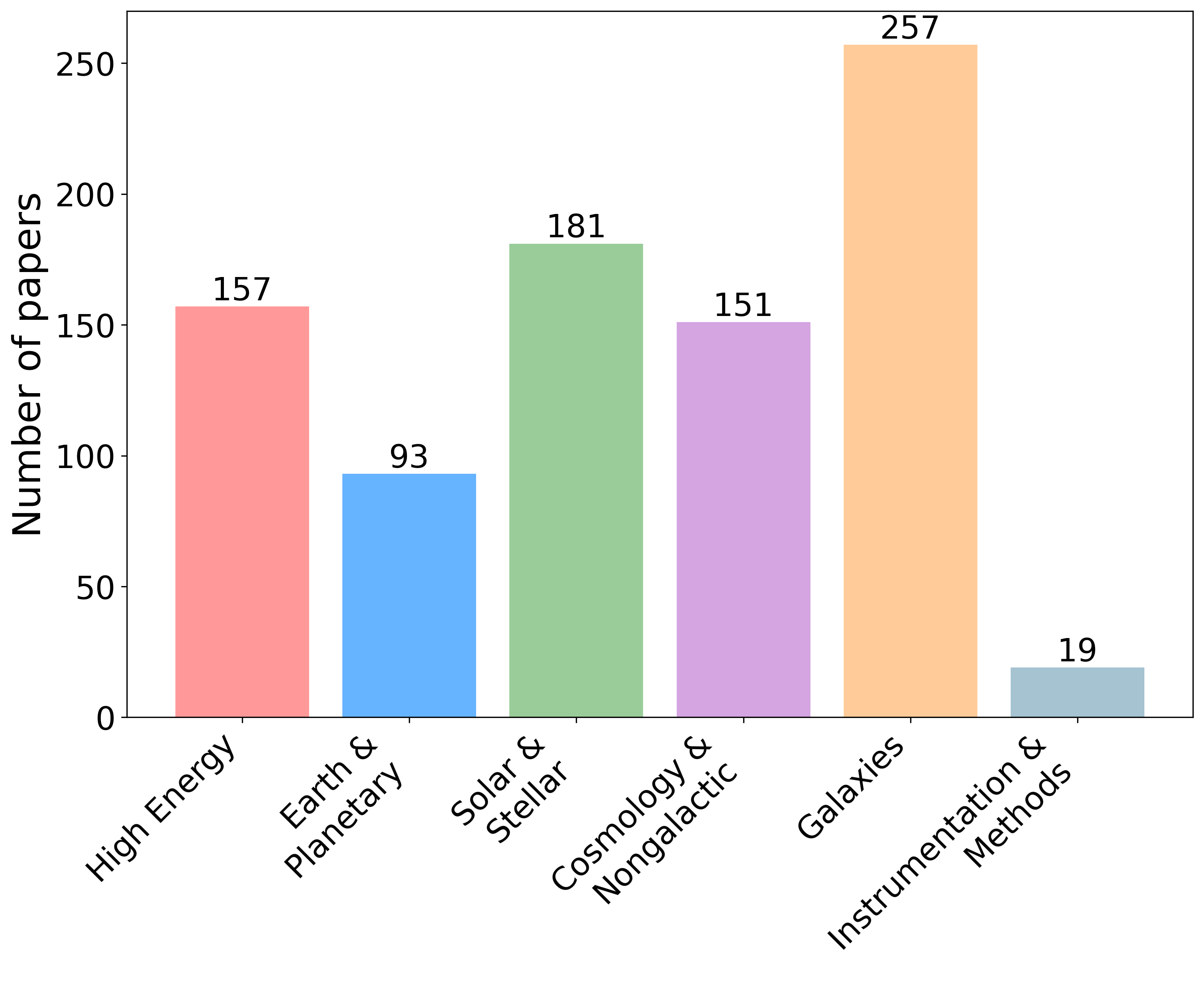}
    \caption{Top 1\% most-cited papers selected from six major subdomains of astronomy}
    \label{fig:highQualityPaper}
\end{figure}

\clearpage
\section{G. Visualization of Samples in AstroChart}
\label{app:visualization_astrochart}

\begin{figure}[ht]
    \centering
    \includegraphics[scale=0.45]{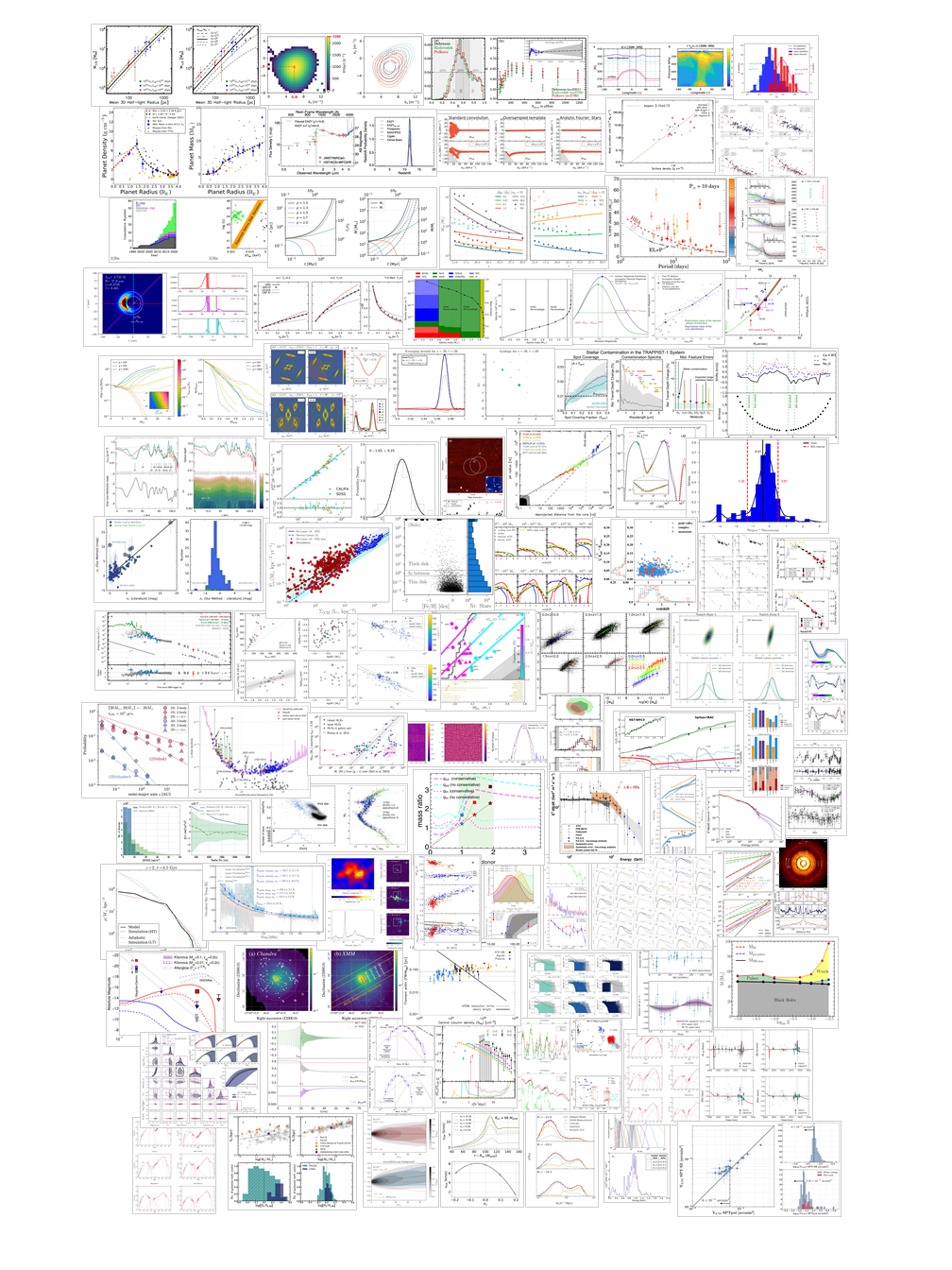}
    \caption{Visualization of Sample Charts in AstroChart}
    \label{fig:astrocqa_visualization}
\end{figure}

Fig. \ref{fig:astrocqa_visualization} visualizes sample charts from the AstroChart benchmark, illustrating its diversity and complexity.

\clearpage

\section{H. Reliability of Automated Filtering with GPT-4o}
\label{sec:gpt4o-filter}
To evaluate the reliability of the automated filtering step in identifying low-quality QA pairs, we conduct a manual verification experiment on a randomly selected set of $200$ KB-Inference QA pairs. Among these, GPT-4o filters out $19$ QA pairs. Human annotators then independently re-evaluate the same $200$ pairs using the identical filtering criteria.

We find that GPT-4o and human judgments disagree on $7$ cases: in $6$ instances, GPT-4o marked the pair for deletion while human experts judged them to be valid (false positives); in $1$ instance, the human annotators identified a pair for deletion that GPT-4o retained (false negative). The remaining $193$ pairs are consistent across both methods.

Based on this comparison, GPT-4o achieves an overall accuracy of $96.5\%$, a precision of $68.4\%$, and a recall of $92.9\%$ in identifying deletable QA pairs. Notably, most disagreements are conservative false positives, suggesting that GPT-4o tends to over-filter rather than under-filter. These results indicate that GPT-4o serves as a reasonably reliable filter for large-scale QA dataset curation, with minor manual corrections recommended for high-stakes benchmarks.

GPT-4o also demonstrates substantial agreement with human annotators, with a Cohen’s Kappa coefficient of $0.77$, indicating strong consistency in identifying deletable items. These results indicate that GPT-4o serves as a reasonably reliable filter for large-scale QA dataset curation, with minor manual corrections recommended for high-stakes benchmarks.
\begin{table}[ht]
\centering
\small
\begin{tabular}{lcc}
\toprule
 & \textbf{Human: Delete} & \textbf{Human: Keep} \\
\midrule
\textbf{GPT-4o: Delete} & 13 (True Positive) & 6 (False Positive) \\
\textbf{GPT-4o: Keep}   & 1 (False Negative) & 180 (True Negative) \\
\bottomrule
\end{tabular}
\caption{Confusion matrix for GPT-4o filtering decisions on 200 QA pairs.}
\label{tab:gpt_filter_confusion}
\end{table}

\clearpage
\section{I. Examples of AstroChart}
\label{app:examples_astrochart}
\begin{figure}[t]
    \centering
    \includegraphics[width=\columnwidth]{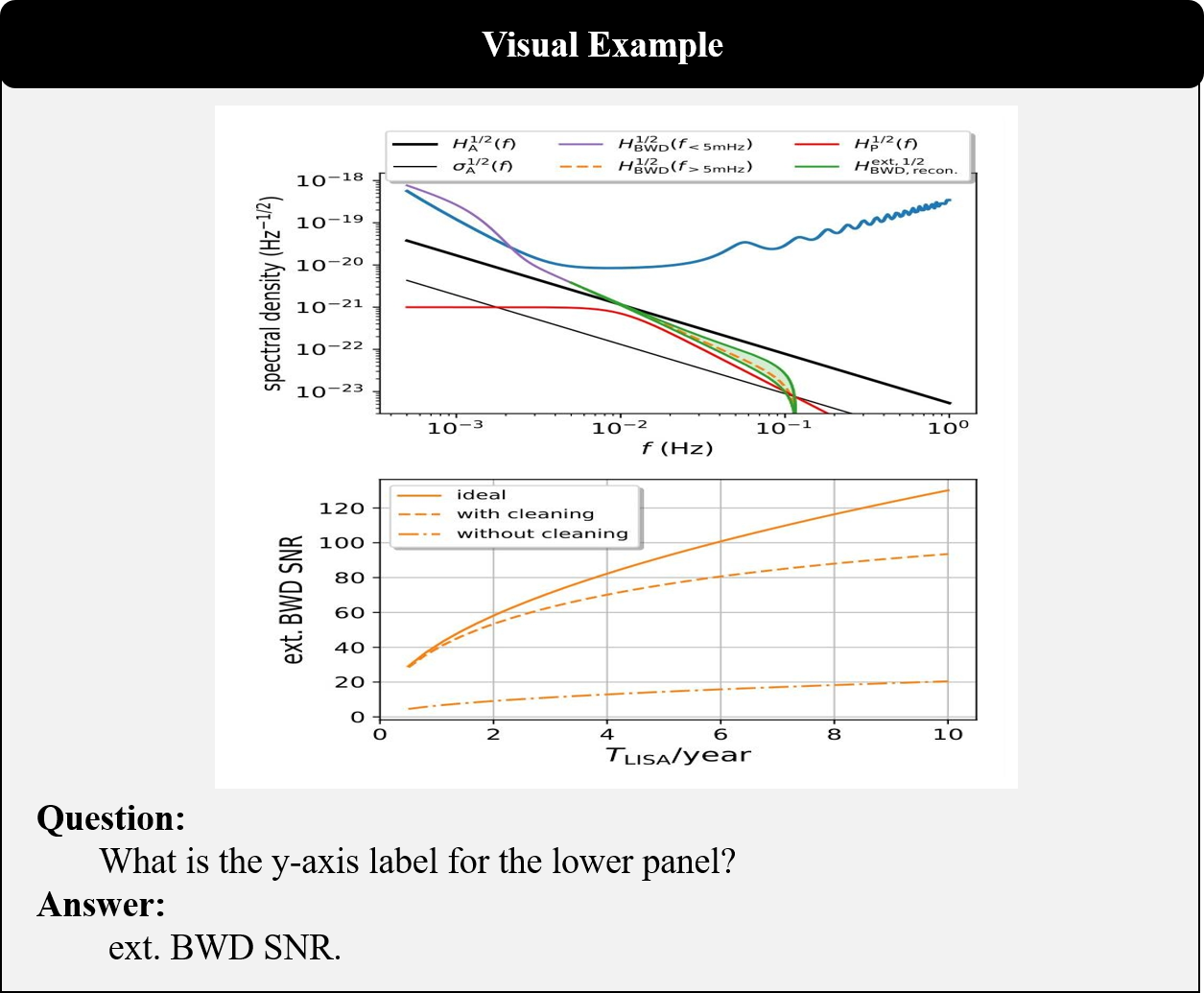}
    \caption{Example for visual question-answer pair.}
    \label{fig:visual_text}
\end{figure}

\begin{figure}[t]
    \centering
    \includegraphics[width=\columnwidth]{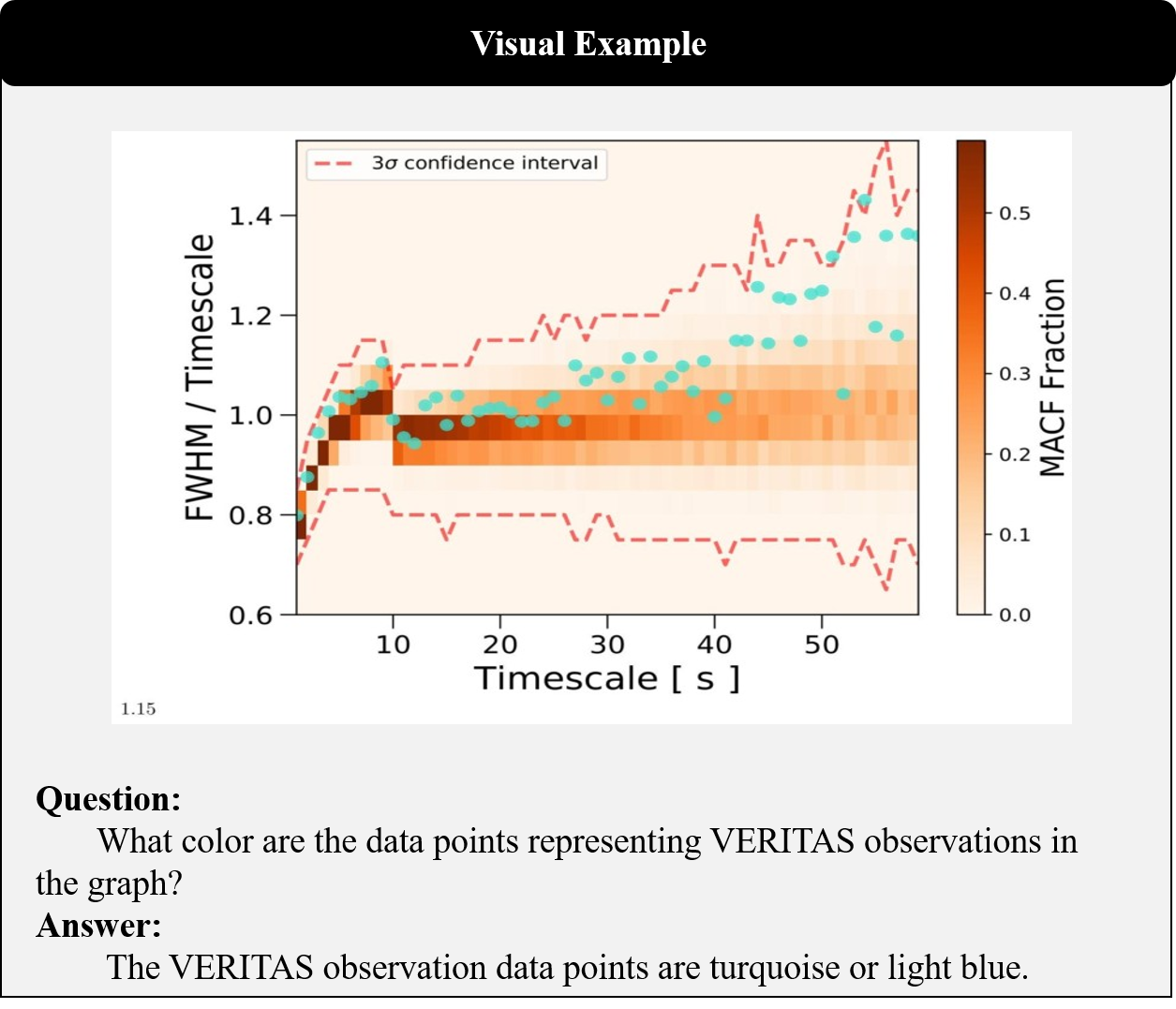}
    \caption{Example for visual question-answer pair.}
    \label{fig:visual_color}
\end{figure}

\begin{figure}[t]
    \centering
    \includegraphics[width=\columnwidth]{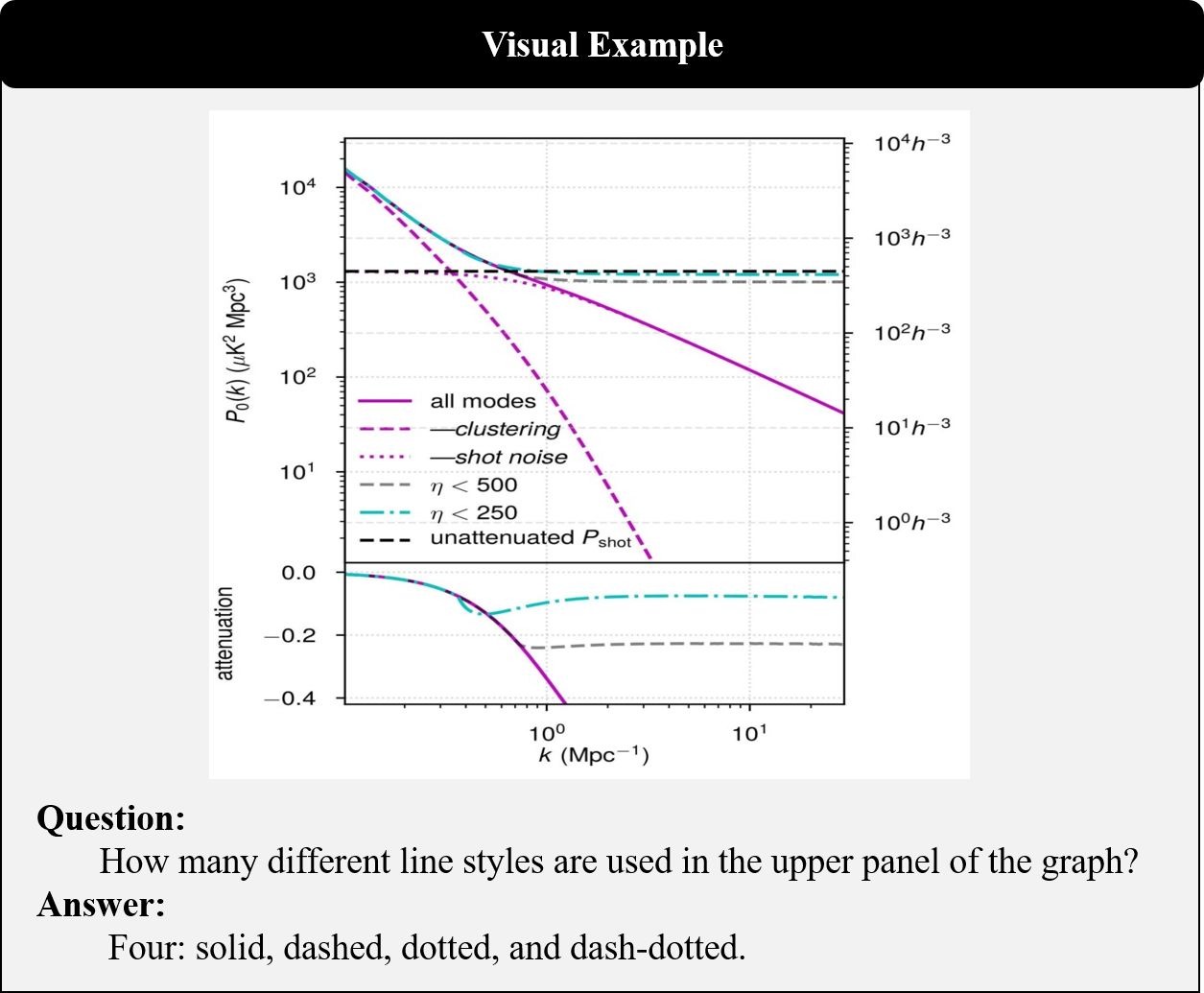}
    \caption{Example for visual question-answer pair.}
    \label{fig:visual_style}
\end{figure}

\begin{figure}[t]
    \centering
    \includegraphics[width=\columnwidth]{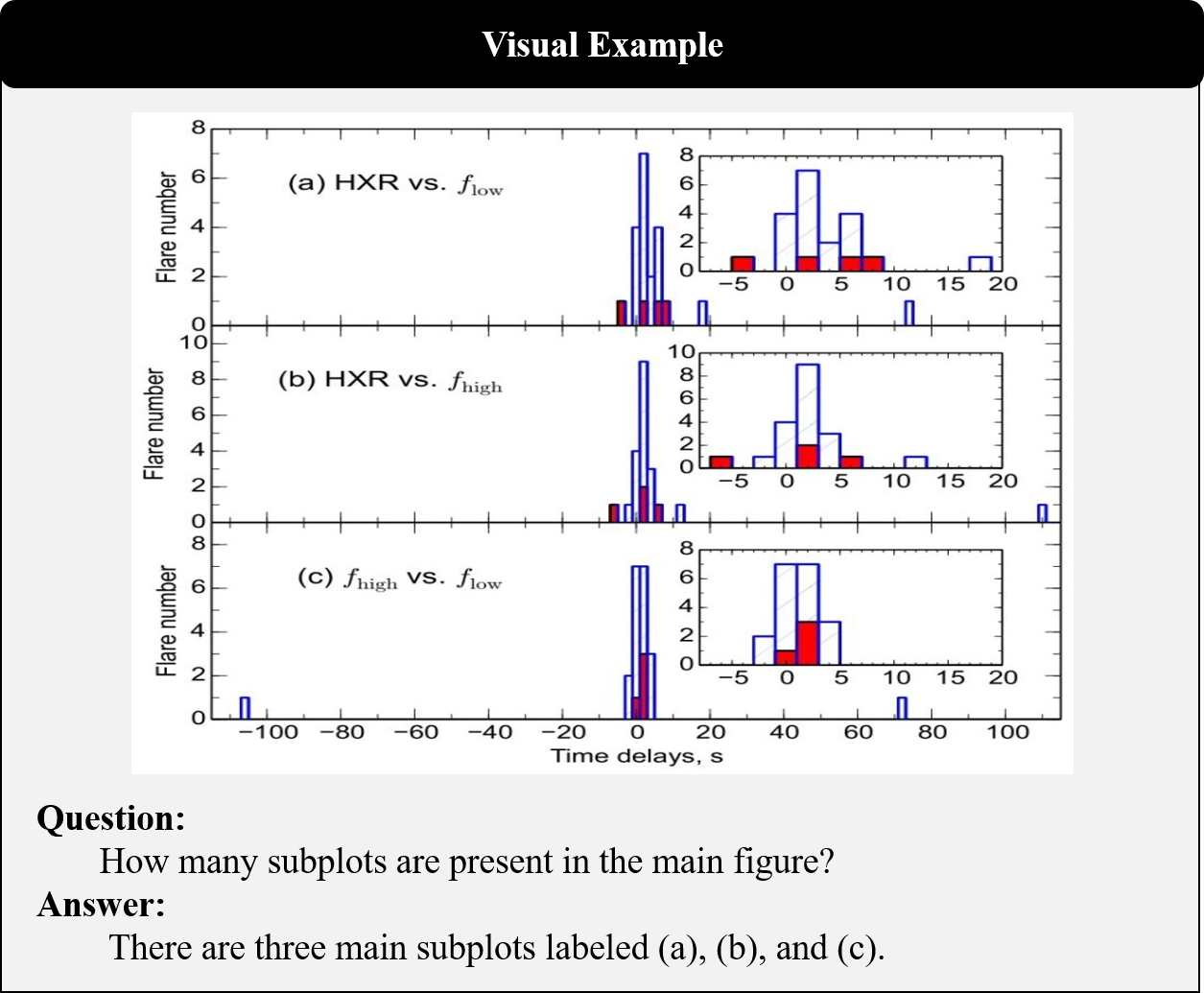}
    \caption{Example for visual question-answer pair.}
    \label{fig:visual_layout}
\end{figure}

\begin{figure}[t]
    \centering
    \includegraphics[width=\columnwidth]{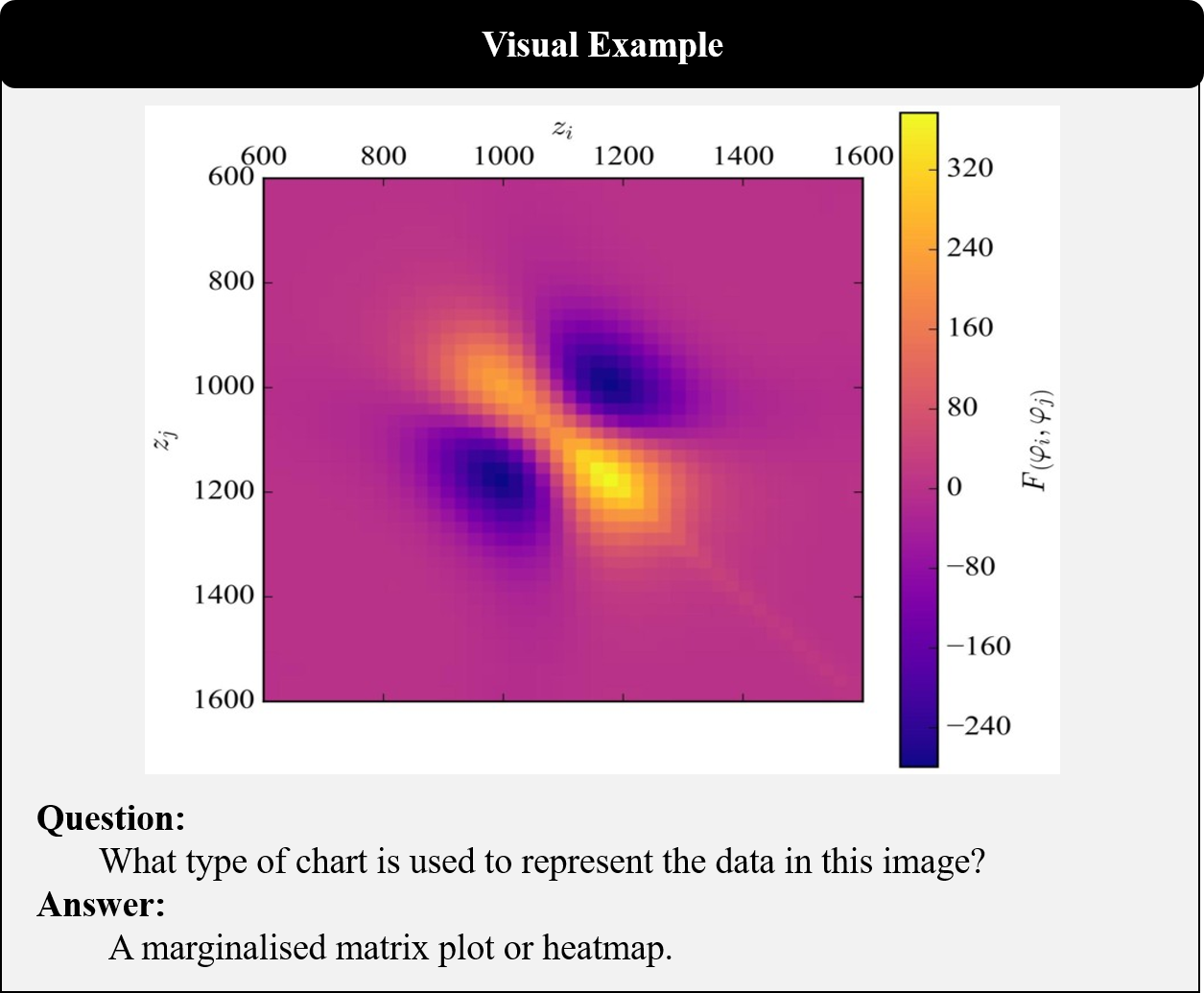}
    \caption{Example for visual question-answer pair.}
    \label{fig:visual_layout_2}
\end{figure}

\begin{figure}[t]
    \centering
    \includegraphics[width=\columnwidth]{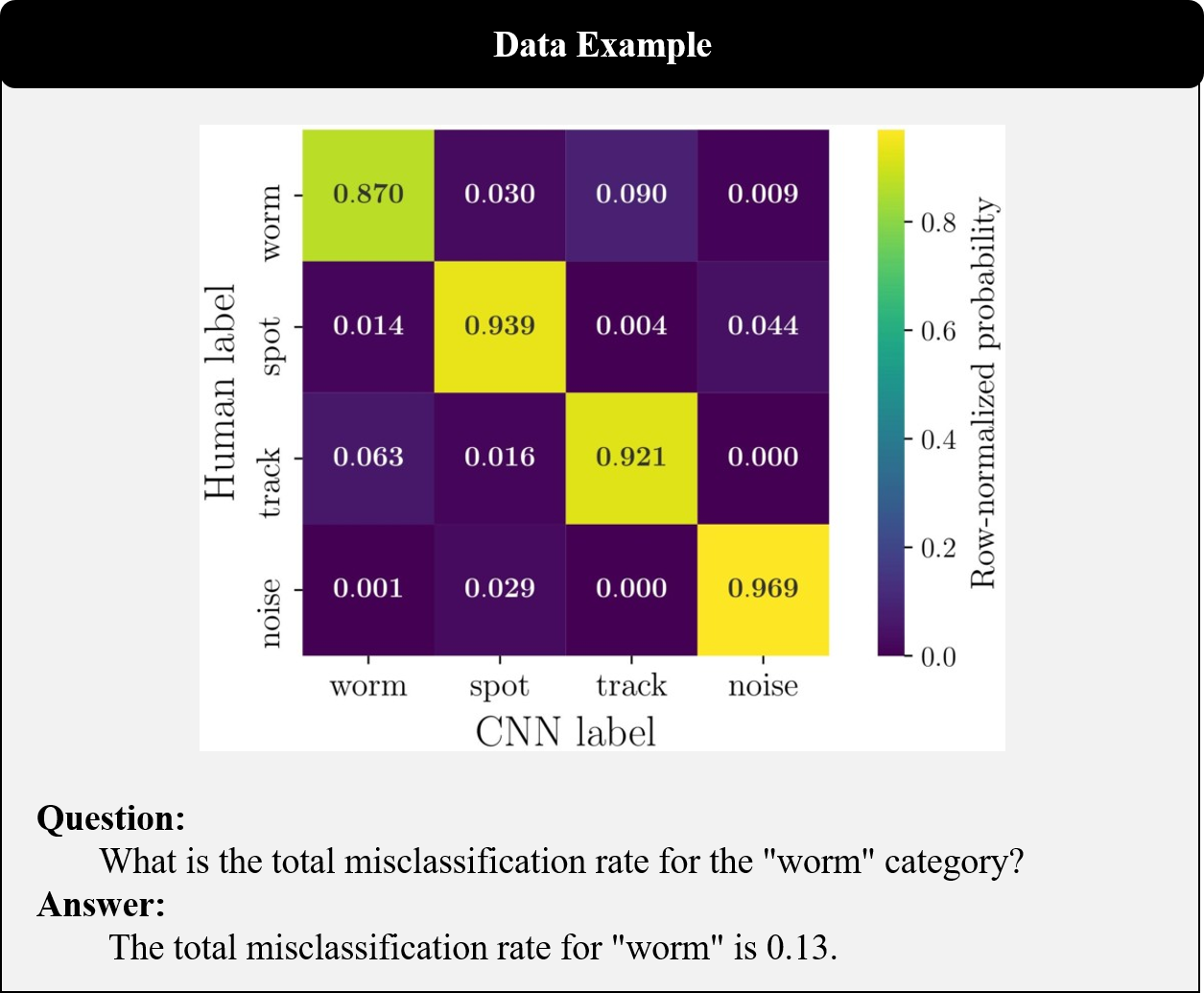}
    \caption{Example for data question-answer pair.}
    \label{fig:data_cal_1}
\end{figure}

\begin{figure}[t]
    \centering
    \includegraphics[width=\columnwidth]{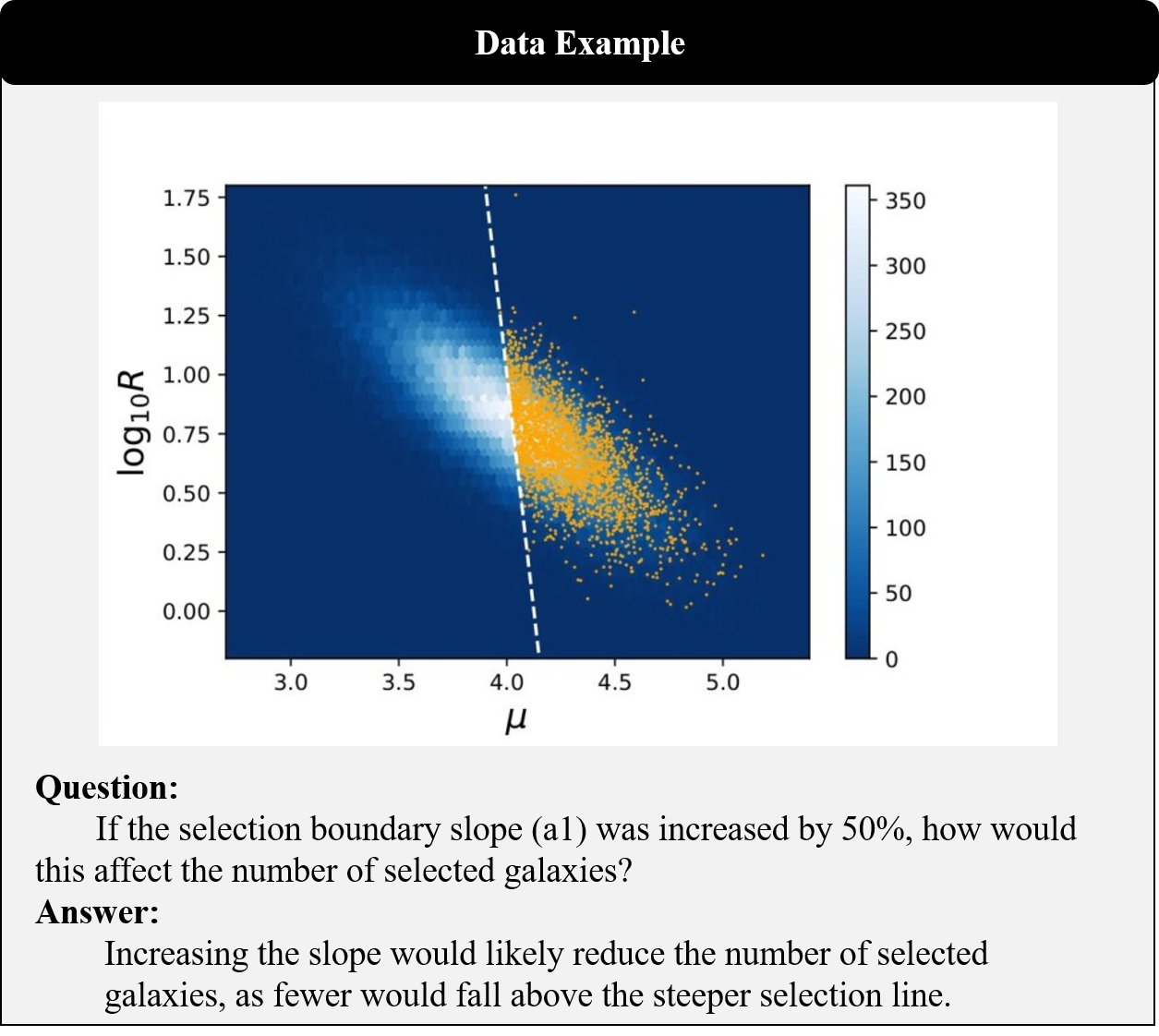}
    \caption{Example for data question-answer pair.}
    \label{fig:data_cal_2}
\end{figure}

\begin{figure}[t]
    \centering
    \includegraphics[width=\columnwidth]{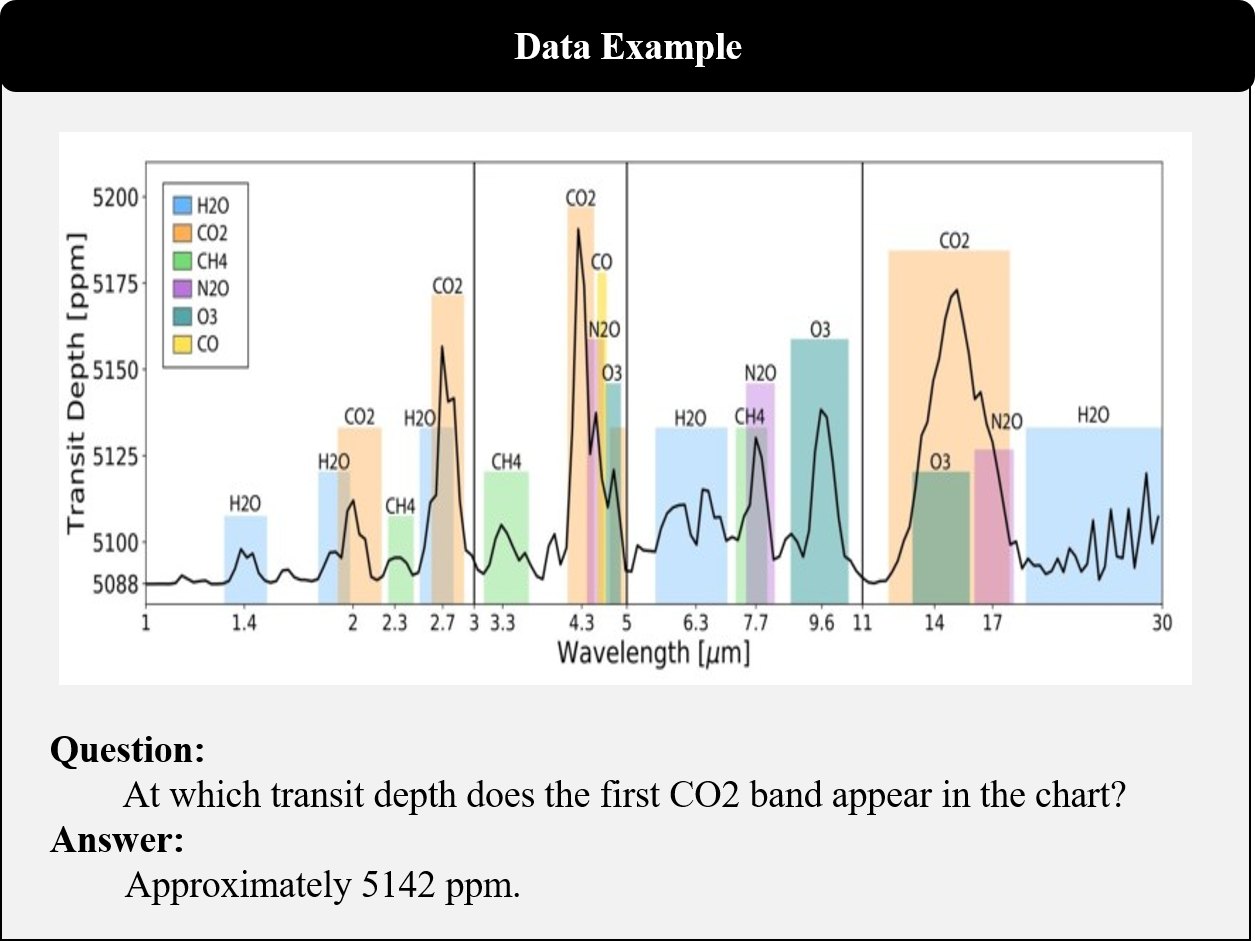}
    \caption{Example for data question-answer pair.}
    \label{fig:data_point}
\end{figure}

\begin{figure}[t]
    \centering
    \includegraphics[width=\columnwidth]{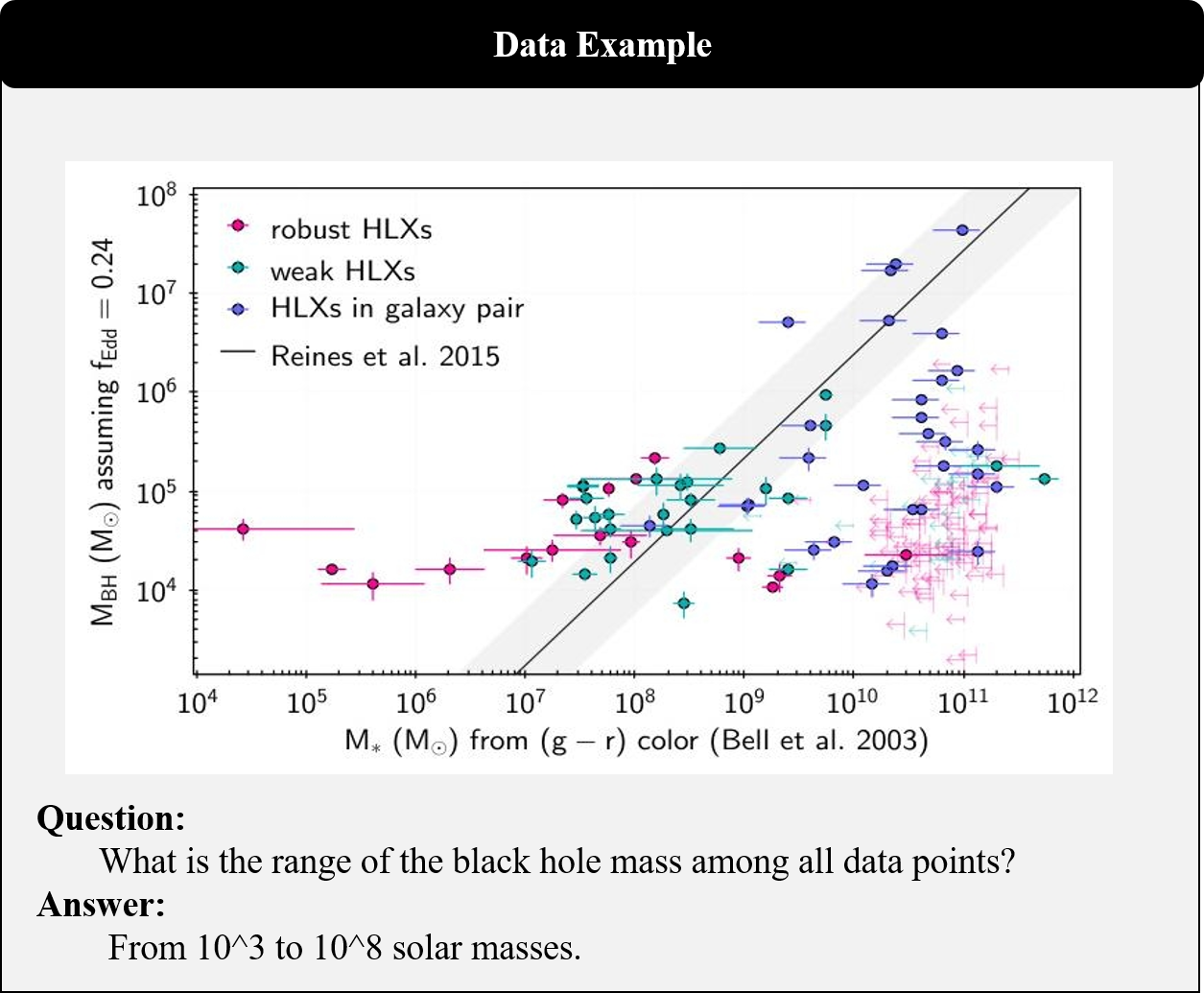}
    \caption{Example for data question-answer pair.}
    \label{fig:data_inter}
\end{figure}

\begin{figure}[t]
    \centering
    \includegraphics[width=\columnwidth]{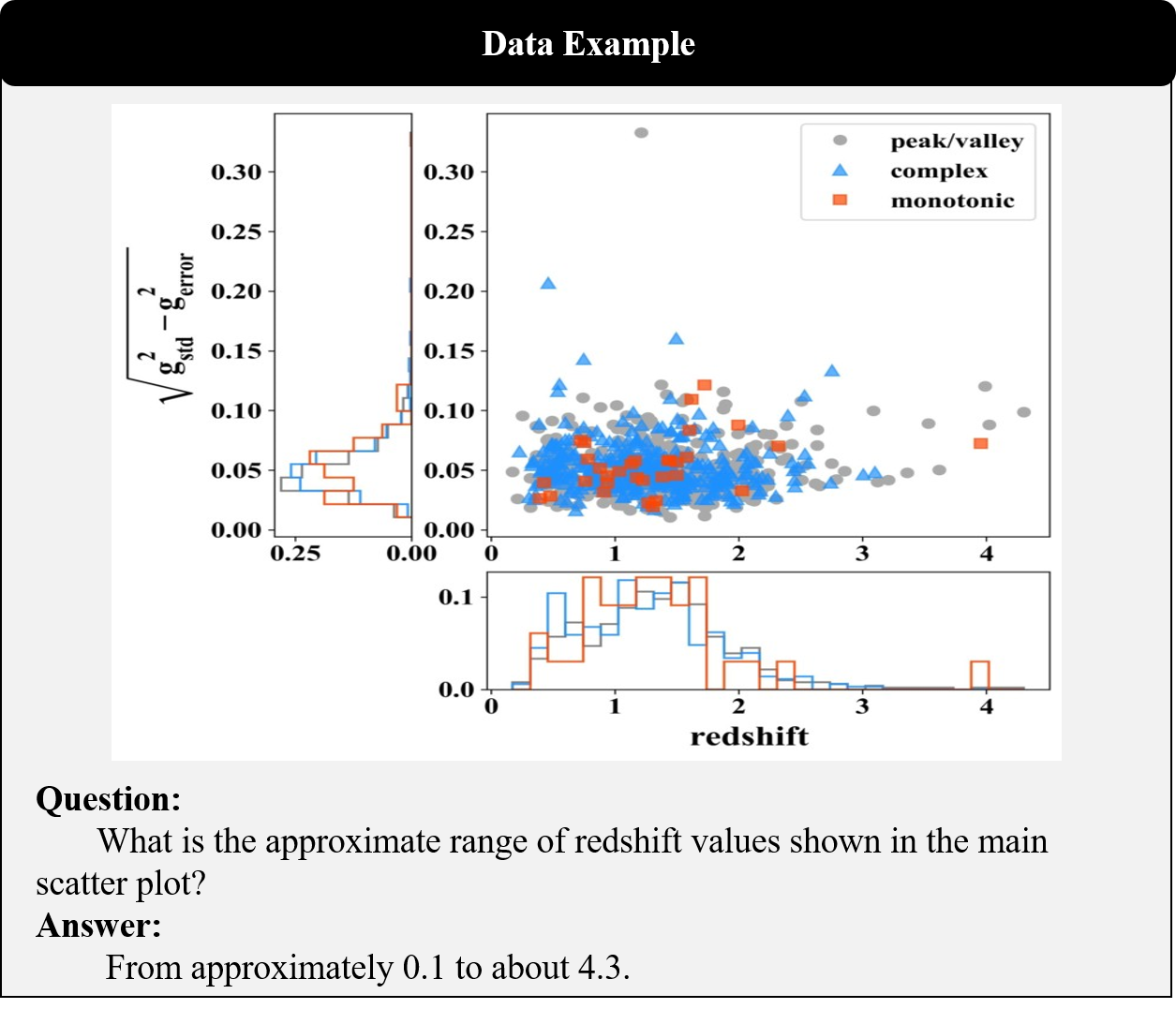}
    \caption{Example for data question-answer pair.}
    \label{fig:data_inter_2}
\end{figure}

\begin{figure}[t]
    \centering
    \includegraphics[width=\columnwidth]{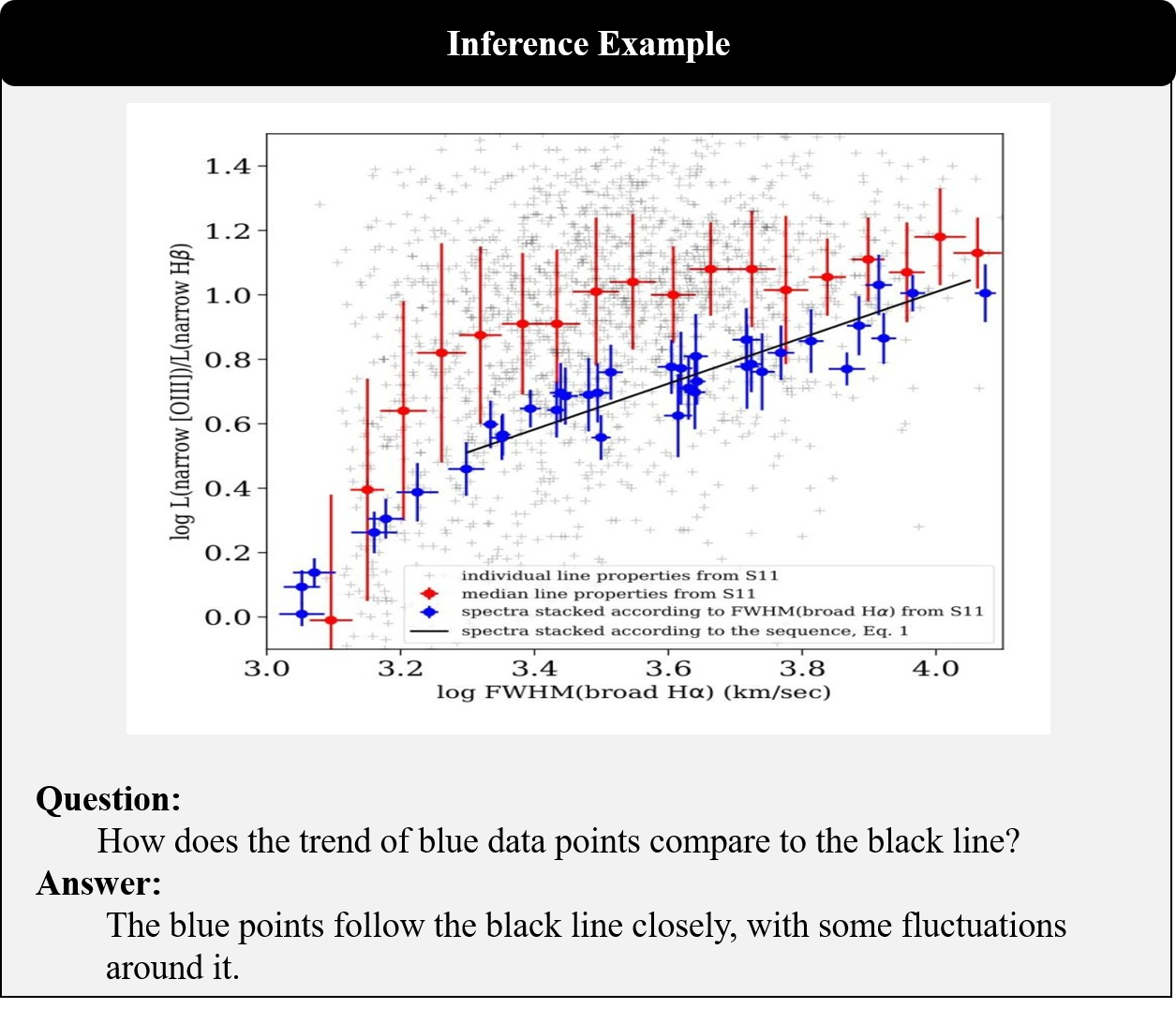}
    \caption{Example for inference question-answer pair.}
    \label{fig:inter_1}
\end{figure}

\begin{figure}[t]
    \centering
    \includegraphics[width=\columnwidth]{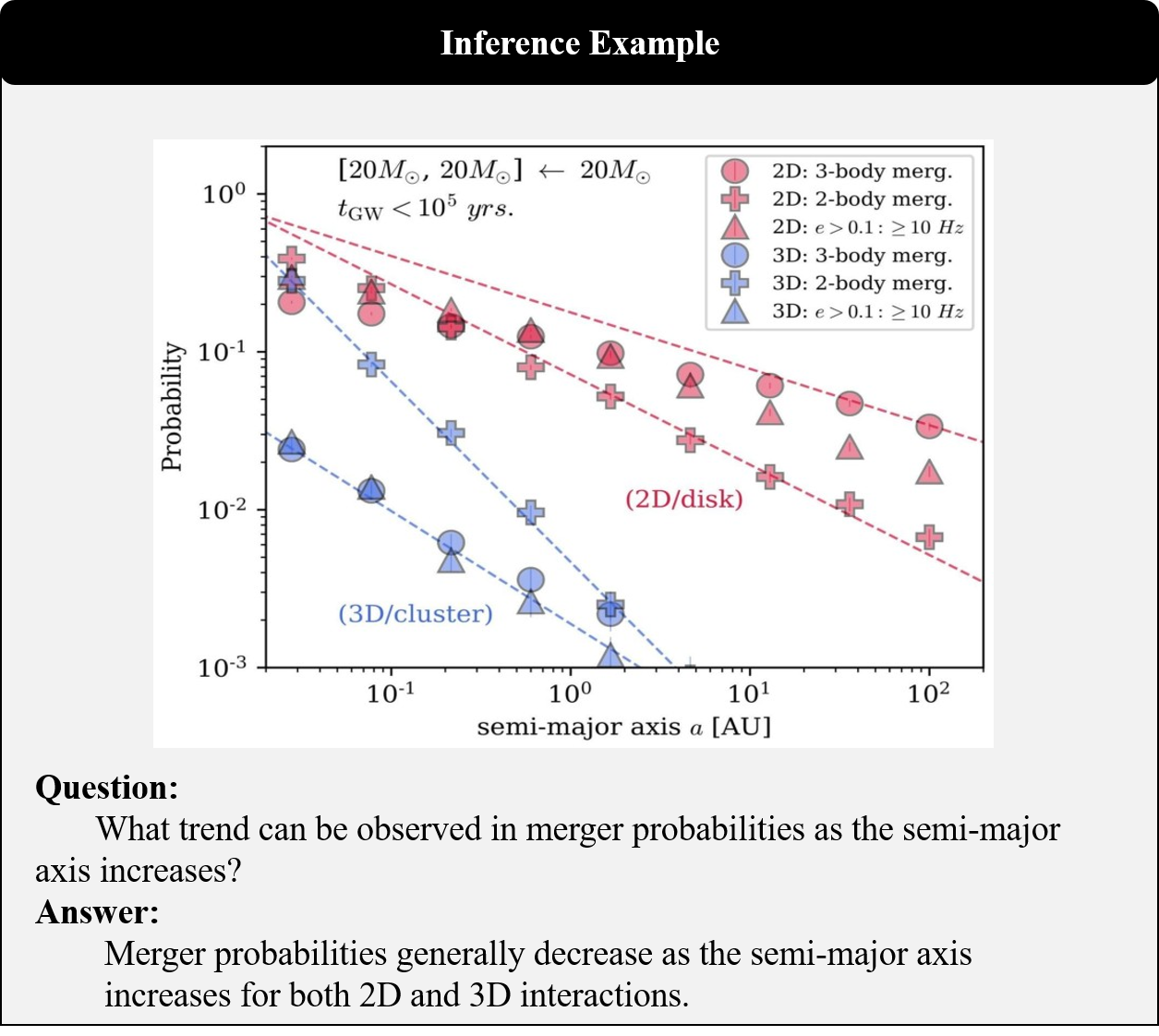}
    \caption{Example for inference question-answer pair.}
    \label{fig:inter_2}
\end{figure}

\begin{figure}[t]
    \centering
    \includegraphics[width=\columnwidth]{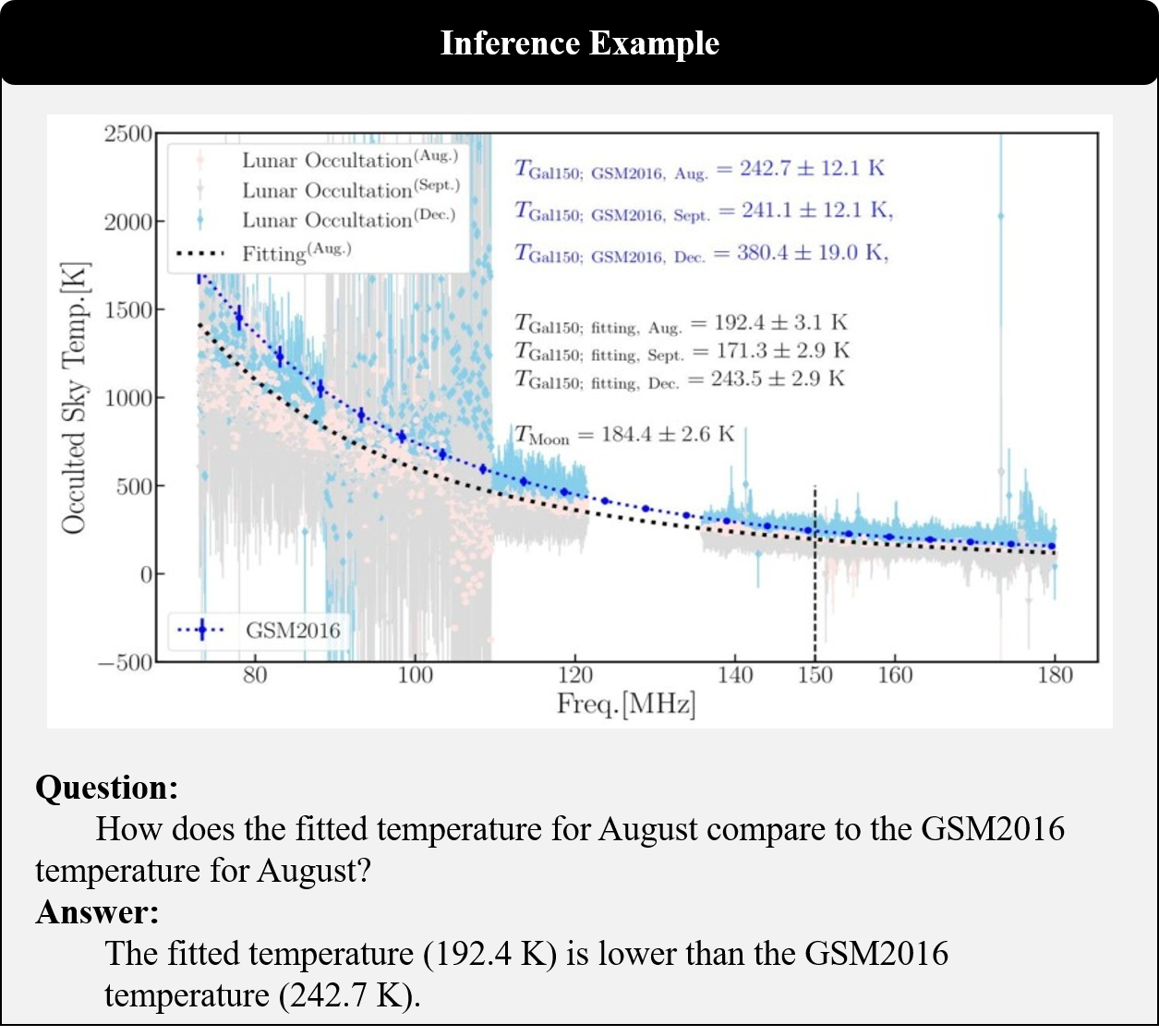}
    \caption{Example for inference question-answer pair.}
    \label{fig:inter_3}
\end{figure}

\begin{figure}[t]
    \centering
    \includegraphics[width=\columnwidth]{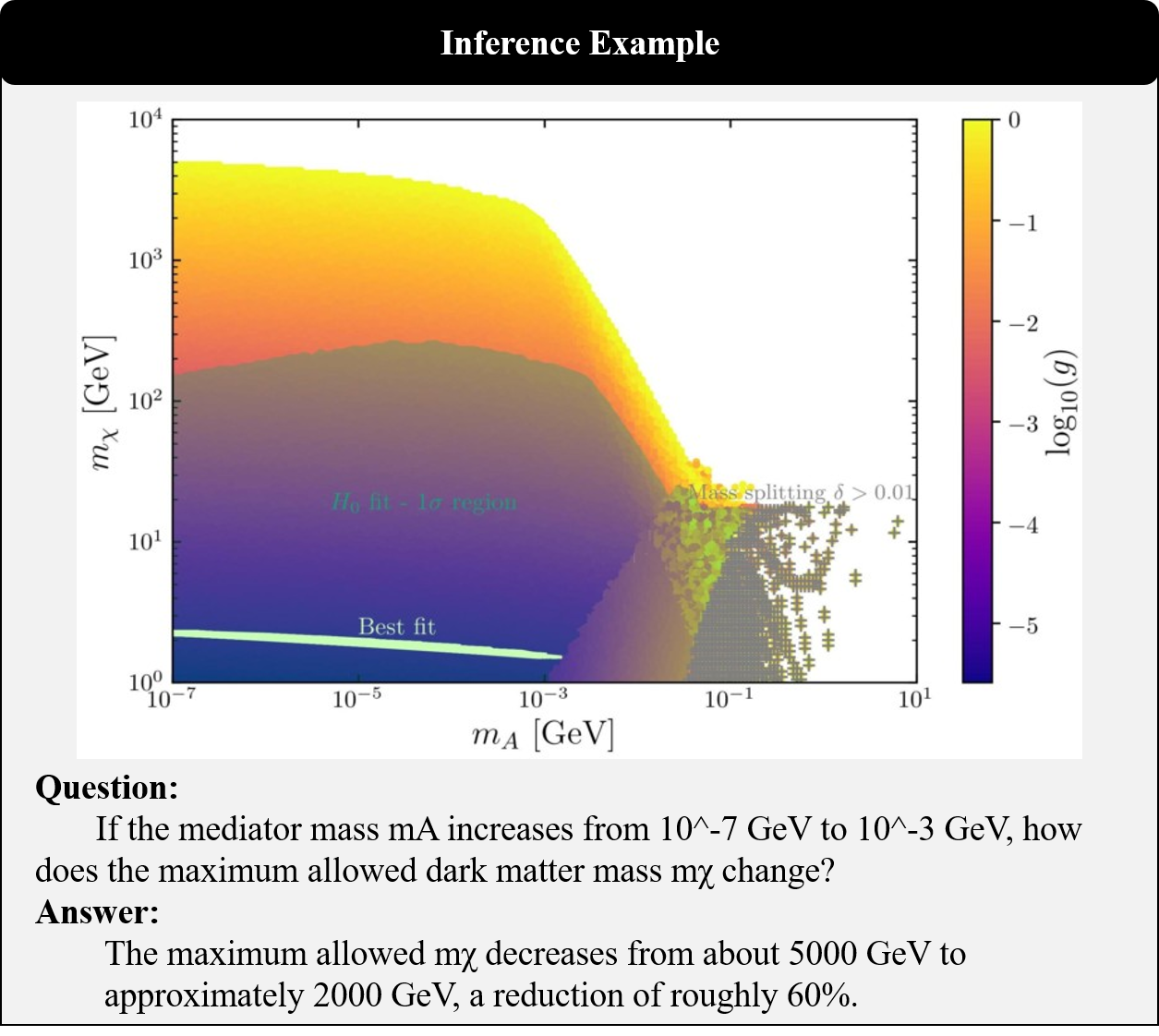}
    \caption{Example for inference question-answer pair.}
    \label{fig:inter_4}
\end{figure}

\begin{figure}[t]
    \centering
    \includegraphics[width=\columnwidth]{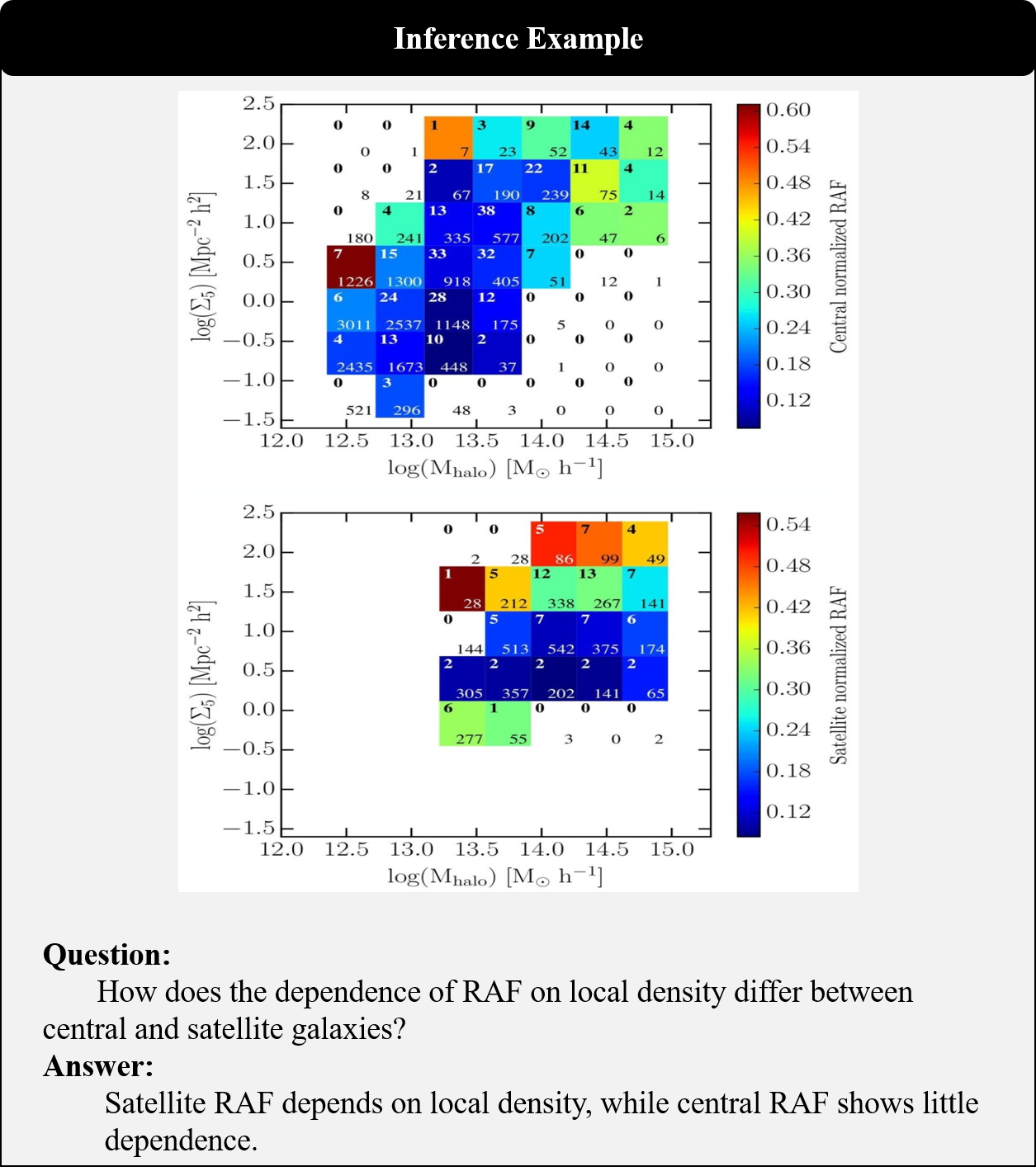}
    \caption{Example for inference question-answer pair.}
    \label{fig:inter_5}
\end{figure}

\begin{figure}[t]
    \centering
    \includegraphics[width=\columnwidth]{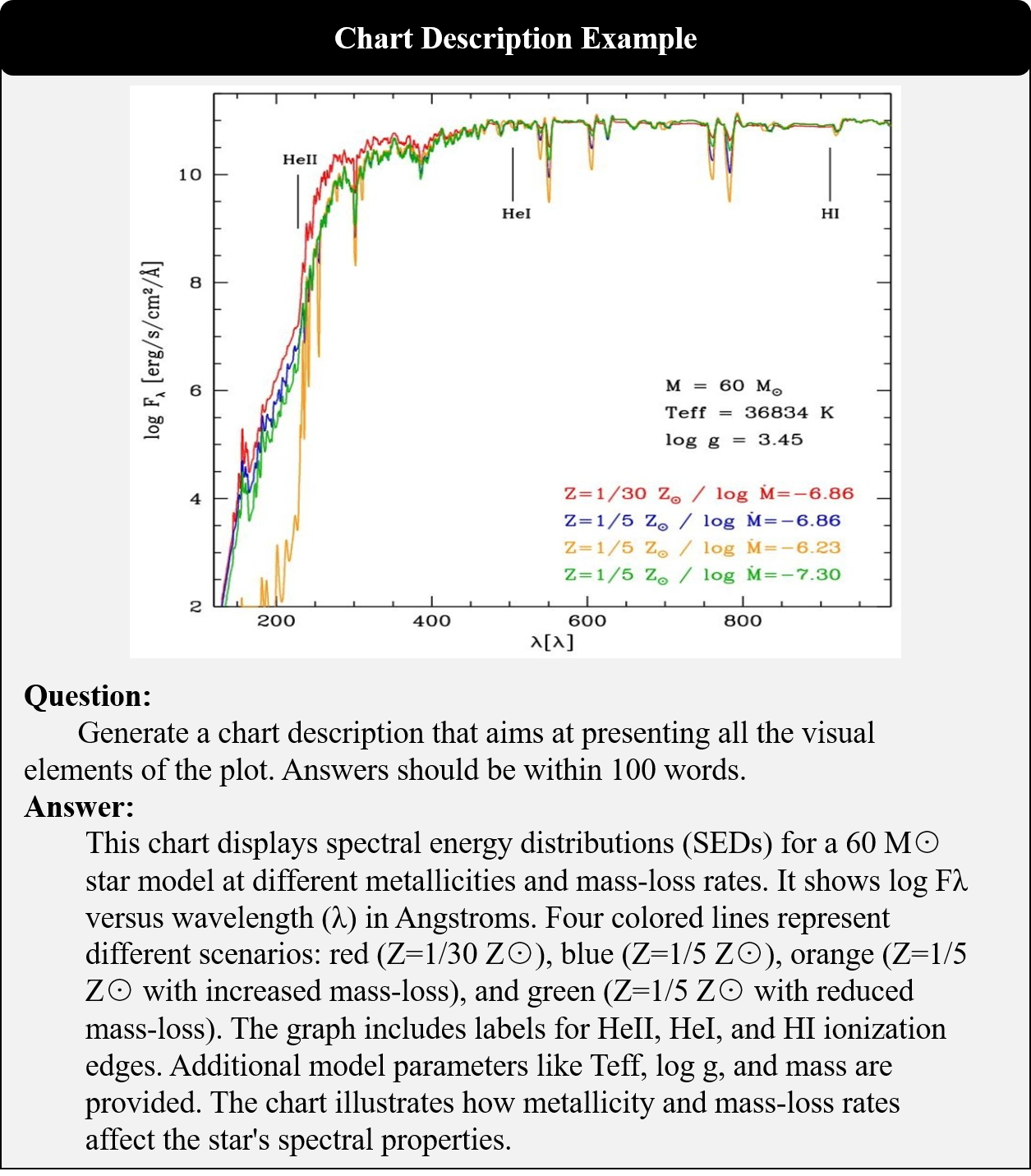}
    \caption{Example for chart description question-answer pair.}
    \label{fig:summ_1}
\end{figure}

\begin{figure}[t]
    \centering
    \includegraphics[width=\columnwidth]{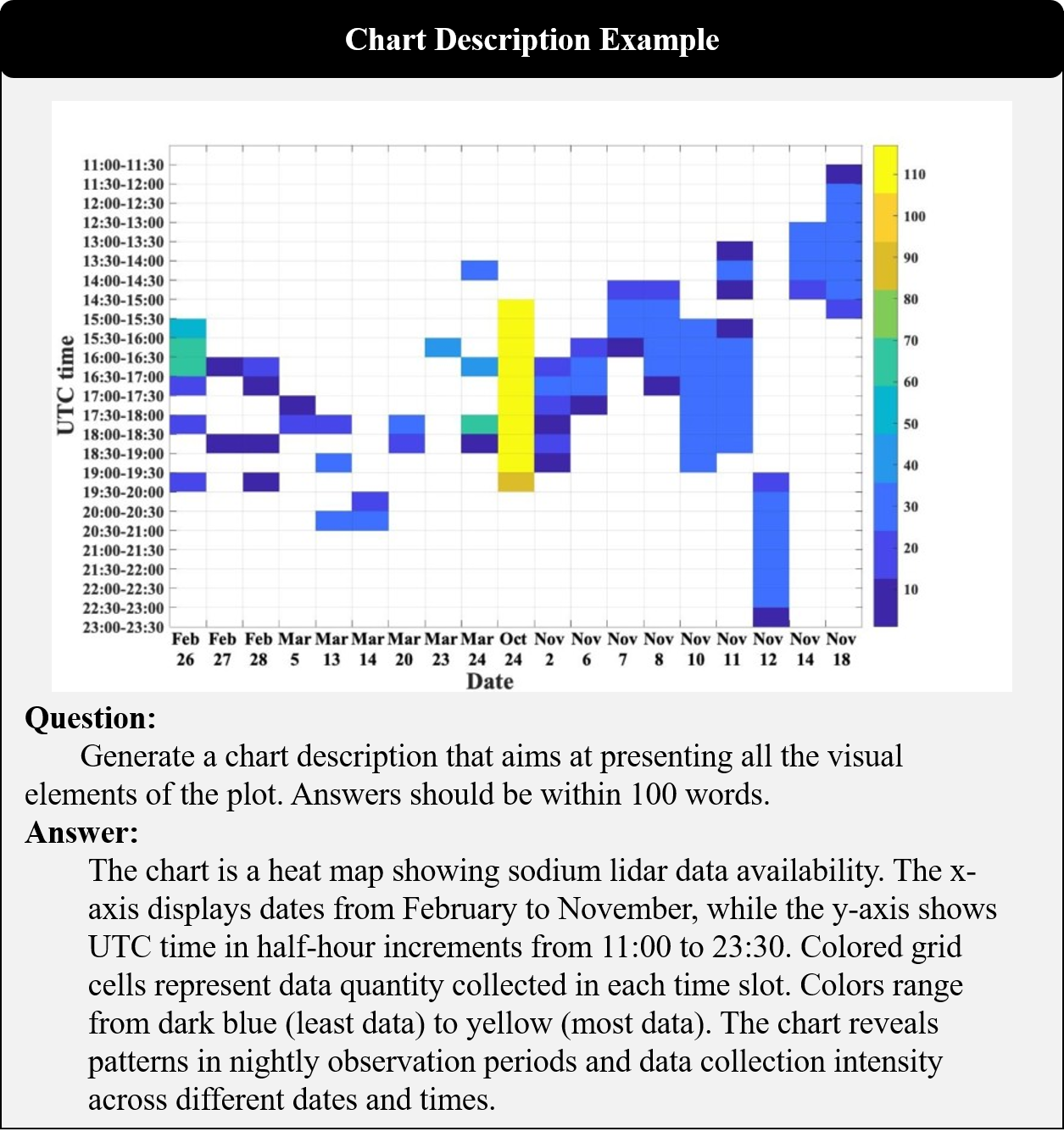}
    \caption{Example for chart description question-answer pair.}
    \label{fig:summ_2}
\end{figure}

\begin{figure}[t]
    \centering
    \includegraphics[width=\columnwidth]{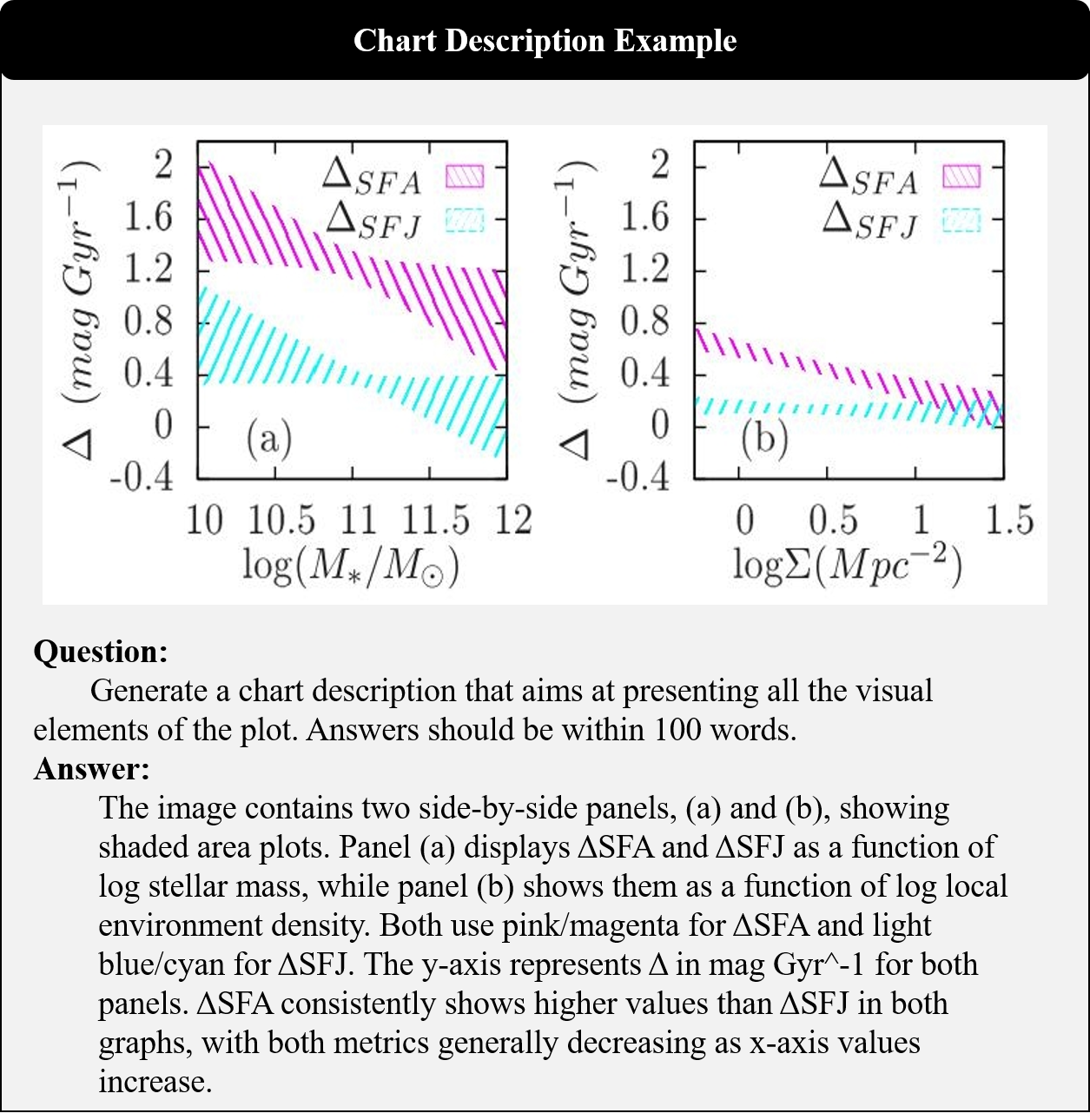}
    \caption{Example for chart description question-answer pair.}
    \label{fig:summ_3}
\end{figure}

\begin{figure}[t]
    \centering
    \includegraphics[width=\columnwidth]{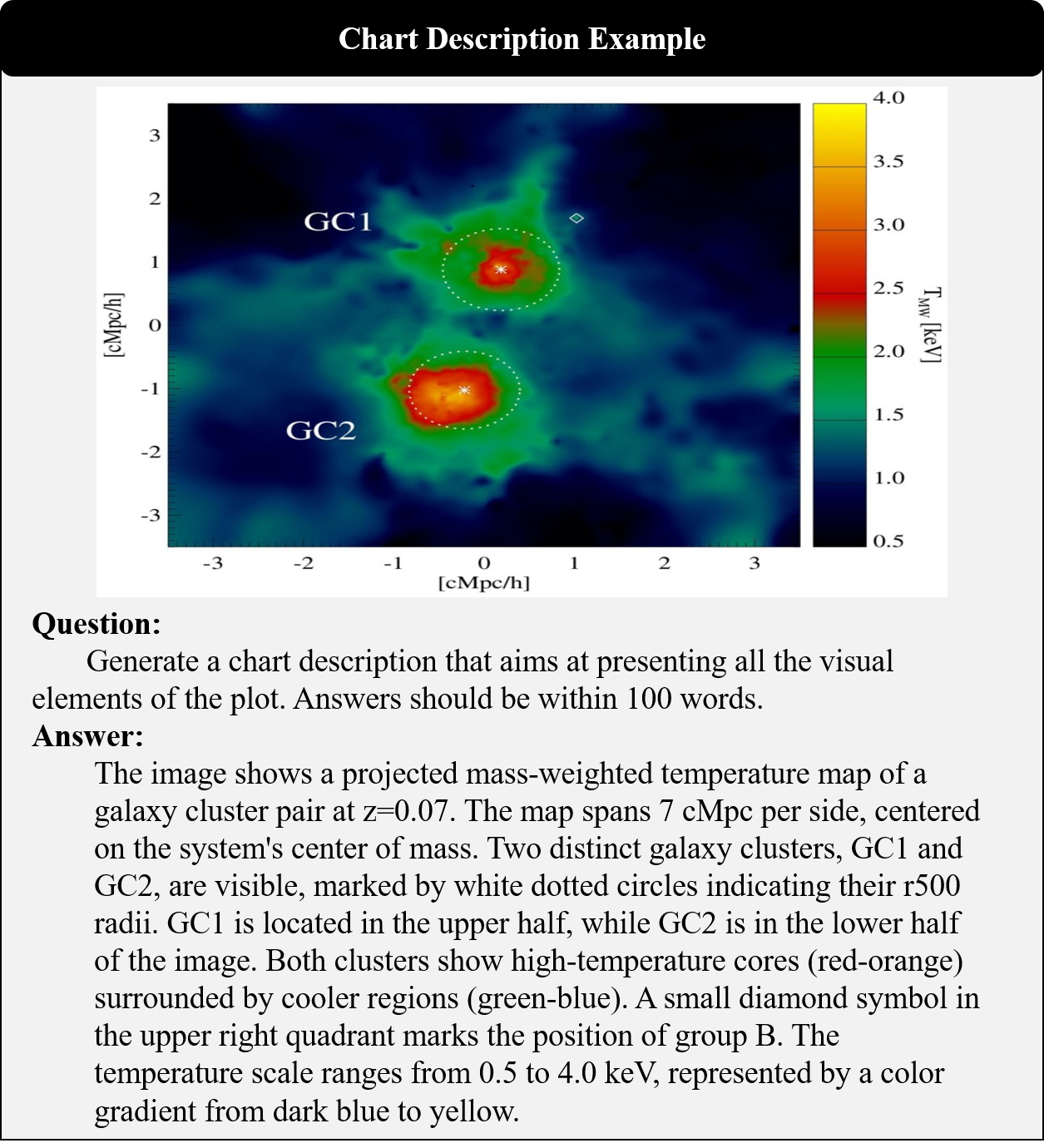}
    \caption{Example for chart description question-answer pair.}
    \label{fig:summ_4}
\end{figure}

\begin{figure}[t]
    \centering
    \includegraphics[width=\columnwidth]{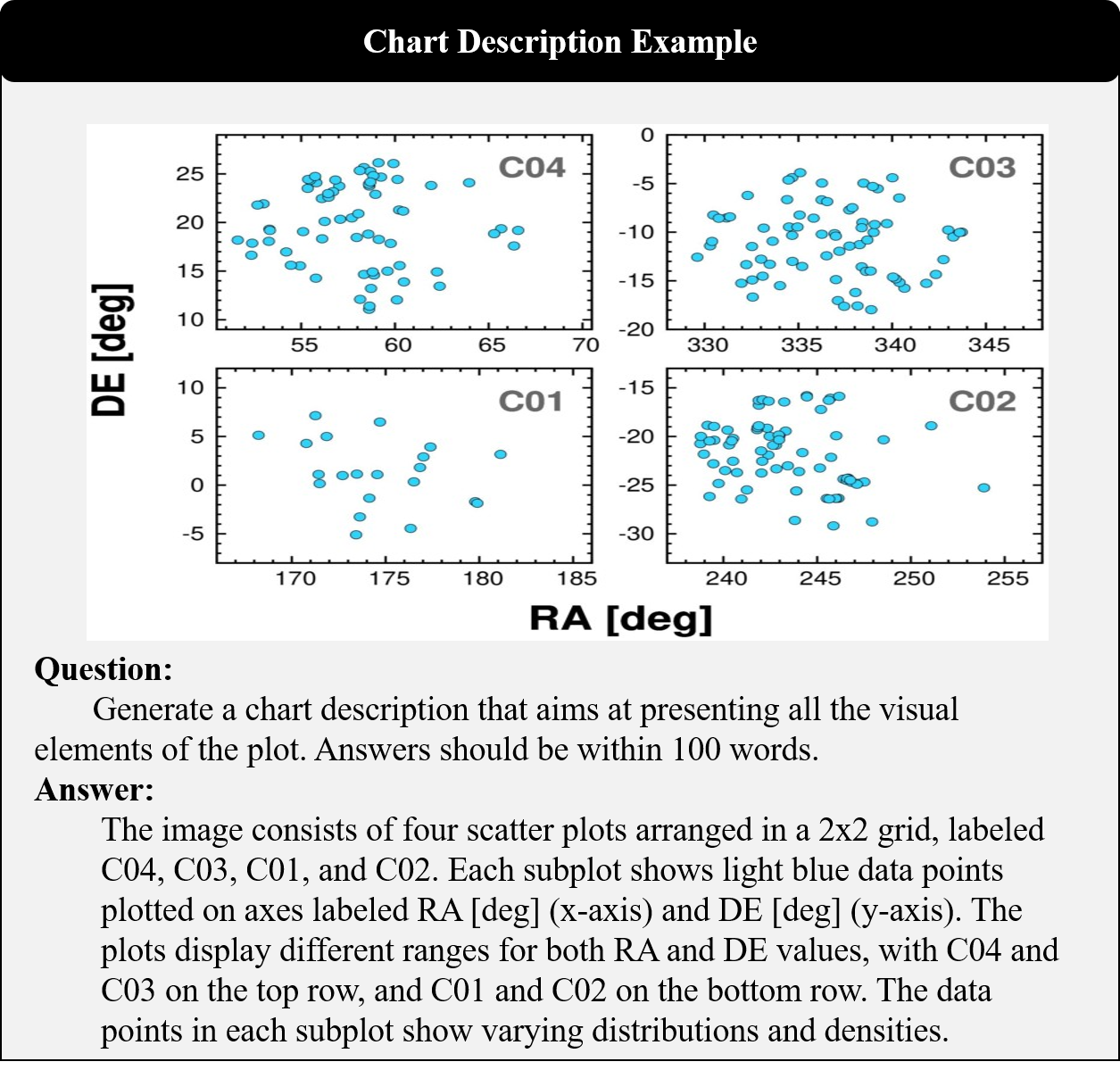}
    \caption{Example for chart description question-answer pair.}
    \label{fig:summ_5}
\end{figure}

We generated a total of 1890 question-answer pairs, consisting of 1,509 FQA pairs and 381 AQA pairs. The FQA pairs are further divided into four subcategories:

\begin{itemize}
    \item \textbf{Visual question-answer pairs}: 603 in total, covering four Types—Text (\cref{fig:visual_text}), Color (\cref{fig:visual_color}), Style (\cref{fig:visual_style}), and Layout (\cref{fig:visual_layout}, \cref{fig:visual_layout_2}).
    \item \textbf{Data question-answer pairs}: 315 in total, categorized into Calculation (\cref{fig:data_cal_1}, \cref{fig:data_cal_2}), Point (\cref{fig:data_point}), and Interval (\cref{fig:data_inter}, \cref{fig:data_inter_2}).
    \item \textbf{Inference questions-answer pair}: 289 pairs (\cref{fig:inter_1}, \cref{fig:inter_2}, \cref{fig:inter_3}, \cref{fig:inter_4}, \cref{fig:inter_5}).
    \item \textbf{Summary questions-answer pair}: 302 pairs (\cref{fig:summ_1}, \cref{fig:summ_2}, \cref{fig:summ_3}, \cref{fig:summ_4}, \cref{fig:summ_5}).
\end{itemize}

\begin{figure}[t]
    \centering
    \includegraphics[width=\columnwidth]{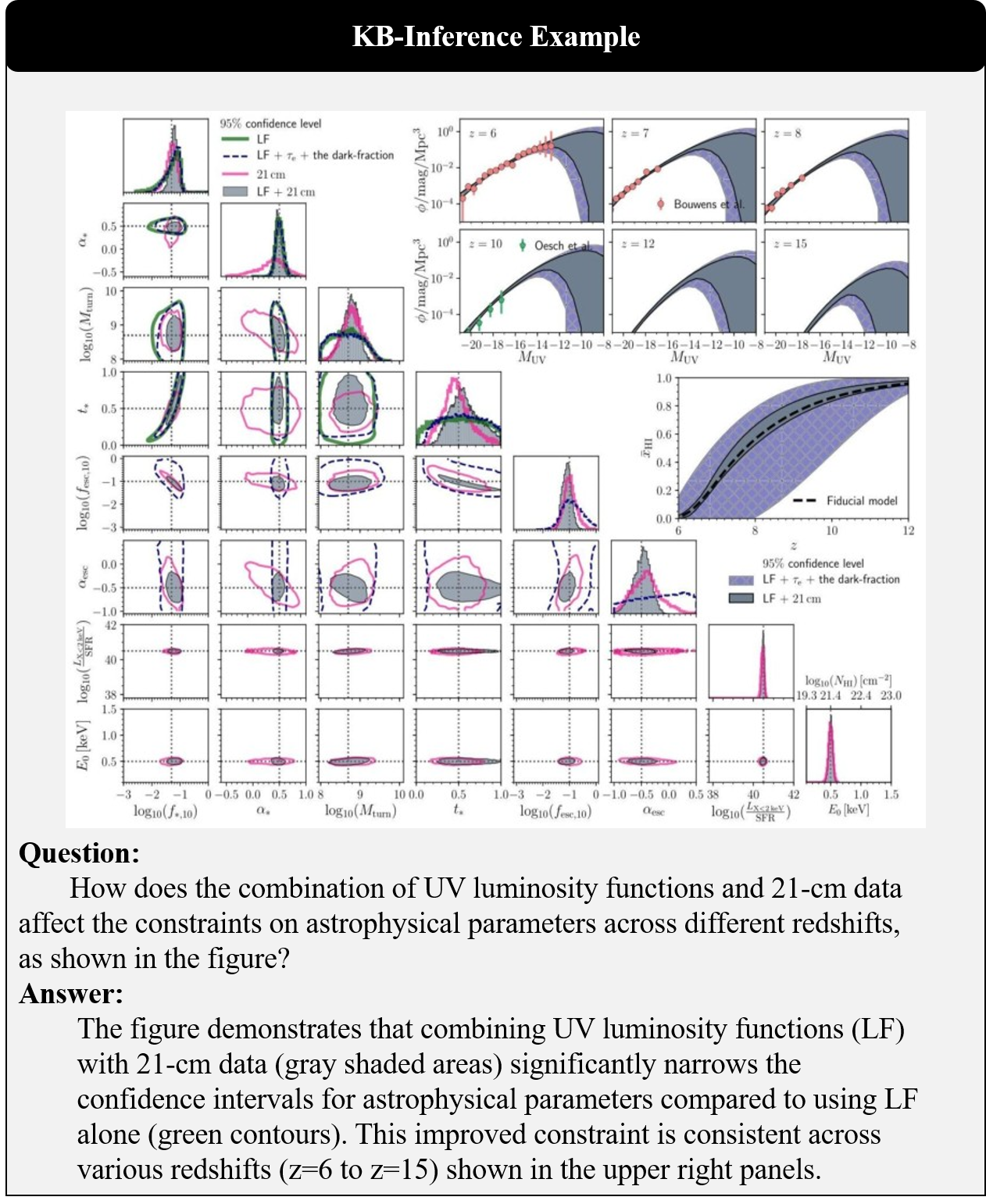}
    \caption{Example for KB-inference question-answer pair.}
    \label{fig:kb-in_1}
\end{figure}

\begin{figure}[t]
    \centering
    \includegraphics[width=\columnwidth]{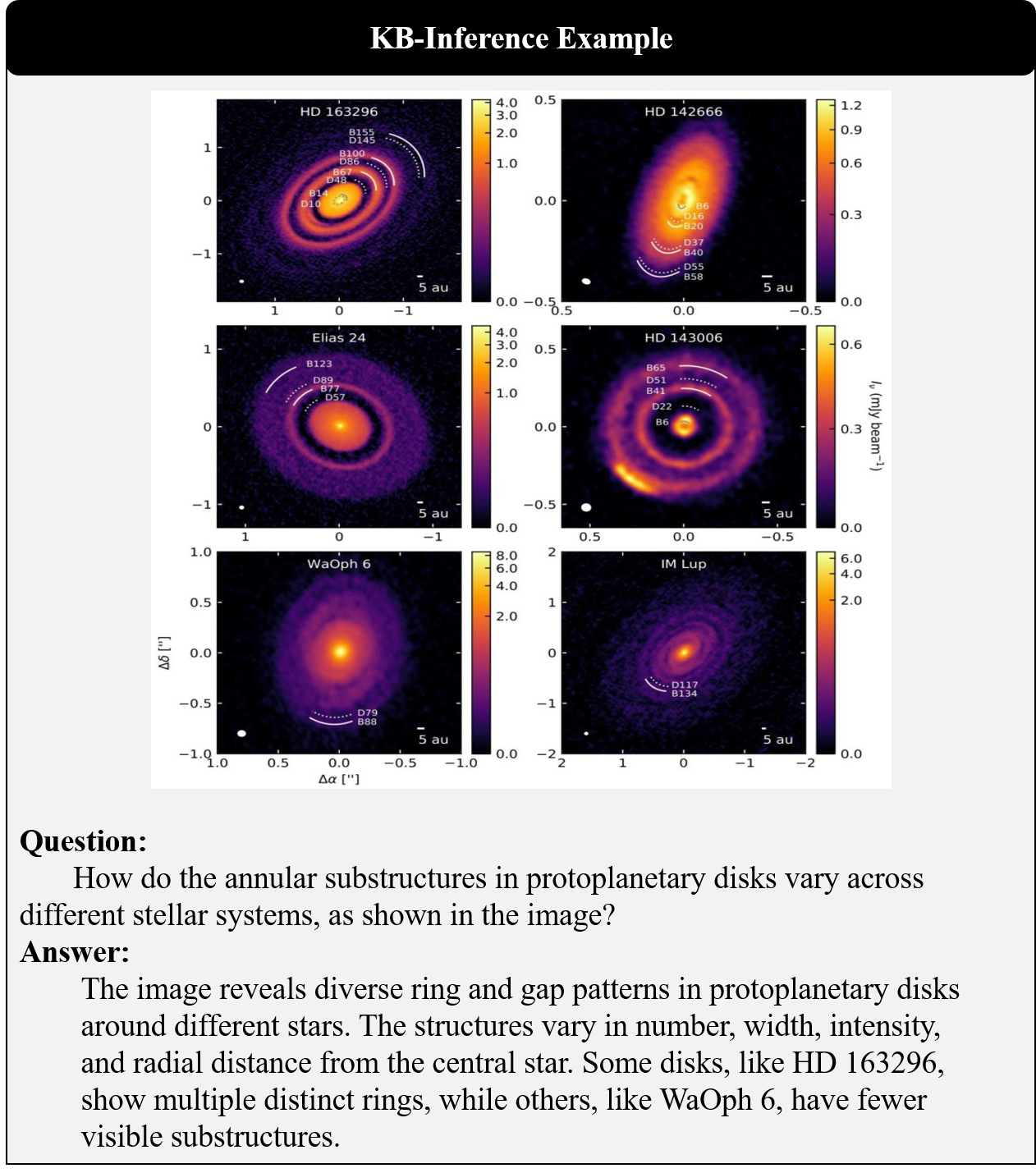}
    \caption{Example for KB-inference question-answer pair.}
    \label{fig:kb-in_2}
\end{figure}

\begin{figure}[t]
    \centering
    \includegraphics[width=\columnwidth]{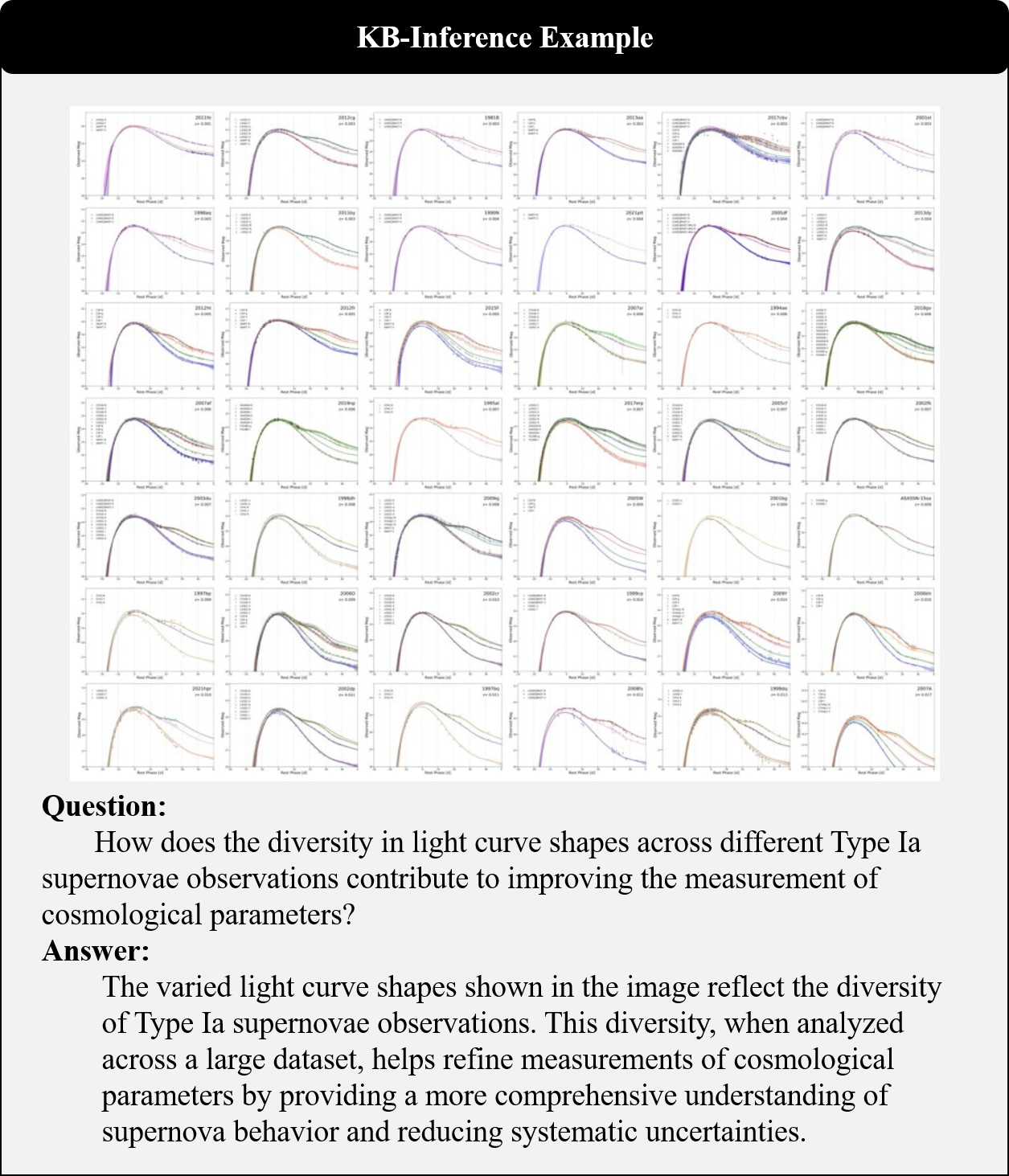}
    \caption{Example for KB-inference question-answer pair.}
    \label{fig:kb-in_3}
\end{figure}

\begin{figure}[t]
    \centering
    \includegraphics[width=\columnwidth]{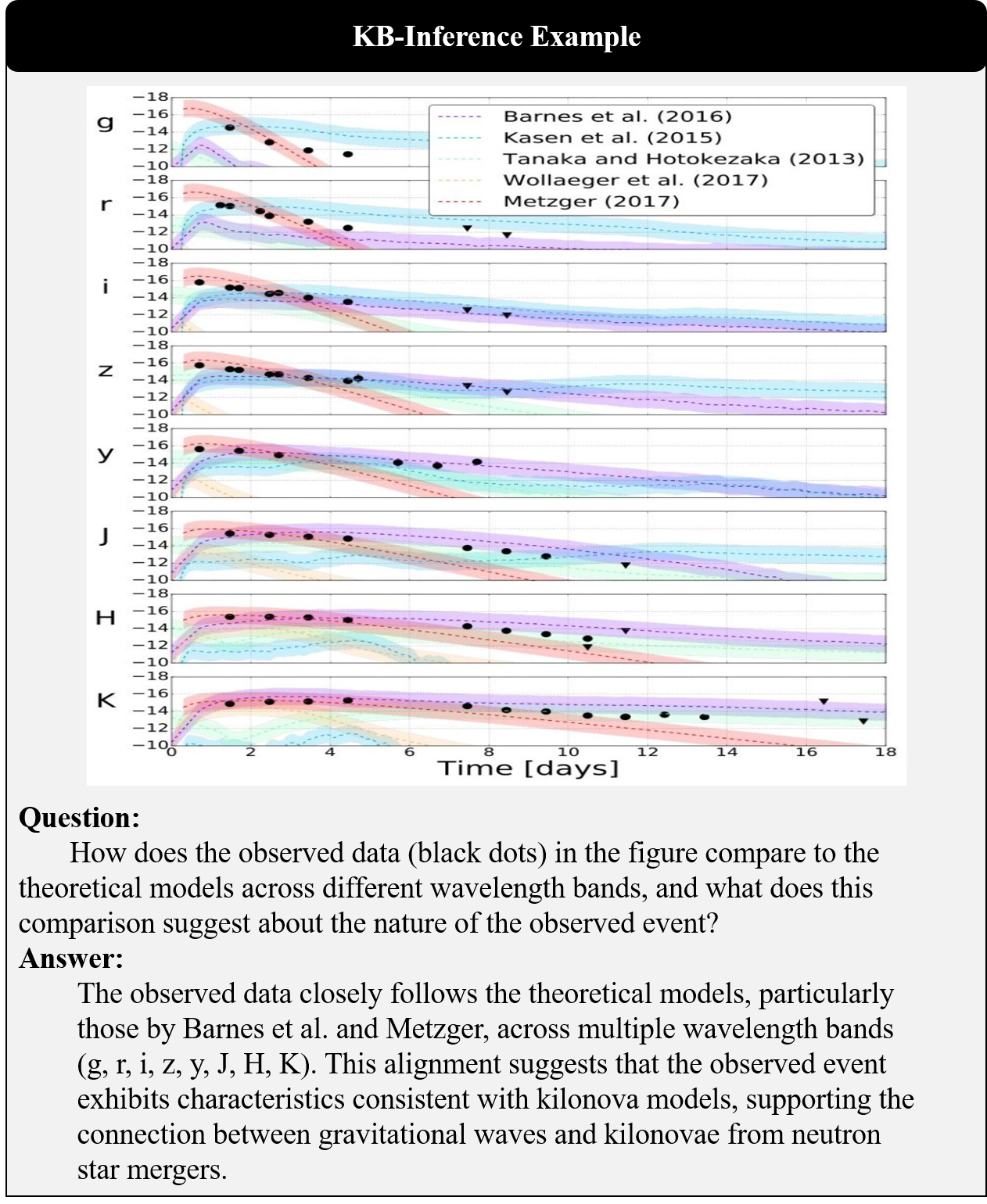}
    \caption{Example for KB-inference question-answer pair.}
    \label{fig:kb-in_4}
\end{figure}

\begin{figure}[t]
    \centering
    \includegraphics[width=\columnwidth]{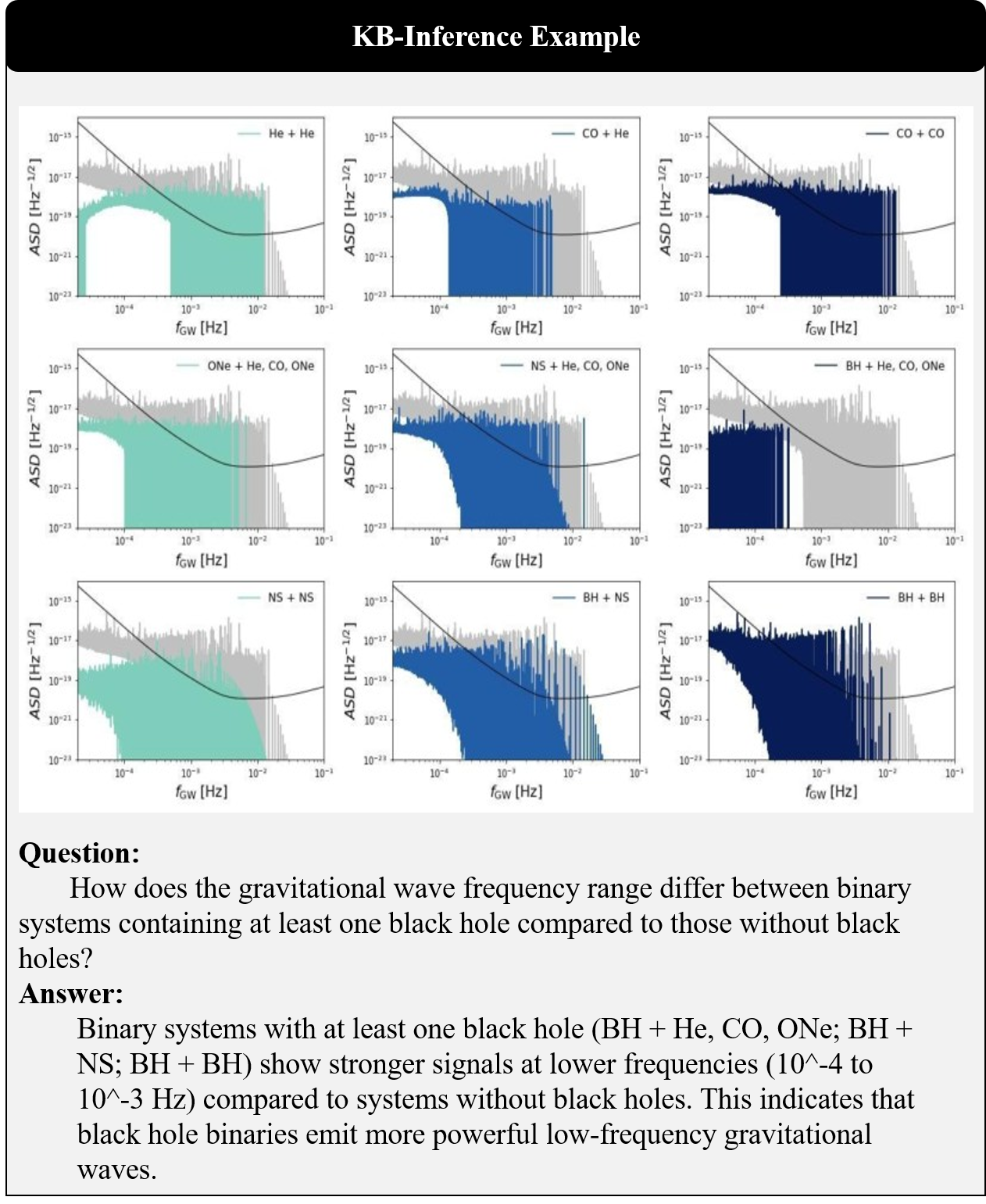}
    \caption{Example for KB-inference question-answer pair.}
    \label{fig:kb-in_5}
\end{figure}

The AQA pairs are divided into two types:
\begin{itemize}
    \item \textbf{KB-Inference questions-answer pair}: 200 pairs (\cref{fig:kb-in_1}, \cref{fig:kb-in_2}, \cref{fig:kb-in_3}, \cref{fig:kb-in_4}, \cref{fig:kb-in_5}).
\end{itemize}

\clearpage





\clearpage
\section{J. Expert Proofreading Website Screenshot}
\label{app:website_screenshot}

\begin{figure}[htbp]
    \centering
    \includegraphics[width=1\linewidth]{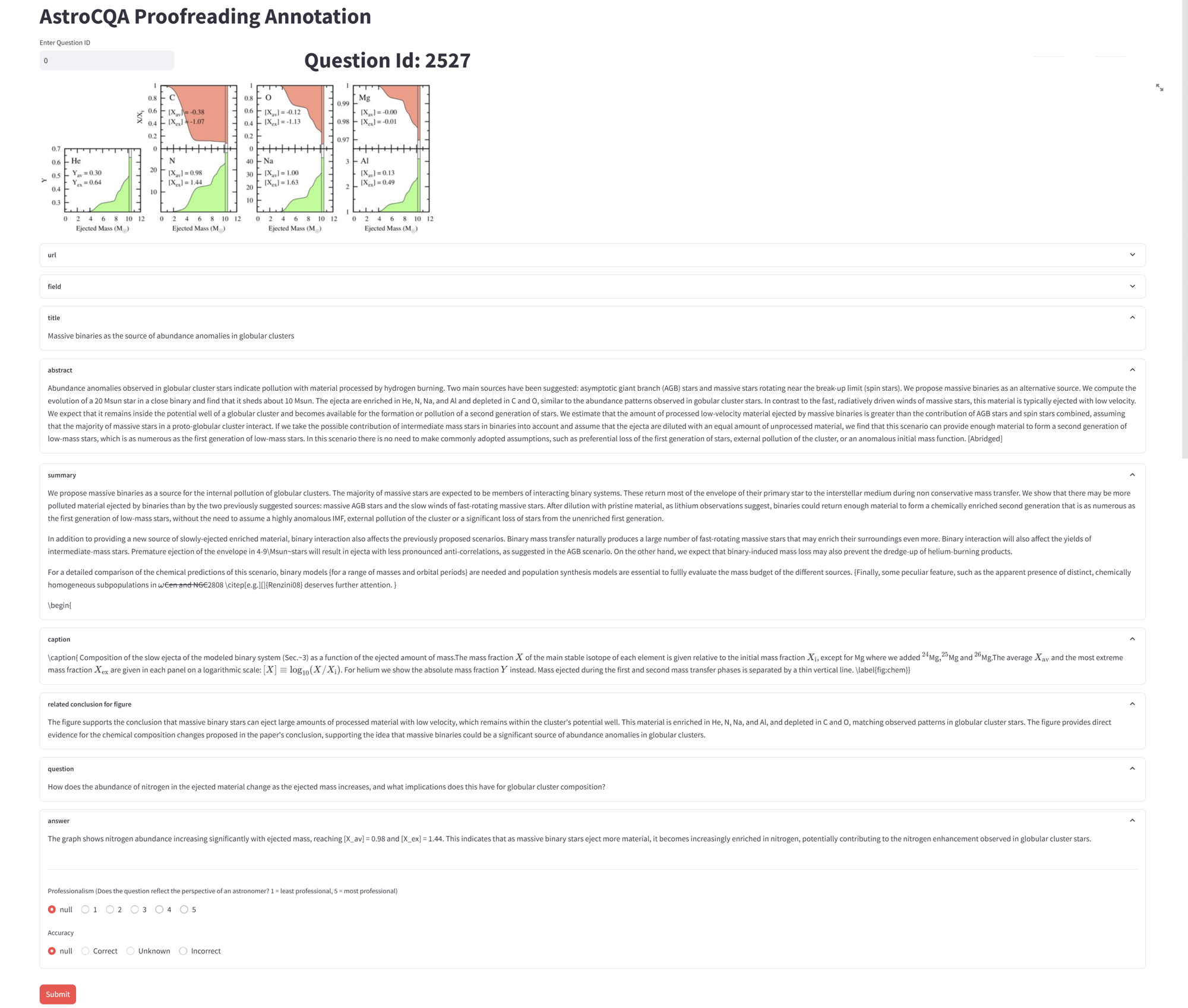}
    \caption{Screenshot of the expert proofreading website for QA validation, where validators review question-answer pairs with research excerpts, rate professionalism, and accuracy, and provide corrections or comments.}
    \label{fig:web_demo}
\end{figure}

We developed a website(\cref{fig:web_demo}) for question-answer pair validation. Validators can log into the website to review the question-answer pairs along with the corresponding research paper excerpts. They can assess the professionalism and accuracy of the pairs by providing scores. If any errors are found, validators can input the correct information as a comment.

\clearpage
\section{K. Details of Evaluation Metrics}
\label{app:details_evaluation_metrics}

In this appendix, we detail the evaluation framework used to assess both numerical responses and open-ended responses, as described in the main text.

\subsection{K.1. Evaluation of Numerical Responses}

For numerical responses, we categorize evaluation into data retrieval and data derivation. Data retrieval focuses on extracting specific data points or value ranges from charts. Data derivation involves structural element prediction (e.g., number of bars, colors, legends) and math reasoning.

\textbf{Data Retrieval Evaluation.} To ensure a scale-aware evaluation, we normalize the relative error using the axis range. The scoring process follows \cref{proc:numerical_scoring}.

\begin{algorithm}
\caption{Numerical Value Extraction and Scoring}
\label{proc:numerical_scoring}
\begin{algorithmic}[1]
\Require Reference values, Predicted values
\Ensure Final Score as $S_{final}$
\Procedure{ScoreValues}{Reference, Prediction}
    \State Extract numerical values from both Reference and Prediction
    \If{Number of reference values $>$ Number of predicted values}
        \State \Return $S_{final}=0$
    \ElsIf{Number of predicted values $>$ Number of reference values}
        \State Compute the mean of predicted values
    \EndIf
    \State Construct pairs \( V_i = \{ (\mathrm{Predict}_i, \mathrm{True}_i) \} \)

    \If{Chart axis is logarithmic}
        \State Apply logarithmic transformation (or retain exponent)
    \EndIf

    \State Initialize $S_{final}=0$
    \For{each pair \( V_i \)}
        \State Compute relative error $R_i$:
        \[
        R_i = \frac{|\mathrm{True}_i - \mathrm{Predict}_i|}{D_{range}}
        \]
        where $D_{range}$ is length of the axis.
        \State Compute Score $S_i$:
        \[
        S_i = (1 - R_i) \times I\left((1 - R_i) > 0.9 \right)
        \]
        \State Accumulate Score
    \EndFor

    \State Compute Final Score:
    \[
    S_{final} = \frac{1}{N} \sum_{i=1}^{N} \mathrm{Score}_i
    \]
    \State \Return $S_{final}$
\EndProcedure
\end{algorithmic}
\end{algorithm}

\textbf{Data Derivation Evaluation}. For data derivation, an LLM extracts numerical values, and correctness is determined by exact numerical matching, ensuring that only fully correct answers are considered accurate.

\subsection{K.2. Evaluation of Open-ended Responses}
\begin{figure}[htbp]
    \centering
    \includegraphics[width=0.80\linewidth]{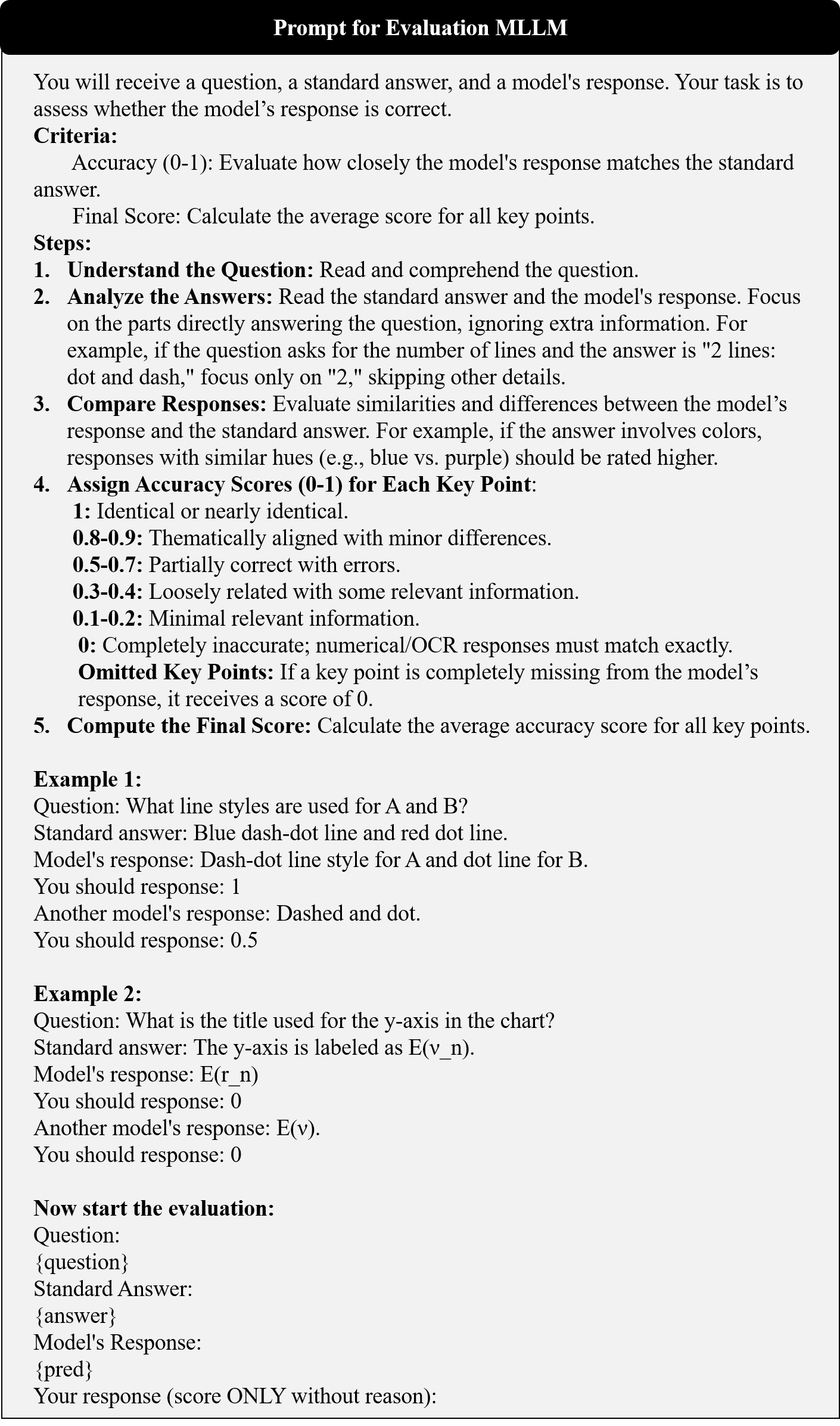}
    \caption{Prompt for evaluation MLLMs.}
    \label{fig:evaluation_prompt}
\end{figure}

We employ an LLM-based judging framework to evaluate open-ended responses. A dedicated judging model assigns a score between 0 and 1 based on predefined criteria, ensuring consistency and scalability. Our approach first extracts key points from both the generated and reference answers, then performs fine-grained matching to assess correctness. The final score is computed using an averaging strategy, providing a more nuanced evaluation. The evaluation prompt design is illustrated in \cref{fig:evaluation_prompt}.

\clearpage
\vspace{-55pt}
\section{Other Evaluation Metrics and Results}
\label{app:other_evaluation_results}

We further evaluate the models using L3Score(\cref{tab:l3score}), BLEU-4(\cref{tab:bleu-4}), and ROUGE-L(\cref{tab:Rouge-L}), which are commonly used metrics for assessing text generation quality. L3Score captures semantic alignment in long-form answers, BLEU-4 evaluates n-gram precision, and ROUGE-L measures lexical overlap. In the following tables, bold numbers indicate the best-performing model among proprietary and open-source MLLMs, respectively.

\begin{table}[http]
\centering
\setlength{\tabcolsep}{2.5pt} 
\scriptsize  
\renewcommand{\arraystretch}{1.25}
\begin{tabular}{lccccccccccccc}
\hline
\multicolumn{1}{c}{}                                 & \multicolumn{11}{c}{\textbf{FQA}}& \textbf{AQA} & \multicolumn{1}{c}{}\\ 
\cmidrule(lr){2-12} \cmidrule(lr){13-13}
\multicolumn{1}{c}{} & \multicolumn{5}{c}{\textbf{Visual}}         & \multicolumn{4}{c}{\textbf{Data}} & \multicolumn{1}{c}{} & \multicolumn{1}{c}{} & \multicolumn{1}{c}{}                                        & \multicolumn{1}{c}{}                                   \\ \cmidrule(lr){2-6} \cmidrule(lr){7-10}
\multirow{-3}{*}{\textbf{Model}} & \textbf{All}     & \textbf{color}   & \textbf{style}   & \textbf{text}    & \textbf{layout}  & \textbf{All}     & \textbf{point}   & \textbf{interval} & \textbf{calculation} & \multicolumn{1}{c}{\multirow{-2}{*}{\textbf{Inference}}} & \multicolumn{1}{c}{\multirow{-2}{*}{\textbf{Chart Desc.}}} & \multicolumn{1}{c}{\multirow{-2}{*}{\textbf{KB-Infer.}}} & \multicolumn{1}{c}{\multirow{-3}{*}{\textbf{Overall}}} \\ \hline

\multicolumn{14}{c}{\textbf{Proprietary Multimodal Large Language Models}} \\
\hline
\textbf{Gemini-2.5-pro}         & \cellcolor[HTML]{E7E6E6}89.18      & 85.97        & 85.82        & 93.76       & 92.50         & \cellcolor[HTML]{E7E6E6}72.15      & 78.22        & 71.72           & 63.29              & 80.24                                & 94.94                    & 75.82                                   & 84.07          \\
\textbf{Gemini-2.5-flash}       & \cellcolor[HTML]{E7E6E6}86.06      & 83.69        & 86.46        & 89.30       & 82.03         & \cellcolor[HTML]{E7E6E6}64.88      & 65.32        & 63.86           & 65.45              & 80.86                                & 95.18                    & 73.41                                   & 81.49          \\
\textbf{GPT-4o}                 & \cellcolor[HTML]{E7E6E6}85.90      & 85.48        & 82.53        & 87.11       & 91.12         & \cellcolor[HTML]{E7E6E6}52.03      & 50.72        & 59.86           & 44.57              & 73.80                                & 93.16                    & 71.84                                   & 77.29          \\
\textbf{Qwen-VL-Max}            & \cellcolor[HTML]{E7E6E6}82.29      & 83.97        & 69.84        & 86.41       & 89.35         & \cellcolor[HTML]{E7E6E6}51.76      & 50.74        & 58.87           & 44.71              & 71.17                                & 87.71                    & 66.84                                   & 74.00          \\

\hline
\multicolumn{14}{c}{\textbf{Open-source Multimodal Large Language Models}} \\
\hline   
\textbf{TinyChart-3B}           & \cellcolor[HTML]{E7E6E6}24.99      & 38.65        & 20.89        & 14.84       & 18.45         & \cellcolor[HTML]{E7E6E6}12.61      & 18.47        & 9.38            & 7.44               & 14.20                                & 0.33                     & 3.73                                    & 14.15          \\
\textbf{Llava1.5-7B}            & \cellcolor[HTML]{E7E6E6}23.02      & 33.48        & 19.12        & 15.20       & 20.20         & \cellcolor[HTML]{E7E6E6}8.06       & 8.11         & 8.47            & 7.48               & 34.75                                & 0.00                     & 26.17                                   & 18.45          \\
\textbf{Llava1.6-mistral-7B}    & \cellcolor[HTML]{E7E6E6}37.54      & 46.12        & 30.67        & 32.49       & 38.14         & \cellcolor[HTML]{E7E6E6}14.97      & 17.76        & 14.64           & 11.04              & 36.66                                & 1.94                     & 27.08                                   & 25.69          \\
\textbf{Qwen-VL-Chat-7B}        & \cellcolor[HTML]{E7E6E6}37.02      & 41.70        & 32.47        & 33.35       & 42.56         & \cellcolor[HTML]{E7E6E6}11.23      & 15.34        & 7.68            & 9.18               & 29.89                                & 0.82                     & 22.35                                   & 22.94          \\
\textbf{deepseek-janus-pro-7B}  & \cellcolor[HTML]{E7E6E6}62.22      & 65.42        & 61.55        & 58.04       & 67.85         & \cellcolor[HTML]{E7E6E6}31.73      & 35.11        & 39.96           & 16.49              & 49.97                                & 35.66                    & 36.16                                   & 46.89          \\
\textbf{MiniCPM-V2.6-8B} & \cellcolor[HTML]{E7E6E6}67.06      & 68.72        & 53.05        & 76.14       & 56.68         & \cellcolor[HTML]{E7E6E6}33.61      & 32.71        & 44.64           & 21.59              & 46.33                                & 46.67                    & 40.04                                   & 50.72          \\

\textbf{InternVL3‑8B}           & \cellcolor[HTML]{E7E6E6}63.19      & 65.87        & 54.05        & 64.27       & 69.46         & \cellcolor[HTML]{E7E6E6}34.92      & 37.73        & 43.80           & 19.77              & 47.68                                & 35.58                    & 43.04                                   & 48.16          \\
\textbf{mPLUG-Owl2-8.2B}        & \cellcolor[HTML]{E7E6E6}21.57      & 26.45        & 20.79        & 16.73       & 21.77         & \cellcolor[HTML]{E7E6E6}10.00      & 10.39        & 11.26           & 7.86               & 27.86                                & 0.00                     & 18.08                                   & 16.25          \\

\textbf{Pixtral-12B}            & \cellcolor[HTML]{E7E6E6}76.73      & 76.30        & 71.72        & 79.61       & 76.76         & \cellcolor[HTML]{E7E6E6}50.12      & 51.40        & 61.55           & 34.26              & 69.10                                & 87.80                    & 61.80                                   & 70.83          \\

\textbf{Llava1.6-vicuna-13B}    & \cellcolor[HTML]{E7E6E6}41.81      & 51.19        & 33.95        & 38.25       & 34.10         & \cellcolor[HTML]{E7E6E6}13.69      & 16.31        & 11.35           & 12.49              & 34.46                                & 2.13                     & 31.91                                   & 27.14          \\
\textbf{SPHINX-v2-13B}          & \cellcolor[HTML]{E7E6E6}26.14      & 34.57        & 23.65        & 21.19       & 14.77         & \cellcolor[HTML]{E7E6E6}6.83       & 12.65        & 1.47            & 4.36               & 26.11                                & 0.00                     & 21.32                                   & 17.34          \\
\textbf{Llama-4-Maverick-17B}   & \cellcolor[HTML]{E7E6E6}82.04      & 80.05        & 73.37        & 88.43       & 85.49         & \cellcolor[HTML]{E7E6E6}56.75      & 53.72        & 61.56           & 55.61              & 73.92                                & 85.07                    & 69.83                                   & 75.16          \\
\textbf{CogVLM2-19B}            & \cellcolor[HTML]{E7E6E6}61.51      & 70.16        & 45.00        & 63.81       & 58.24         & \cellcolor[HTML]{E7E6E6}26.59      & 27.63        & 35.52           & 14.16              & 43.97                                & 34.81                    & 31.60                                   & 44.01          \\
\textbf{Gemma-3-27B}            & \cellcolor[HTML]{E7E6E6}67.18      & 64.81        & 61.15        & 69.81       & 80.38         & \cellcolor[HTML]{E7E6E6}35.66      & 34.10        & 49.48           & 21.30              & 52.94                                & 53.19                    & 52.70                                   & 54.80          \\
\textbf{Llava1.6-34B}           & \cellcolor[HTML]{E7E6E6}45.17      & 52.36        & 37.01        & 43.61       & 39.01         & \cellcolor[HTML]{E7E6E6}18.19      & 16.18        & 24.85           & 13.23              & 39.38                                & 8.38                     & 42.14                                   & 32.24          \\
\textbf{Qwen2.5-VL-72B}         & \cellcolor[HTML]{E7E6E6}81.92      & 82.16        & 72.06        & 88.16       & 78.01         & \cellcolor[HTML]{E7E6E6}52.87      & 51.53        & 60.09           & 46.17              & 68.73                                & 89.09                    & 65.01                                   & 73.70          \\
\textbf{Pixtral-large-124B}     & \cellcolor[HTML]{E7E6E6}85.30      & 83.13        & 79.31        & 90.85       & 84.23         & \cellcolor[HTML]{E7E6E6}56.96      & 59.22        & 65.84           & 42.66              & 77.66                                & 91.25                    & 70.05                                   & 78.12          \\

\hline
\multicolumn{14}{c}{\textbf{Fine-tuned}} \\
\hline

\textbf{MiniCPM-V2.6-8B-fine-tuned}       & \cellcolor[HTML]{E7E6E6}74.82      & 72.47        & 70.52        & 78.00       & 81.38         & \cellcolor[HTML]{E7E6E6}37.68      & 37.70        & 50.59           & 21.97              & 52.64                                & 52.23                    & 41.20                                   & 56.45          \\

\hline
\end{tabular}

\caption{Accuracy (\%) on AstroChart benchmark using L3Score.}
\label{tab:l3score}
\end{table}

\begin{table*}[t]
\centering
\setlength{\tabcolsep}{2.5pt} 
\scriptsize  
\renewcommand{\arraystretch}{1.25}
\begin{tabular}{lccccccccccccc}
\hline
\multicolumn{1}{c}{}                                 & \multicolumn{11}{c}{\textbf{FQA}}& \textbf{AQA} & \multicolumn{1}{c}{}\\ 
\cmidrule(lr){2-12} \cmidrule(lr){13-13}
\multicolumn{1}{c}{} & \multicolumn{5}{c}{\textbf{Visual}}         & \multicolumn{4}{c}{\textbf{Data}} & \multicolumn{1}{c}{} & \multicolumn{1}{c}{} & \multicolumn{1}{c}{}                                        & \multicolumn{1}{c}{}                                   \\ \cmidrule(lr){2-6} \cmidrule(lr){7-10}
\multirow{-3}{*}{\textbf{Model}} & \textbf{All}     & \textbf{color}   & \textbf{style}   & \textbf{text}    & \textbf{layout}  & \textbf{All}     & \textbf{point}   & \textbf{interval} & \textbf{calculation} & \multicolumn{1}{c}{\multirow{-2}{*}{\textbf{Inference}}} & \multicolumn{1}{c}{\multirow{-2}{*}{\textbf{Chart Desc.}}} & \multicolumn{1}{c}{\multirow{-2}{*}{\textbf{KB-Infer.}}} & \multicolumn{1}{c}{\multirow{-3}{*}{\textbf{Overall}}} \\ \hline

\multicolumn{14}{c}{\textbf{Proprietary Multimodal Large Language Models}} \\
\hline
\textbf{Gemini-2.5-pro}         & \cellcolor[HTML]{E7E6E6}19.48      & 20.91        & 15.64        & 21.71       & 12.54         & \cellcolor[HTML]{E7E6E6}12.97      & 14.16        & 13.98           & 9.89               & 7.16                                 & 9.32                     & 9.88                                    & 13.31          \\
\textbf{Gemini-2.5-flash}       & \cellcolor[HTML]{E7E6E6}25.14      & 30.61        & 20.86        & 23.51       & 18.58         & \cellcolor[HTML]{E7E6E6}12.81      & 12.87        & 13.54           & 11.81              & 7.93                                 & 9.46                     & 10.66                                   & 15.54          \\
\textbf{GPT-4o}                 & \cellcolor[HTML]{E7E6E6}21.26      & 23.42        & 20.64        & 21.61       & 11.07         & \cellcolor[HTML]{E7E6E6}10.83      & 11.91        & 11.10           & 8.84               & 11.41                                & 12.69                    & 15.00                                   & 15.43          \\
\textbf{Qwen-VL-Max}            & \cellcolor[HTML]{E7E6E6}21.34      & 22.91        & 20.32        & 23.27       & 7.27          & \cellcolor[HTML]{E7E6E6}11.90      & 12.61        & 13.26           & 9.12               & 12.34                                & 11.83                    & 11.15                                   & 15.25          \\

\hline
\multicolumn{14}{c}{\textbf{Open-source Multimodal Large Language Models}} \\
\hline   

\textbf{TinyChart-3B}           & \cellcolor[HTML]{E7E6E6}3.00       & 6.20         & 1.81         & 1.05        & 0.77          & \cellcolor[HTML]{E7E6E6}0.61       & 0.24         & 1.30            & 0.34               & 3.99                                 & 1.61                     & 4.52                                    & 2.64           \\
\textbf{Llava1.5-7B}            & \cellcolor[HTML]{E7E6E6}2.47       & 5.18         & 0.44         & 1.39        & 0.62          & \cellcolor[HTML]{E7E6E6}0.95       & 0.86         & 1.05            & 0.96               & 15.36                                & 2.83                     & 9.00                                    & 5.15           \\
\textbf{Llava1.6-mistral-7B}    & \cellcolor[HTML]{E7E6E6}5.40       & 8.11         & 1.83         & 5.45        & 2.90          & \cellcolor[HTML]{E7E6E6}6.12       & 6.24         & 3.49            & 9.14               & 2.69                                 & 2.98                     & 3.44                                    & 4.43           \\
\textbf{Qwen-VL-Chat-7B}        & \cellcolor[HTML]{E7E6E6}2.87       & 2.35         & 0.69         & 5.02        & 1.36          & \cellcolor[HTML]{E7E6E6}0.55       & 0.37         & 1.20            & 0.04               & 14.06                                & 5.06                     & 8.39                                    & 5.33           \\
\textbf{deepseek-janus-pro-7B}  & \cellcolor[HTML]{E7E6E6}10.57      & 9.98         & 11.01        & 11.49       & 7.09          & \cellcolor[HTML]{E7E6E6}11.39      & 11.32        & 8.82            & 14.62              & 11.66                                & 10.26                    & 11.85                                   & 10.99          \\

\textbf{MiniCPM-V2.6-8B} & \cellcolor[HTML]{E7E6E6}14.09      & 14.43        & 12.10        & 17.33       & 2.40          & \cellcolor[HTML]{E7E6E6}5.67       & 5.22         & 8.12            & 3.39               & 9.23                                 & 8.25                     & 10.84                                   & 10.29          \\
\textbf{InternVL3‑8B}           & \cellcolor[HTML]{E7E6E6}16.95      & 20.52        & 13.58        & 17.36       & 8.90          & \cellcolor[HTML]{E7E6E6}11.10      & 11.21        & 12.61           & 9.11               & 9.96                                 & 7.33                     & 9.04                                    & 12.10          \\
\textbf{mPLUG-Owl2-8.2B}        & \cellcolor[HTML]{E7E6E6}2.42       & 3.56         & 0.62         & 2.11        & 3.87          & \cellcolor[HTML]{E7E6E6}0.90       & 0.59         & 1.49            & 0.66               & 15.50                                & 2.78                     & 8.65                                    & 5.10           \\

\textbf{Pixtral-12B}            & \cellcolor[HTML]{E7E6E6}14.37      & 13.30        & 13.44        & 16.49       & 11.78         & \cellcolor[HTML]{E7E6E6}8.25       & 7.75         & 8.05            & 9.25               & 7.99                                 & 10.78                    & 11.64                                   & 11.20          \\
\textbf{Llava1.6-vicuna-13B}    & \cellcolor[HTML]{E7E6E6}5.33       & 7.55         & 1.26         & 5.75        & 4.78          & \cellcolor[HTML]{E7E6E6}0.78       & 0.58         & 1.23            & 0.56               & 12.71                                & 4.61                     & 8.88                                    & 5.99           \\
\textbf{SPHINX-v2-13B}          & \cellcolor[HTML]{E7E6E6}1.52       & 2.12         & 0.39         & 1.82        & 0.58          & \cellcolor[HTML]{E7E6E6}0.06       & 0.09         & 0.07            & 0.01               & 13.17                                & 2.70                     & 9.39                                    & 4.29           \\
\textbf{Llama-4-Maverick-17B}   & \cellcolor[HTML]{E7E6E6}14.05      & 13.92        & 11.09        & 16.73       & 10.05         & \cellcolor[HTML]{E7E6E6}9.27       & 10.29        & 9.28            & 7.66               & 8.54                                 & 12.64                    & 14.10                                   & 11.97          \\
\textbf{CogVLM2-19B}            & \cellcolor[HTML]{E7E6E6}11.43      & 9.09         & 7.65         & 16.91       & 7.19          & \cellcolor[HTML]{E7E6E6}6.64       & 5.40         & 6.34            & 8.93               & 13.25                                & 8.22                     & 9.60                                    & 10.08          \\
\textbf{Gemma-3-27B}            & \cellcolor[HTML]{E7E6E6}14.49      & 14.52        & 16.05        & 15.49       & 5.67          & \cellcolor[HTML]{E7E6E6}7.19       & 7.26         & 6.23            & 8.23               & 7.85                                 & 10.07                    & 8.72                                    & 10.58          \\
\textbf{Llava1.6-34B}           & \cellcolor[HTML]{E7E6E6}5.57       & 8.27         & 1.51         & 5.69        & 4.14          & \cellcolor[HTML]{E7E6E6}1.53       & 1.06         & 2.63            & 0.91               & 13.60                                & 5.38                     & 10.49                                   & 6.68           \\
\textbf{Qwen2.5-VL-72B}         & \cellcolor[HTML]{E7E6E6}19.83      & 20.42        & 18.61        & 22.27       & 9.81          & \cellcolor[HTML]{E7E6E6}12.61      & 14.91        & 13.37           & 8.13               & 11.14                                & 12.10                    & 12.51                                   & 14.83          \\
\textbf{Pixtral-large-124B}     & \cellcolor[HTML]{E7E6E6}21.52      & 24.57        & 19.46        & 21.52       & 12.74         & \cellcolor[HTML]{E7E6E6}15.46      & 15.06        & 15.29           & 16.27              & 10.51                                & 15.01                    & 16.02                                   & 16.75          \\
\hline
\multicolumn{14}{c}{\textbf{Fine-tuned}} \\
\hline

\textbf{MiniCPM-V2.6-8B-fine-tuned}       & \cellcolor[HTML]{E7E6E6}31.09      & 33.14        & 32.58        & 31.19       & 15.03         & \cellcolor[HTML]{E7E6E6}11.52      & 12.78        & 11.52           & 9.57               & 10.02                                & 14.77                    & 14.56                                   & 19.14          \\

\hline
\end{tabular}
\caption{Accuracy (\%) on AstroChart benchmark using BLEU-4.}
\label{tab:bleu-4}
\end{table*}

\begin{table*}[t]
\centering
\setlength{\tabcolsep}{2.5pt} 
\scriptsize  
\renewcommand{\arraystretch}{1.25}
\begin{tabular}{lccccccccccccc}
\hline
\multicolumn{1}{c}{}                                 & \multicolumn{11}{c}{\textbf{FQA}}& \textbf{AQA} & \multicolumn{1}{c}{}\\ 
\cmidrule(lr){2-12} \cmidrule(lr){13-13}
\multicolumn{1}{c}{} & \multicolumn{5}{c}{\textbf{Visual}}         & \multicolumn{4}{c}{\textbf{Data}} & \multicolumn{1}{c}{} & \multicolumn{1}{c}{} & \multicolumn{1}{c}{}                                        & \multicolumn{1}{c}{}                                   \\ \cmidrule(lr){2-6} \cmidrule(lr){7-10}
\multirow{-3}{*}{\textbf{Model}} & \textbf{All}     & \textbf{color}   & \textbf{style}   & \textbf{text}    & \textbf{layout}  & \textbf{All}     & \textbf{point}   & \textbf{interval} & \textbf{calculation} & \multicolumn{1}{c}{\multirow{-2}{*}{\textbf{Inference}}} & \multicolumn{1}{c}{\multirow{-2}{*}{\textbf{Chart Desc.}}} & \multicolumn{1}{c}{\multirow{-2}{*}{\textbf{KB-Infer.}}} & \multicolumn{1}{c}{\multirow{-3}{*}{\textbf{Overall}}} \\ \hline

\multicolumn{14}{c}{\textbf{Proprietary Multimodal Large Language Models}} \\
\hline
\textbf{Gemini-2.5-pro}         & \cellcolor[HTML]{E7E6E6}41.97      & 41.04        & 39.87        & 46.33       & 32.23         & \cellcolor[HTML]{E7E6E6}38.72      & 37.21        & 44.23           & 34.36              & 31.41                                & 40.92                    & 36.73                                   & 38.81          \\
\textbf{Gemini-2.5-flash}       & \cellcolor[HTML]{E7E6E6}46.98      & 50.29        & 43.98        & 48.76       & 31.42         & \cellcolor[HTML]{E7E6E6}39.27      & 37.65        & 44.37           & 35.57              & 33.72                                & 40.36                    & 37.94                                   & 41.12          \\
\textbf{GPT-4o}                 & \cellcolor[HTML]{E7E6E6}44.65      & 44.85        & 44.57        & 48.79       & 25.72         & \cellcolor[HTML]{E7E6E6}36.90      & 36.22        & 41.72           & 32.09              & 38.27                                & 44.00                    & 42.52                                   & 41.76          \\
\textbf{Qwen-VL-Max}            & \cellcolor[HTML]{E7E6E6}44.61      & 45.75        & 44.00        & 48.35       & 23.65         & \cellcolor[HTML]{E7E6E6}37.36      & 36.18        & 42.96           & 32.36              & 37.63 \\

\hline
\multicolumn{14}{c}{\textbf{Open-source Multimodal Large Language Models}} \\
\hline   

\textbf{TinyChart-3B}           & \cellcolor[HTML]{E7E6E6}13.95      & 23.82        & 9.14         & 7.62        & 11.38         & \cellcolor[HTML]{E7E6E6}10.95      & 9.05         & 17.63           & 5.79               & 21.11                                & 20.95                    & 28.36                                   & 17.41          \\
\textbf{Llava1.5-7B }           & \cellcolor[HTML]{E7E6E6}12.16      & 22.85        & 3.67         & 8.49        & 4.07          & \cellcolor[HTML]{E7E6E6}12.45      & 9.31         & 18.25           & 10.26              & 39.58                                & 24.97                    & 36.37                                   & 21.78          \\
\textbf{Llava1.6-mistral-7B}    & \cellcolor[HTML]{E7E6E6}20.05      & 27.59        & 10.87        & 18.75       & 17.82         & \cellcolor[HTML]{E7E6E6}21.54      & 17.78        & 22.18           & 26.60              & 23.06                                & 30.91                    & 31.06                                   & 23.96          \\
\textbf{Qwen-VL-Chat-7B }       & \cellcolor[HTML]{E7E6E6}16.03      & 17.90        & 8.48         & 17.95       & 19.96         & \cellcolor[HTML]{E7E6E6}12.27      & 11.16        & 17.45           & 7.68               & 35.91                                & 32.91                    & 34.62                                   & 23.74          \\
\textbf{deepseek-janus-pro-7B}  & \cellcolor[HTML]{E7E6E6}28.79      & 25.57        & 27.79        & 33.46       & 24.80         & \cellcolor[HTML]{E7E6E6}32.27      & 28.19        & 31.93           & 38.98              & 36.79                                & 41.81                    & 39.66                                   & 34.30          \\
\textbf{MiniCPM-V2.6-8B} & \cellcolor[HTML]{E7E6E6}37.61      & 39.26        & 34.47        & 42.21       & 17.80         & \cellcolor[HTML]{E7E6E6}27.72      & 25.21        & 39.70           & 17.04              & 32.81                                & 37.57                    & 37.30                                   & 34.90          \\

\textbf{InternVL3‑8B}           & \cellcolor[HTML]{E7E6E6}38.40      & 41.11        & 35.81        & 40.74       & 23.18         & \cellcolor[HTML]{E7E6E6}32.94      & 29.46        & 39.10           & 30.83              & 34.40                                & 36.44                    & 34.81                                   & 35.96          \\
\textbf{mPLUG-Owl2-8.2B }       & \cellcolor[HTML]{E7E6E6}12.61      & 19.31        & 5.77         & 9.30        & 17.77         & \cellcolor[HTML]{E7E6E6}12.47      & 10.54        & 17.71           & 9.12               & 39.82                                & 25.91                    & 34.39                                   & 21.95          \\
\textbf{Pixtral-12B }           & \cellcolor[HTML]{E7E6E6}37.48      & 36.87        & 33.98        & 41.47       & 32.59         & \cellcolor[HTML]{E7E6E6}28.65      & 24.49        & 31.00           & 32.24              & 33.43                                & 43.82                    & 41.05                                   & 36.65          \\

\textbf{Llava1.6-vicuna-13B}    & \cellcolor[HTML]{E7E6E6}21.96      & 30.53        & 10.08        & 21.44       & 19.94         & \cellcolor[HTML]{E7E6E6}13.01      & 11.00        & 20.26           & 7.33               & 37.53                                & 32.31                    & 36.90                                   & 26.40          \\
\textbf{SPHINX-v2-13B}          & \cellcolor[HTML]{E7E6E6}10.92      & 14.84        & 4.02         & 11.68       & 9.15          & \cellcolor[HTML]{E7E6E6}7.55       & 8.39         & 8.69            & 4.88               & 35.04                                & 24.94                    & 36.74                                   & 19.69          \\
\textbf{Llama-4-Maverick-17B}   & \cellcolor[HTML]{E7E6E6}33.77      & 31.29        & 29.12        & 40.88       & 25.63         & \cellcolor[HTML]{E7E6E6}30.51      & 28.54        & 30.49           & 33.57              & 34.28                                & 42.89                    & 41.37                                   & 35.69          \\
\textbf{CogVLM2-19B}            & \cellcolor[HTML]{E7E6E6}30.34      & 26.26        & 22.80        & 40.25       & 23.90         & \cellcolor[HTML]{E7E6E6}28.50      & 23.20        & 34.74           & 29.13              & 38.47                                & 38.42                    & 36.84                                   & 33.53          \\
\textbf{Gemma-3-27B}            & \cellcolor[HTML]{E7E6E6}36.28      & 36.12        & 38.20        & 38.98       & 19.84         & \cellcolor[HTML]{E7E6E6}29.71      & 28.08        & 31.41           & 30.19              & 31.90                                & 40.04                    & 34.11                                   & 34.74          \\
\textbf{Llava1.6-34B}           & \cellcolor[HTML]{E7E6E6}24.74      & 31.99        & 15.05        & 24.30       & 21.31         & \cellcolor[HTML]{E7E6E6}18.34      & 15.00        & 29.43           & 10.04              & 38.59                                & 34.45                    & 38.02                                   & 29.07          \\
\textbf{Qwen2.5-VL-72B}         & \cellcolor[HTML]{E7E6E6}42.76      & 43.45        & 42.14        & 46.43       & 25.32         & \cellcolor[HTML]{E7E6E6}37.63      & 36.91        & 42.63           & 32.67              & 36.65                                & 42.22                    & 40.83                                   & 40.45          \\
\textbf{Pixtral-large-124B}     & \cellcolor[HTML]{E7E6E6}43.92      & 45.55        & 42.41        & 46.81       & 27.67         & \cellcolor[HTML]{E7E6E6}39.96      & 36.15        & 44.38           & 40.49              & 37.22                                & 47.57                    & 43.84                                   & 42.68          \\

\hline
\multicolumn{14}{c}{\textbf{Fine-tuned}} \\
\hline

\textbf{MiniCPM-V2.6-8B-fine-tuned}       & \cellcolor[HTML]{E7E6E6}53.13      & 54.87        & 54.33        & 54.05       & 36.48         & \cellcolor[HTML]{E7E6E6}38.00      & 37.46        & 43.01           & 32.77              & 35.41                                & 46.77                    & 41.54                                   & 44.89          \\

\hline
\end{tabular}
\caption{Accuracy (\%) on AstroChart benchmark using Rouge-L.}
\label{tab:Rouge-L}
\end{table*}
\clearpage
\section{M. Details of MLLMs in Evaluation}
\label{app:details_evaluation}

Table \ref{tab:model_architecture} summarizes the architecture configurations of the current mainstream open-source MLLMs used for evaluating AstroChart, including model names, Hugging Face checkpoints (HF Checkpoint), LLM branches, and visual branches.

\begin{table}[htbp]
\centering
\renewcommand{\arraystretch}{1.2} 
\resizebox{1\textwidth}{!}{
\begin{tabular}{cccc}
\hline
\textbf{Models} & \textbf{HF Checkpoint} & \textbf{LLM Branch} & \textbf{Visual Branch} \\ \hline
\textbf{TinyChart-3B} & mPLUG/TinyChart-3B-768 & Phi-2 & TinyChart-3B-768-siglip \\
\textbf{Deepseek-Janus-Pro-7B} & deepseek-ai/Janus-Pro-7B & DeepSeek-LLM-7b & SigLIP-L-384 \\
\textbf{Llava1.5-7B} & liuhaotian/llava-v1.5-7b & Vicuna-13B & CLIP ViT-L-14-336 \\
\textbf{Llava1.6-Mistral-7B} & liuhaotian/llava-v1.6-mistral-7b & Mistral-7B & CLIP ViT-L-14-336 \\
\textbf{Qwen-VL-Chat-7B} & Qwen/Qwen-VL-Chat & Qwen-7B & Openclip ViT-bigG \\
\textbf{MiniCPM-Llama3-V2.6-8B} & openbmb/MiniCPM-V-2\_6 & Qwen2-7B & SigLip-400M \\
\textbf{InternVL3‑8B} & OpenGVLab/InternVL3-8B & internlm3-8b-chat & InternViT-300M-448px \\
\textbf{mPLUG-Owl2-8.2B} & MAGAer13/mplug-owl2-llama2-7b & LLaMA-7B & ViT-L -0.3B \\
\textbf{Llava1.6-Vicuna-13B} & liuhaotian/llava-v1.6-vicuna-13b & Vicuna-13B & CLIP ViT-L-14-336 \\
\textbf{SPHINX-v2-13B} & Alpha-VLLM/LLaMA2-Accessory & LLaMA2-13B & DINOv2 VIT-g14 \& OpenCLIP ConvNeXt-XXL \\
\textbf{Llama-4-Maverick-17B} & meta-llama/Llama-4-Maverick-17B-128E-Instruct &  Llama-4 & MetaCLIP \\
\textbf{Cogvlm2-19B} & THUDM/cogvlm2-llama3-chat-19B & Meta-Llama-3-8B-Instruct & EVA2-CLIP-E \\
\textbf{Gemma-3-27B} & google/gemma-3-27b-it & Gemma3CausalLM & SigLIP-400M\\
\textbf{Llava1.6-Yi-34B} & liuhaotian/llava-v1.6-34b & Nous-Hermes-2-Yi-34B & CLIP ViT-L-14-336 \\
\textbf{Qwen2.5-VL-72B} & Qwen/Qwen2.5-VL-72B-Instruct & Qwen2.5 LLM & SwiGLU \\
\textbf{Pixtral-large-124B} & mistralai/Pixtral-Large-Instruct-2411 & Mistral-Large-Instruct-2407 & PixtralViT \\
\hline
\end{tabular}
}
\caption{Open-source MLLM architecture}
\label{tab:model_architecture}
\end{table}
\clearpage
\section{N. Failure cases of AstroChart }
\label{app:failure_cases}

\subsection{Failure cases of Visual question-answer pair}
\begin{figure}[htbp]
    \centering
    \includegraphics[width=\linewidth]{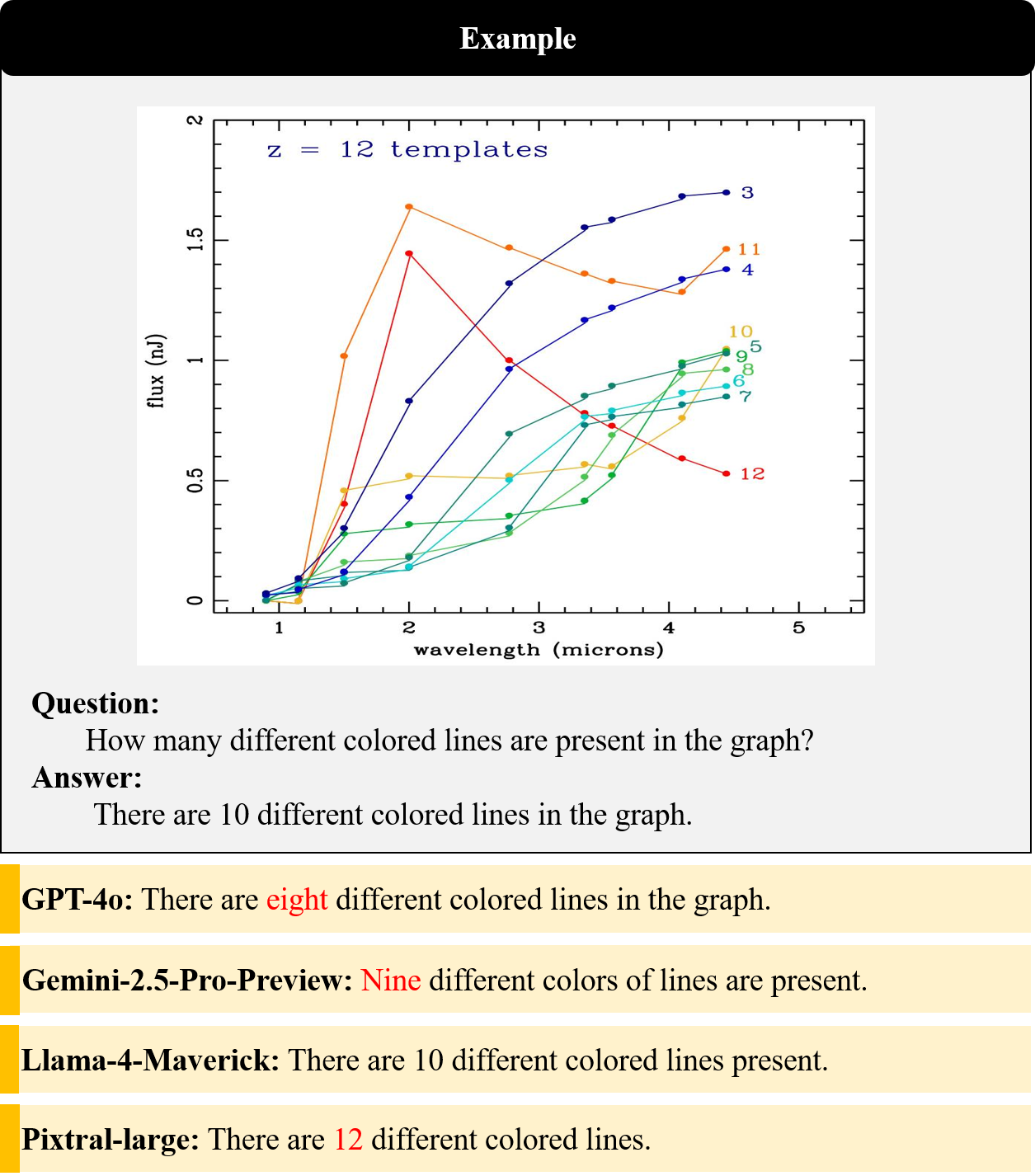}
    \caption{Failure case for visual question-answer pair generation.}
    \label{fig:visual_bad_1}
\end{figure}

\begin{figure}[htbp]
    \centering
    \includegraphics[width=\linewidth]{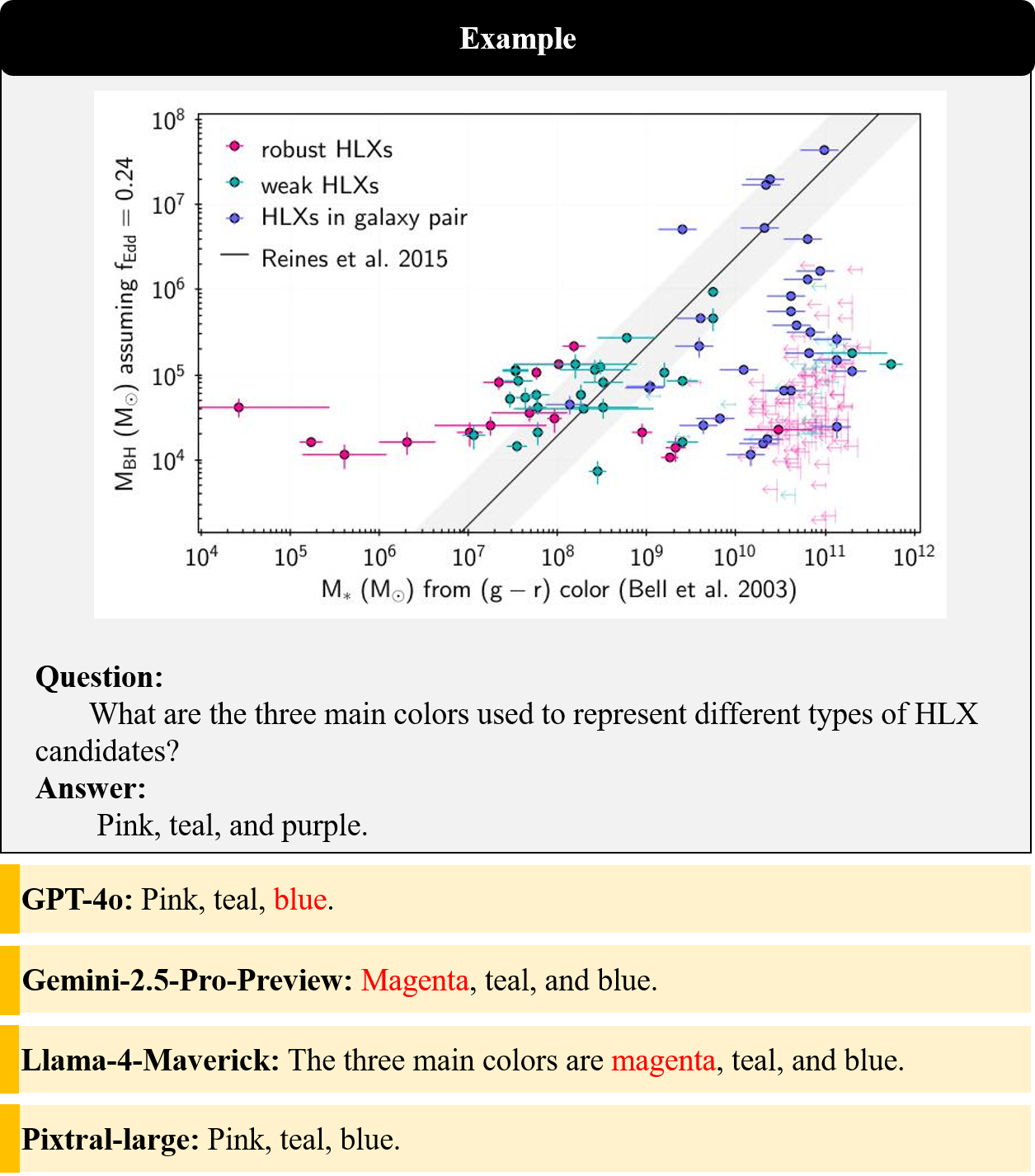}
    \caption{Failure case for visual question-answer pair generation.}
    \label{fig:visual_bad_2}
\end{figure}

\begin{figure}[htbp]
    \centering
    \includegraphics[width=\linewidth]{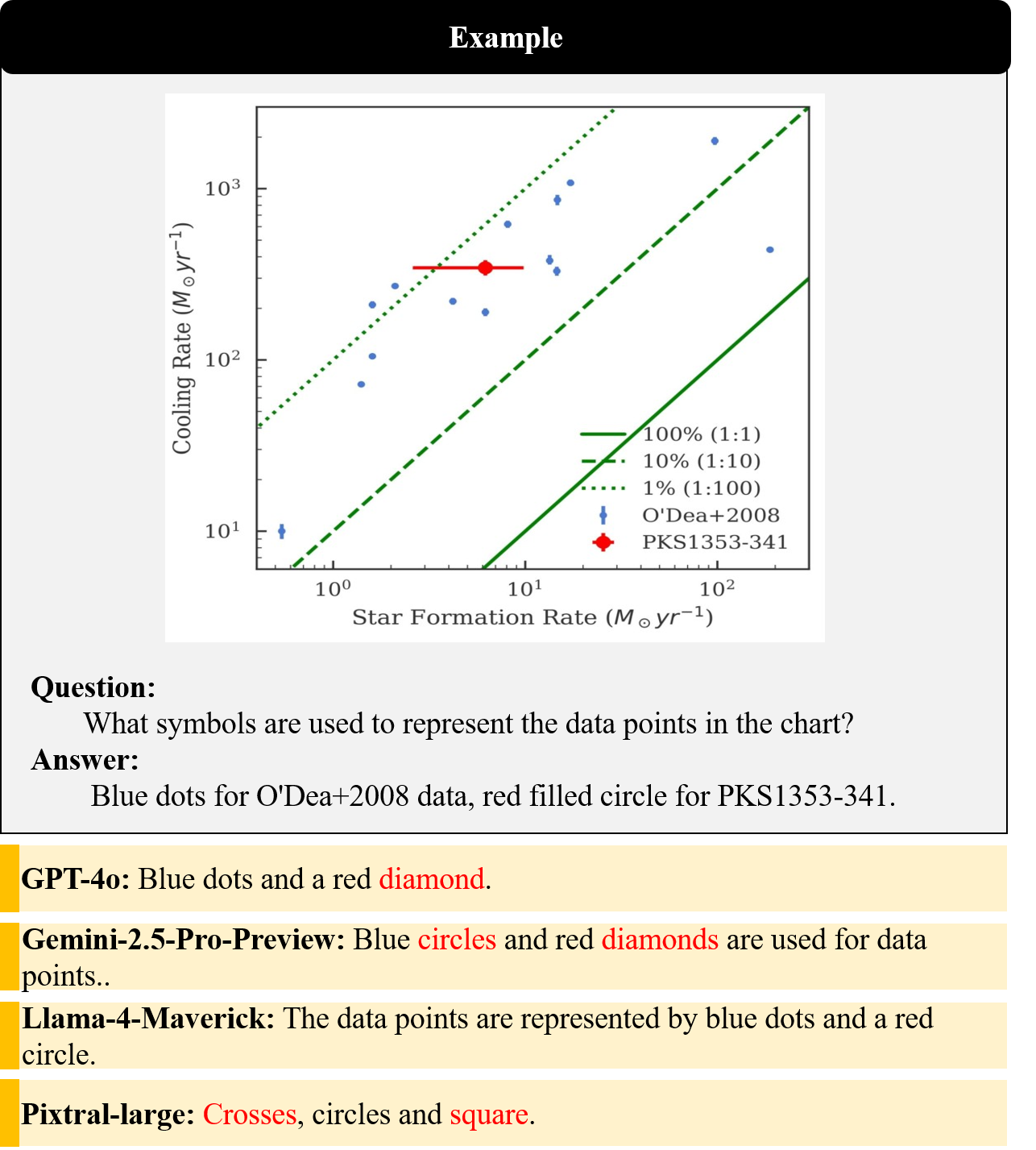}
    \caption{Failure case for visual question-answer pair generation.}
    \label{fig:visual_bad_3}
\end{figure}

Common errors in visual question-answer pair include incorrect color or pattern recognition and errors in counting or tallying (\cref{fig:visual_bad_1},\cref{fig:visual_bad_2}, \cref{fig:visual_bad_3}).

\subsection{Failure cases of Data question-answer pair}
\begin{figure}[htbp]
    \centering
    \includegraphics[width=0.9\linewidth]{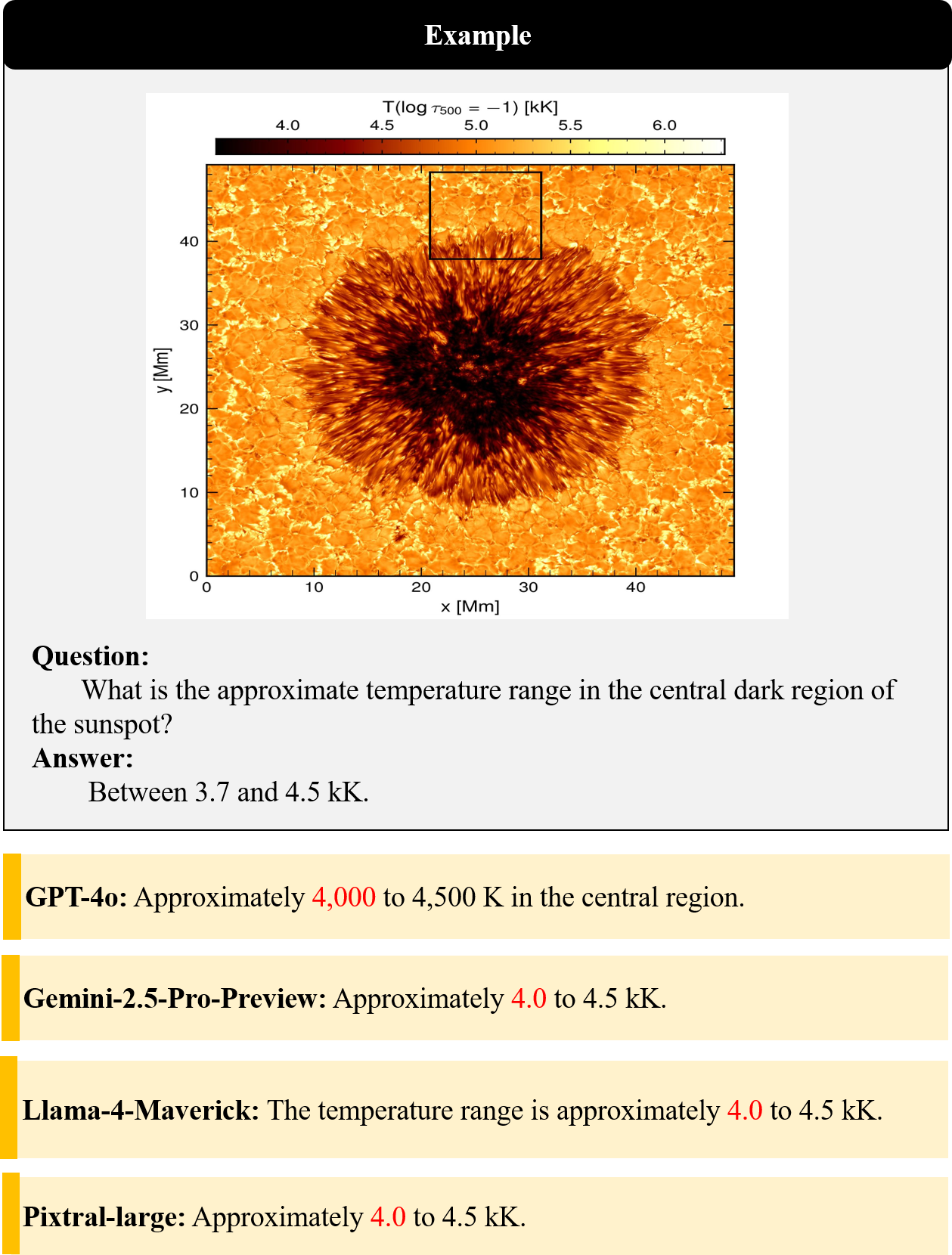}
    \caption{Failure case for data question-answer pair generation.}
    \label{fig:data_bad_1}
\end{figure}

\begin{figure}[htbp]
    \centering
    \includegraphics[width=0.9\linewidth]{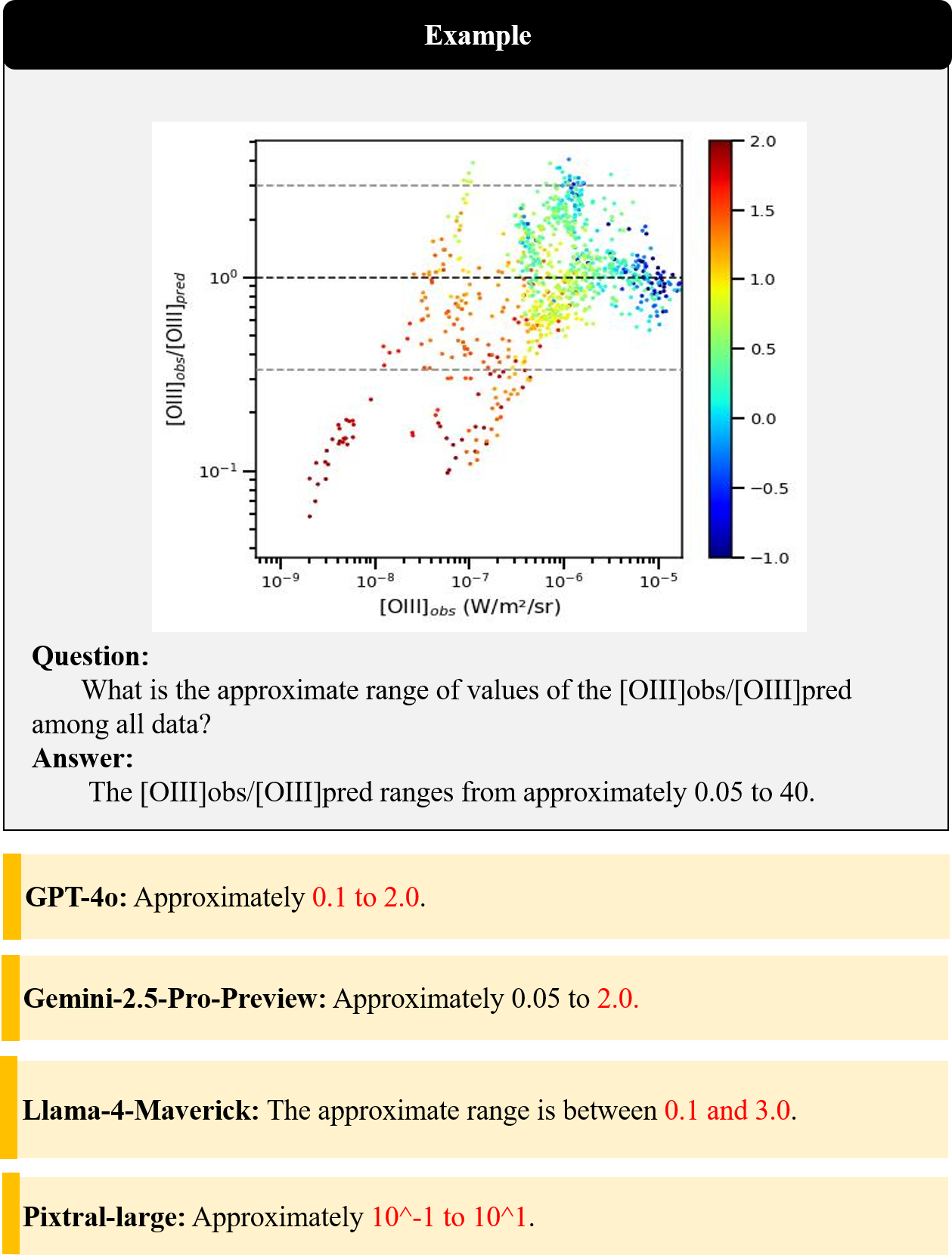}
    \caption{Failure case for data question-answer pair generation.}
    \label{fig:data_bad_2}
\end{figure}

\begin{figure}[htbp]
    \centering
    \includegraphics[width=0.9\linewidth]{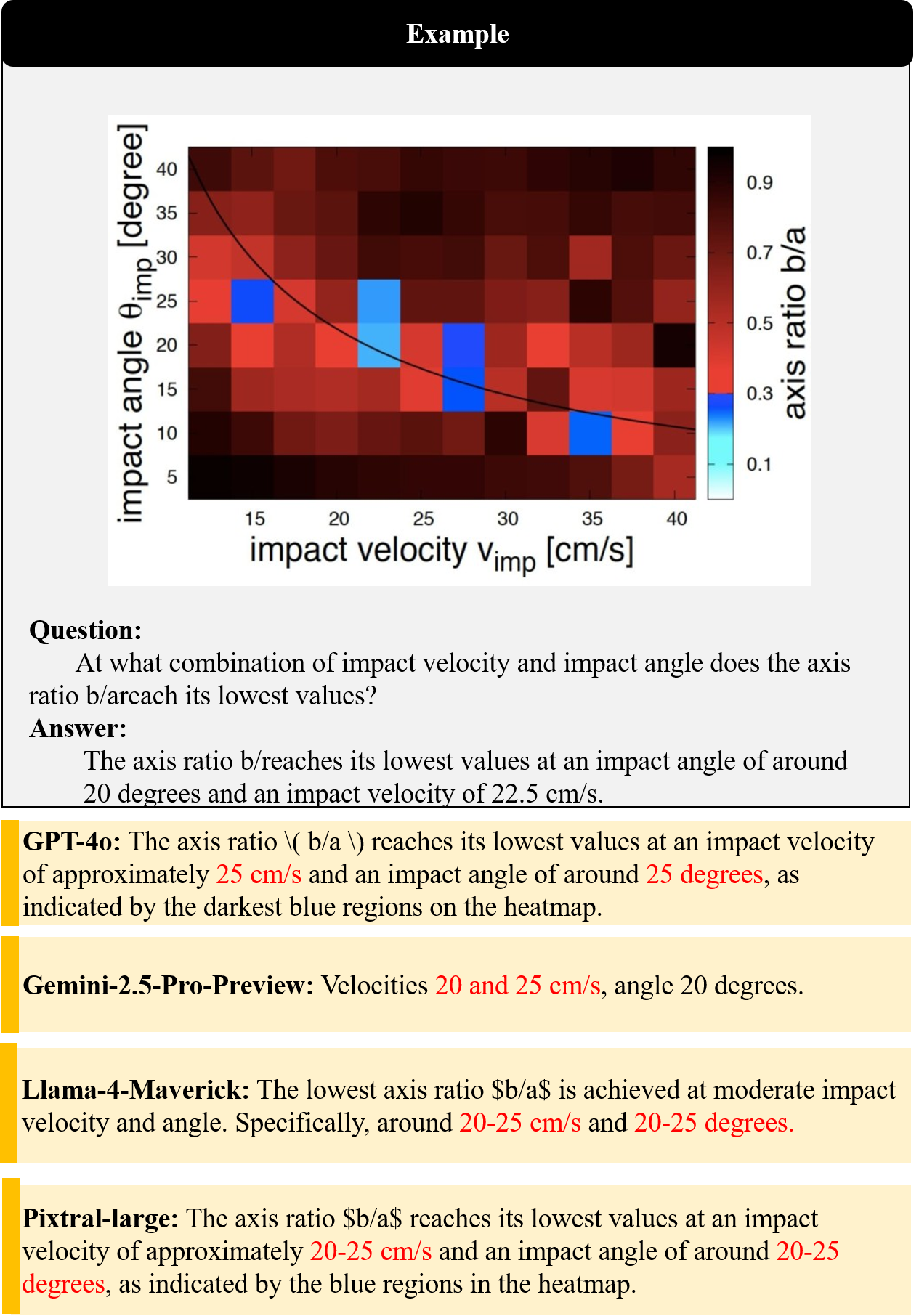}
    \caption{Failure case for data question-answer pair generation.}
    \label{fig:data_bad_3}
\end{figure}

Common errors in data question-answer pair include misreading the numbers. (\cref{fig:data_bad_1},\cref{fig:data_bad_2}, \cref{fig:data_bad_3}).

\subsection{Failure cases of Inference question-answer pair}
\begin{figure}[htbp]
    \centering
    \includegraphics[width=0.9\linewidth]{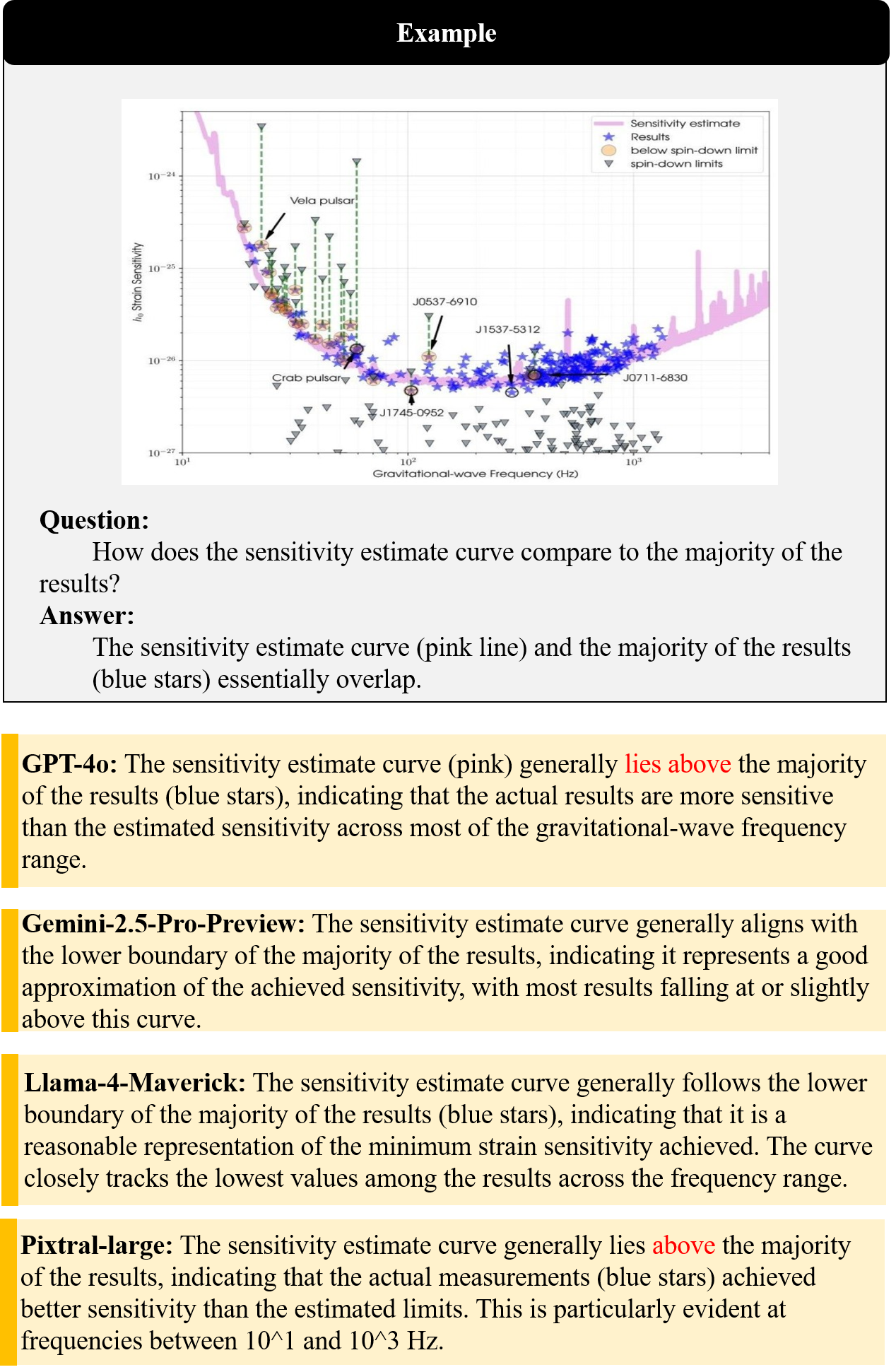}
    \caption{Failure case for inference question-answer pair generation.}
    \label{fig:inference_bad_1}
\end{figure}

\begin{figure}[htbp]
    \centering
    \includegraphics[width=0.9\linewidth]{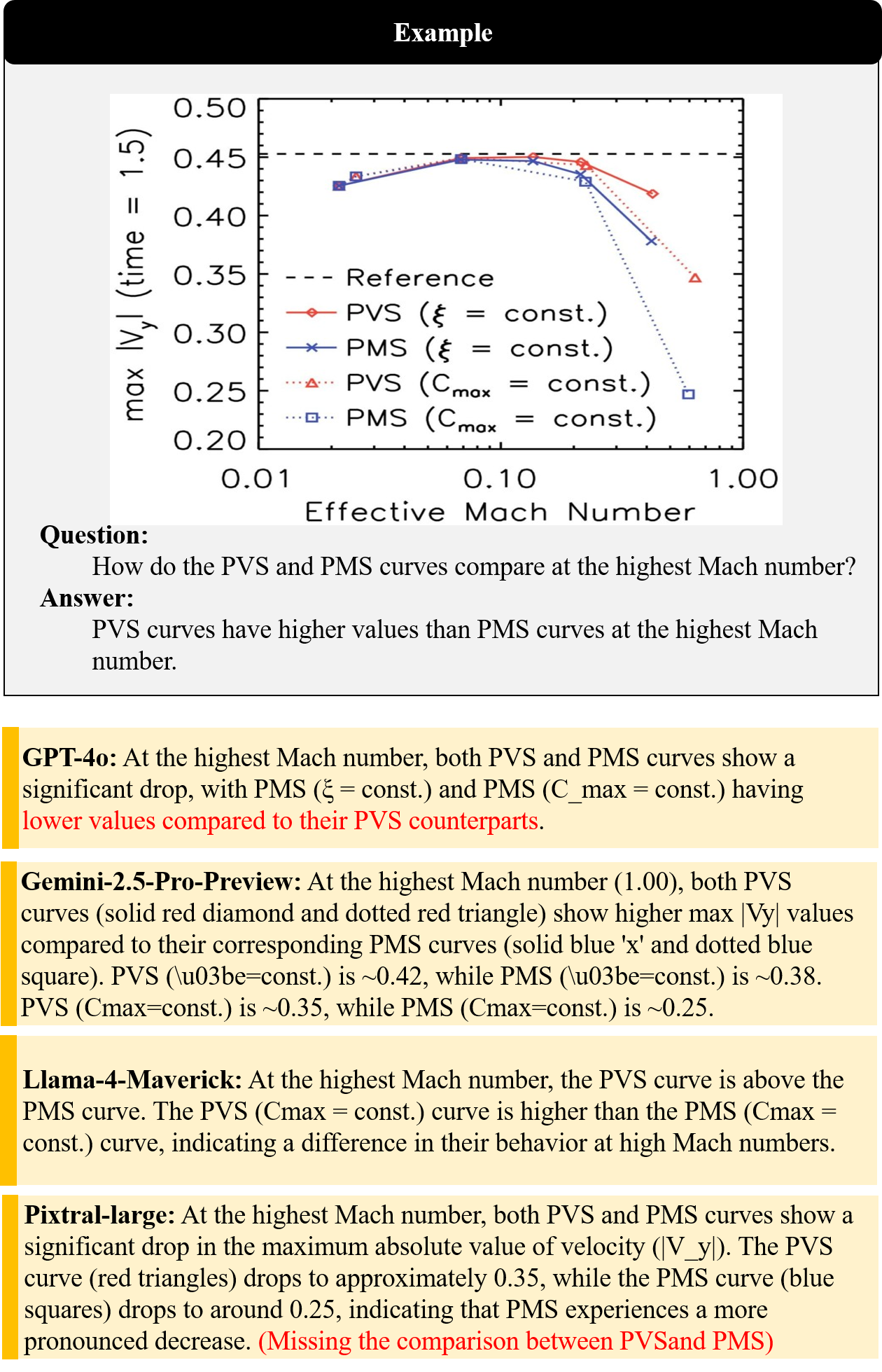}
    \caption{Failure case for inference question-answer pair generation.}
    \label{fig:inference_bad_2}
\end{figure}

\begin{figure}[htbp]
    \centering
    \includegraphics[width=0.9\linewidth]{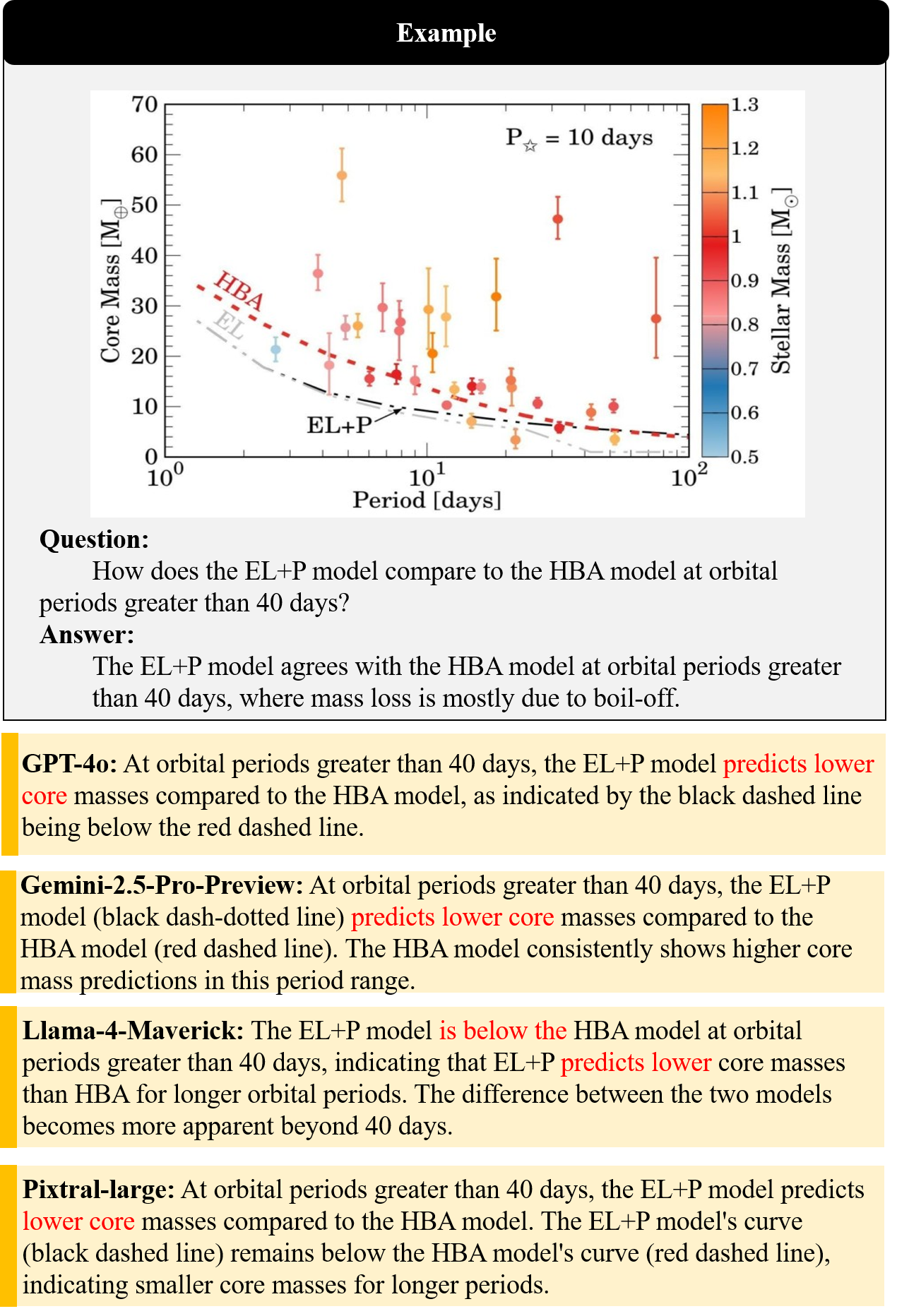}
    \caption{Failure case for inference question-answer pair generation.}
    \label{fig:inference_bad_3}
\end{figure}

Common errors in inference question-answer pairs include misinterpreting the question, providing incorrect answers, and making errors in identifying trends or comparisons. (\cref{fig:inference_bad_1},\cref{fig:inference_bad_2}, \cref{fig:inference_bad_3}).

\subsection{Failure cases of Chart Description question-answer pair}
Common errors in chart description question-answer pair include incorrect descriptions of the chart's patterns and insufficiently comprehensive descriptions of the chart. (\cref{fig:summ_bad_1},\cref{fig:summ_bad_2}, \cref{fig:summ_bad_3}).

\subsection{Failure cases of KB-Inference question-answer pair}Common errors in KB-Inference question-answer pairs include incomplete or incorrect summaries of the chart content and errors or omissions in the inferred conclusions.
(\cref{fig:kb-in_bad_1},\cref{fig:kb-in_bad_2}, \cref{fig:kb-in_bad_3}).







\begin{figure}[htbp]
    \centering
    \includegraphics[width=\linewidth]{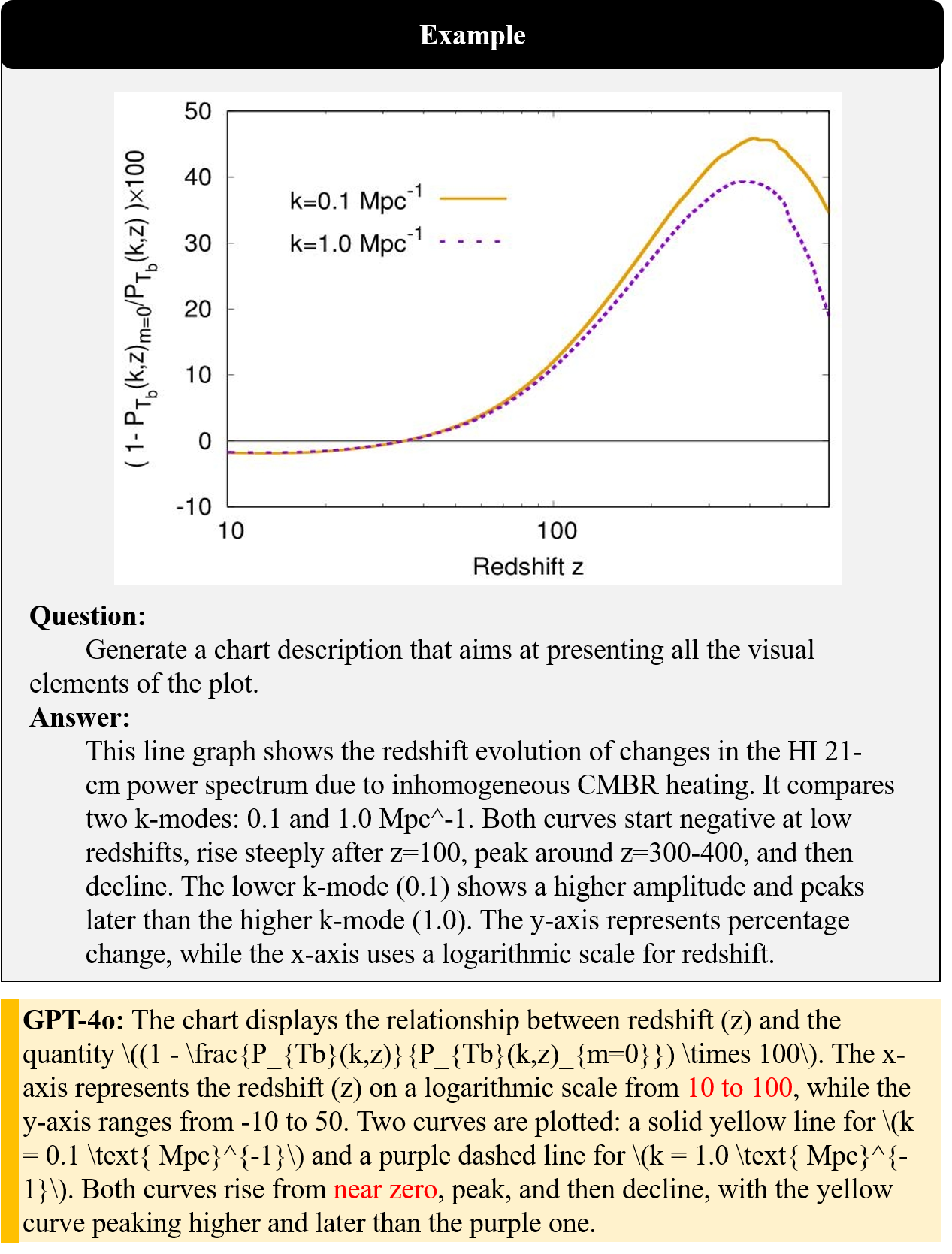}
    \caption{Failure case for chart description question-answer pair generation.}
    \label{fig:summ_bad_1}
\end{figure}

\begin{figure}[htbp]\ContinuedFloat
    \centering
    \includegraphics[width=\linewidth]{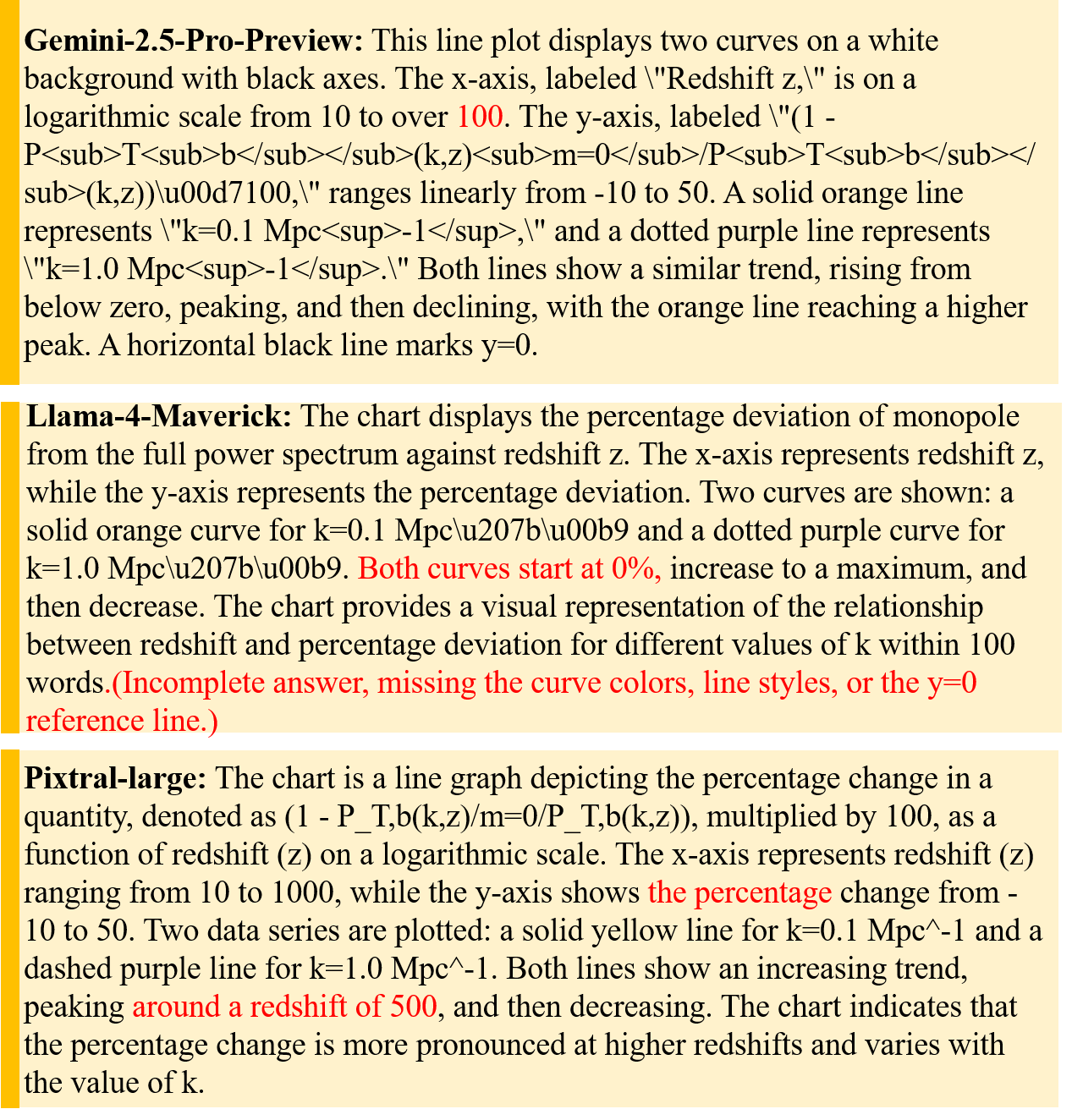}
    \caption{Failure case for chart description question-answer pair generation. (Continued)}
\end{figure}

\begin{figure}[htbp]
    \centering
    \includegraphics[width=\linewidth]{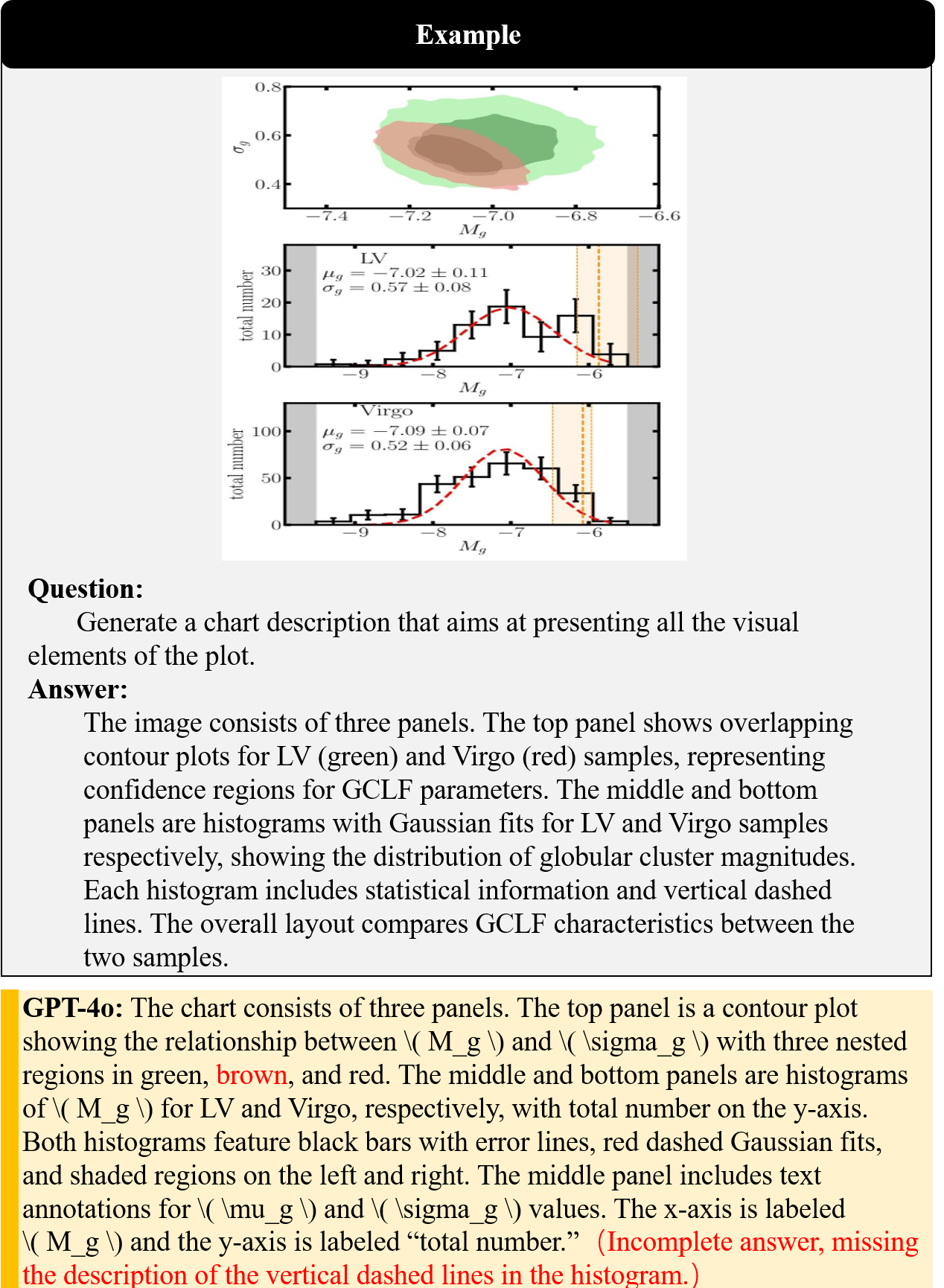}
    \caption{Failure case for chart description question-answer pair generation.}
    \label{fig:summ_bad_2}
\end{figure}
\begin{figure}[htbp]\ContinuedFloat
    \centering
    \includegraphics[width=\linewidth]{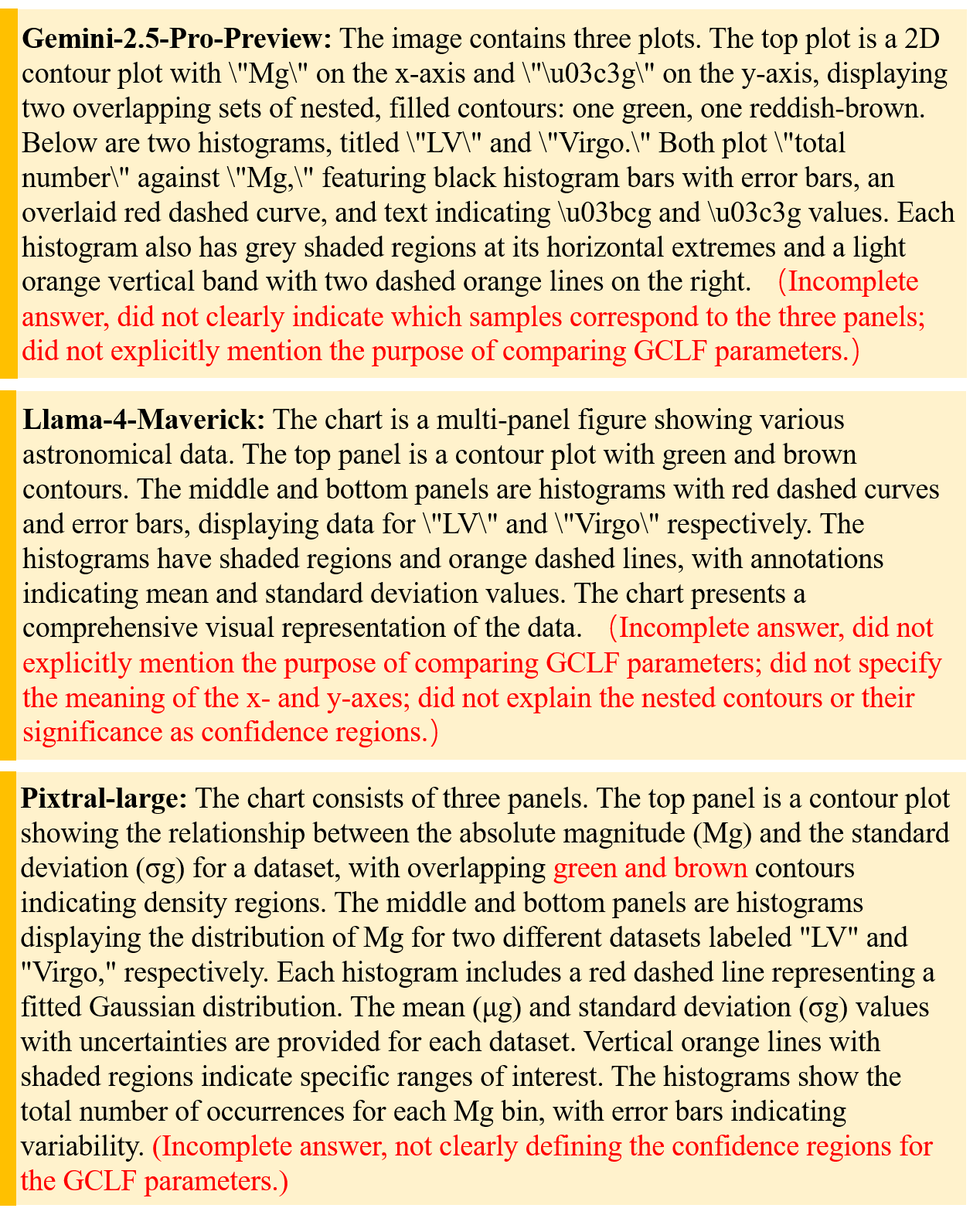}
    \caption{Failure case for chart description question-answer pair generation. (Continued)}
\end{figure}

\begin{figure}[htbp]
    \centering
    \includegraphics[width=\linewidth]{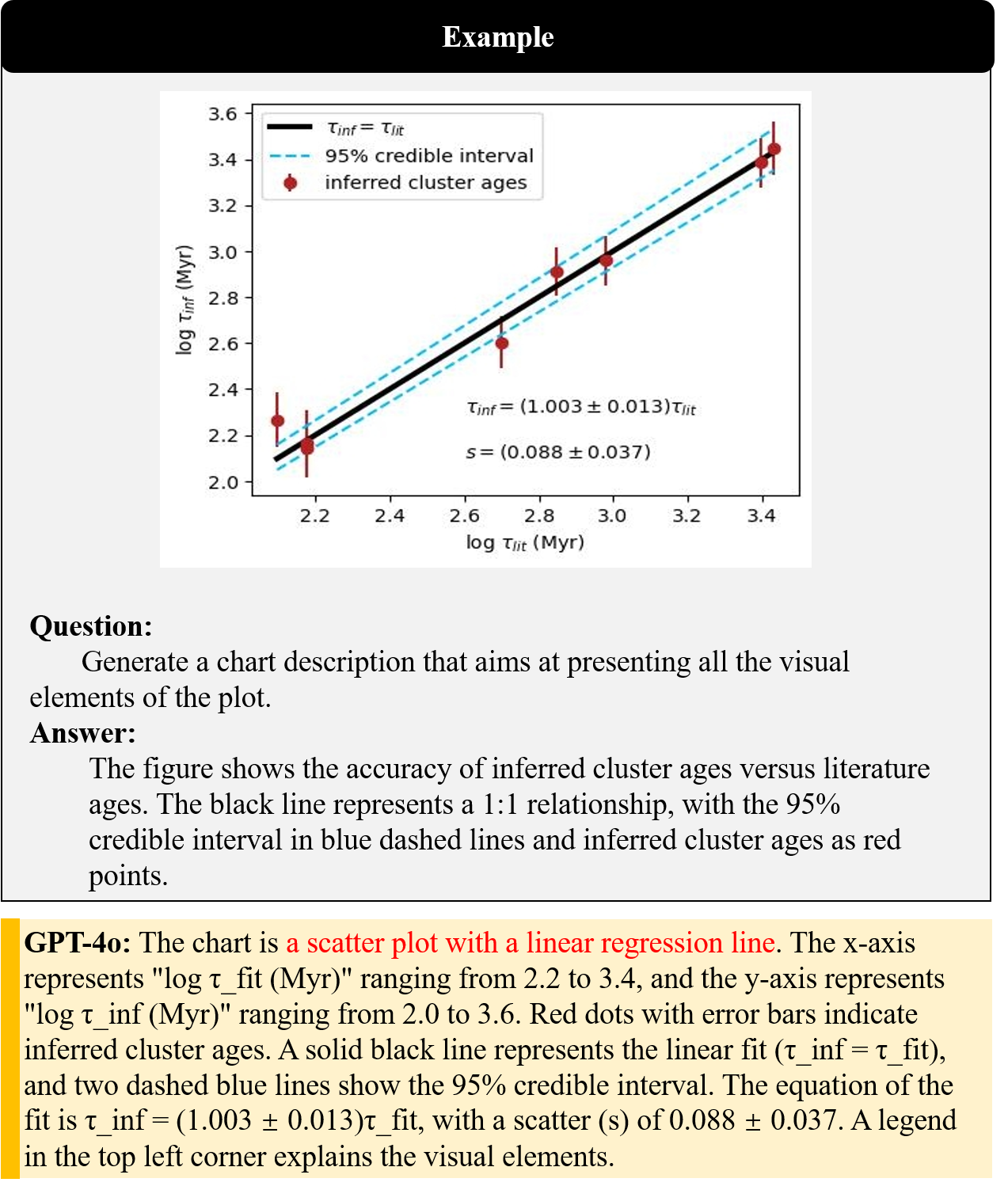}
    \caption{Failure case for chart description question-answer pair generation.}
    \label{fig:summ_bad_3}
\end{figure}
\begin{figure}[htbp]\ContinuedFloat
    \centering
    \includegraphics[width=\linewidth]{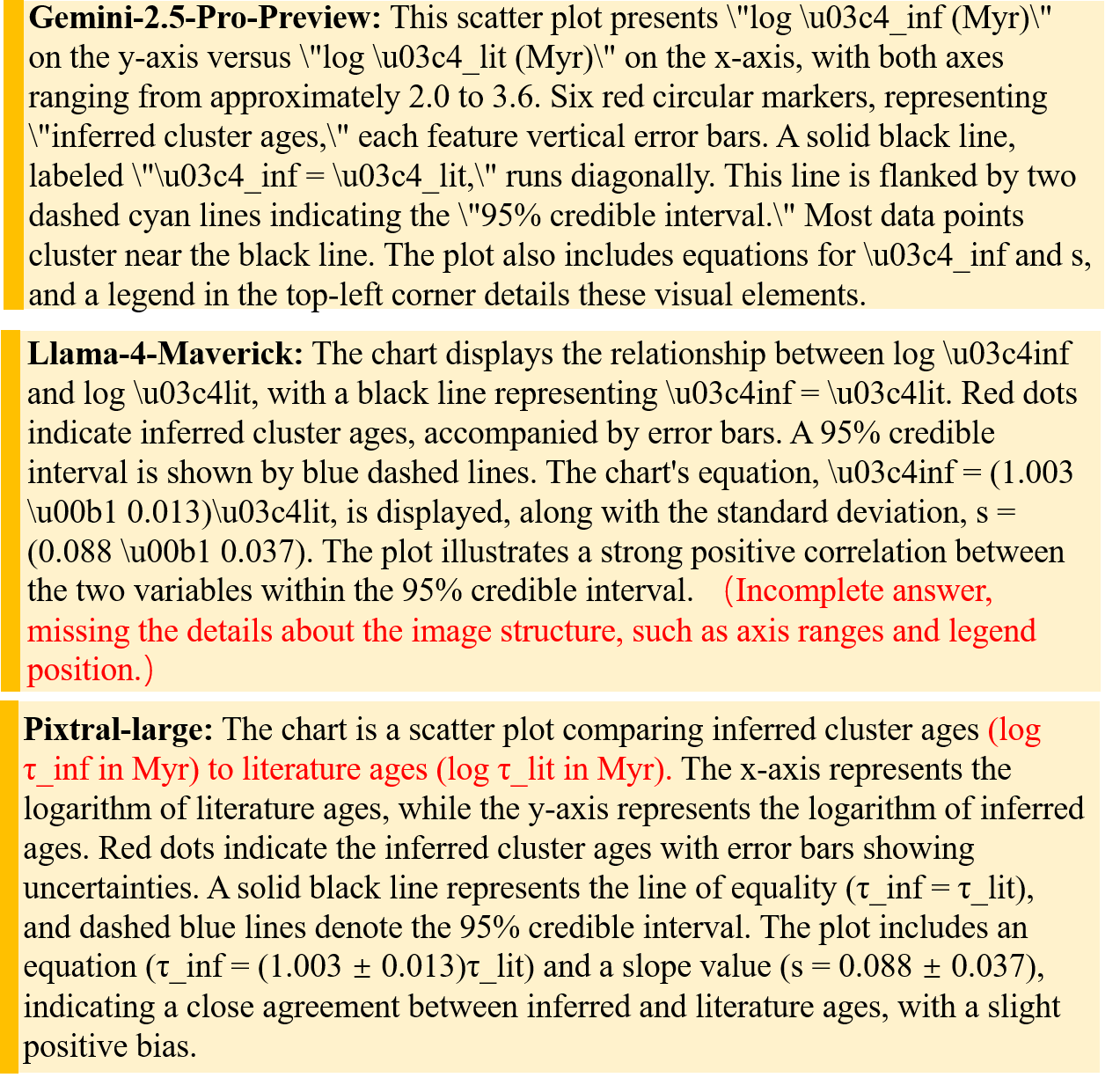}
    \caption{Failure case for chart description question-answer pair generation. (Continued)}
\end{figure}


\begin{figure}[htbp]
    \centering
    \includegraphics[width=\linewidth]{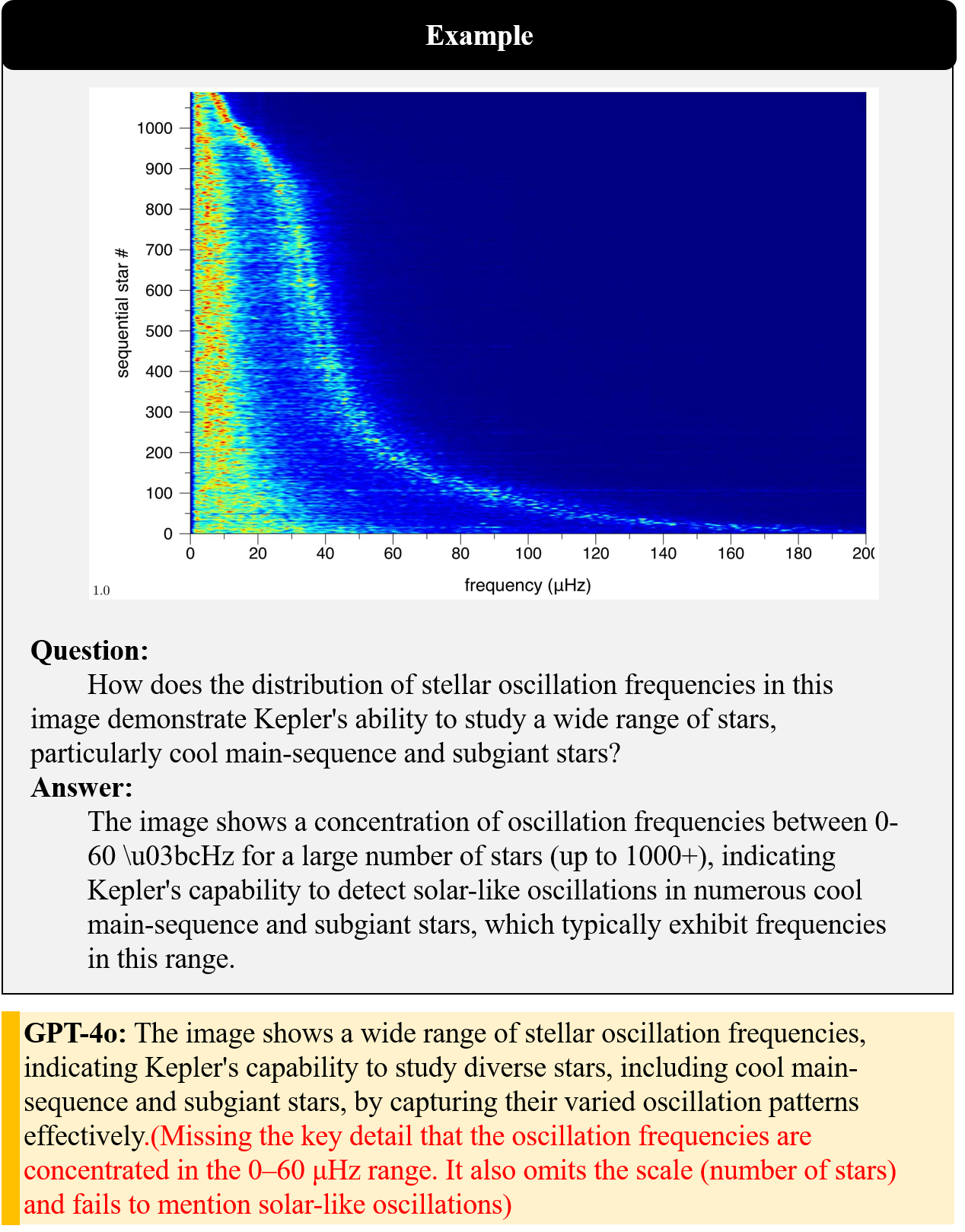}
    \caption{Failure case for KB-Inference question-answer pair generation.}
    \label{fig:kb-in_bad_1}
\end{figure}

\clearpage  

\begin{figure}[t]\ContinuedFloat  
    \centering
    \vspace*{-\baselineskip}  
    \includegraphics[width=\linewidth]{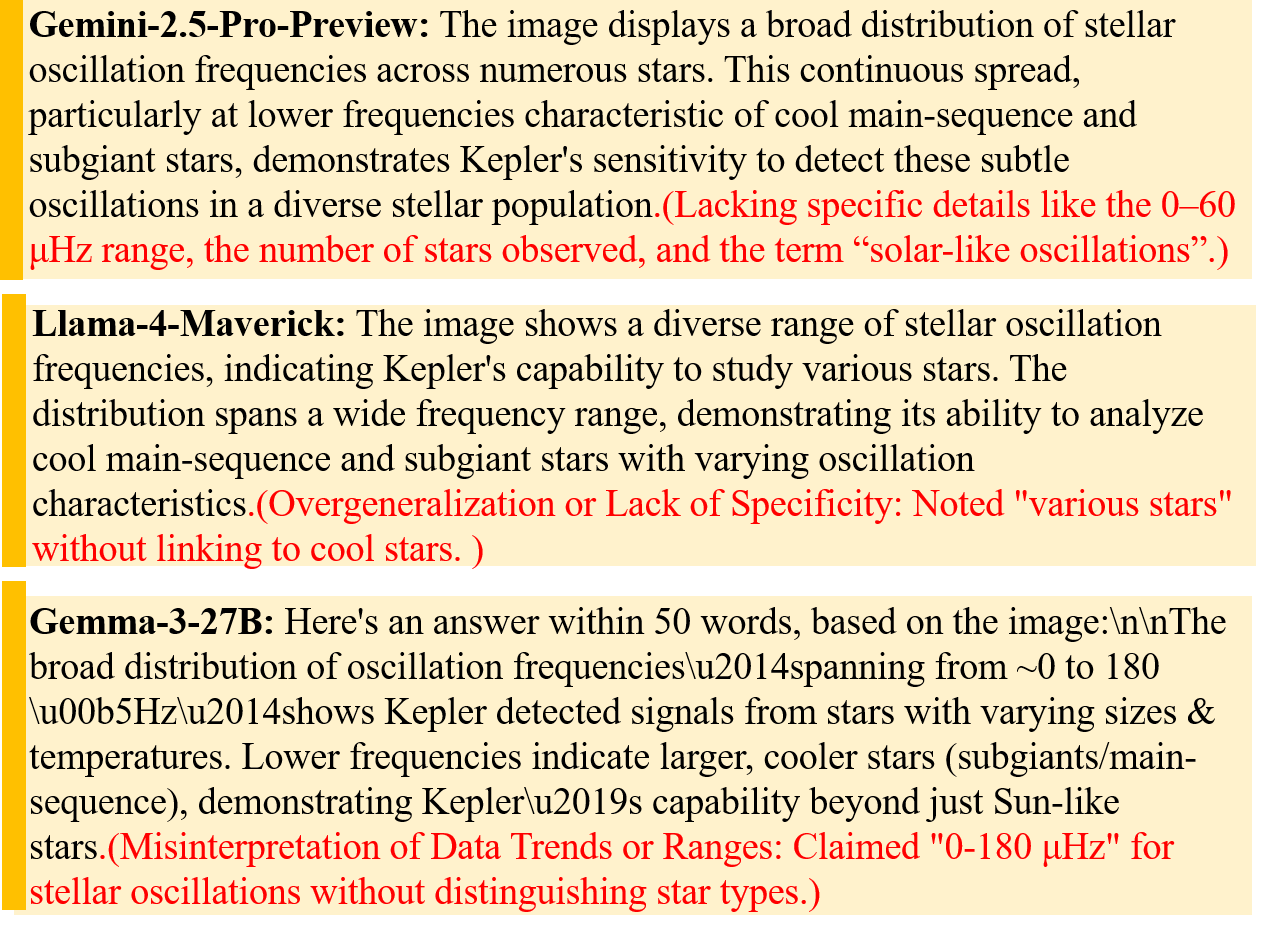}
    \caption{Failure case for KB-Inference question-answer pair generation. (Continued)}
\end{figure}

\begin{figure}[htbp]
    \centering
    \includegraphics[width=\linewidth]{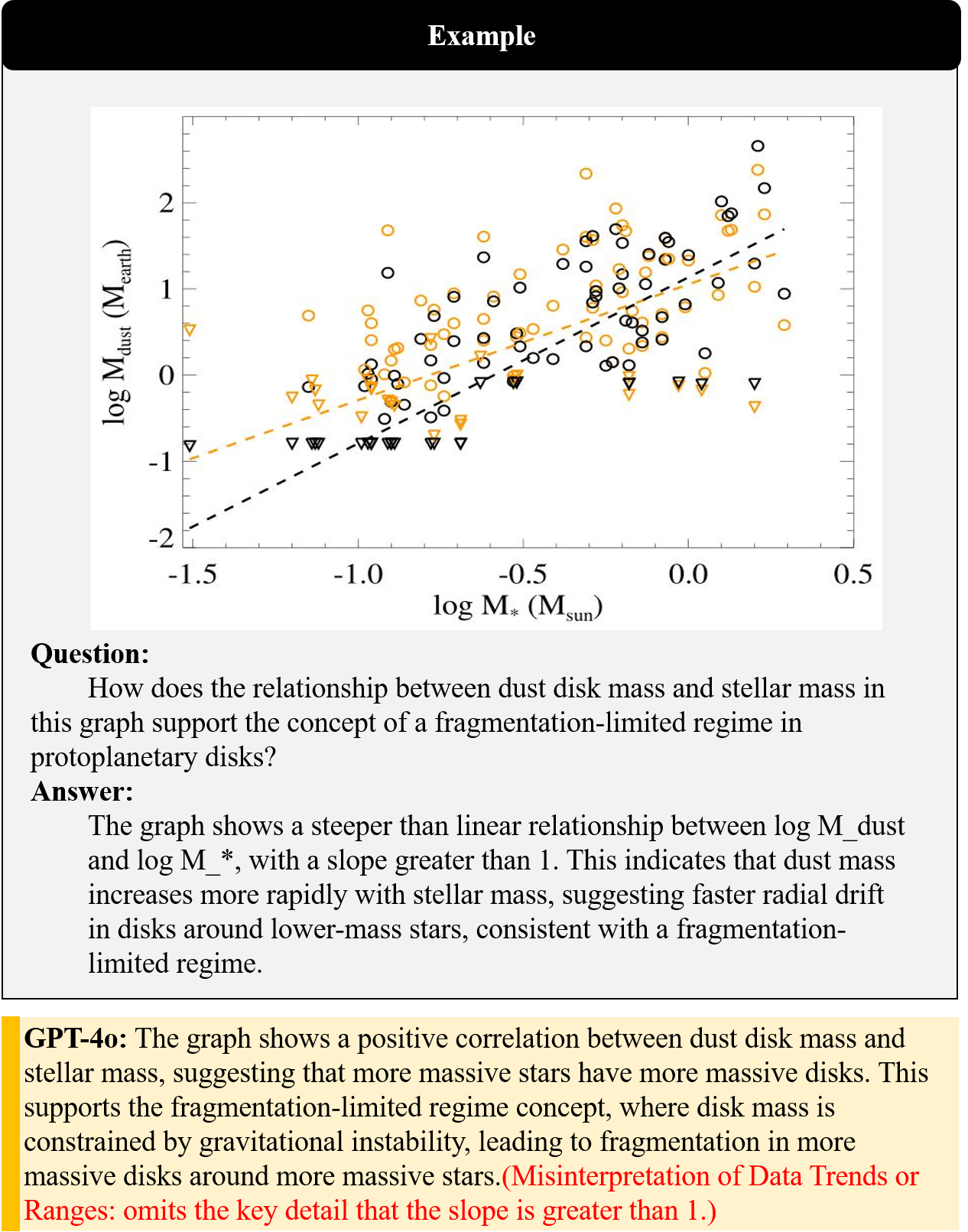}
    \caption{Failure case for KB-Inference question-answer pair generation.}
    \label{fig:kb-in_bad_2}
\end{figure}
\clearpage
\begin{figure}[htbp]\ContinuedFloat
    \centering
    \vspace*{-\baselineskip}
    \includegraphics[width=\linewidth]{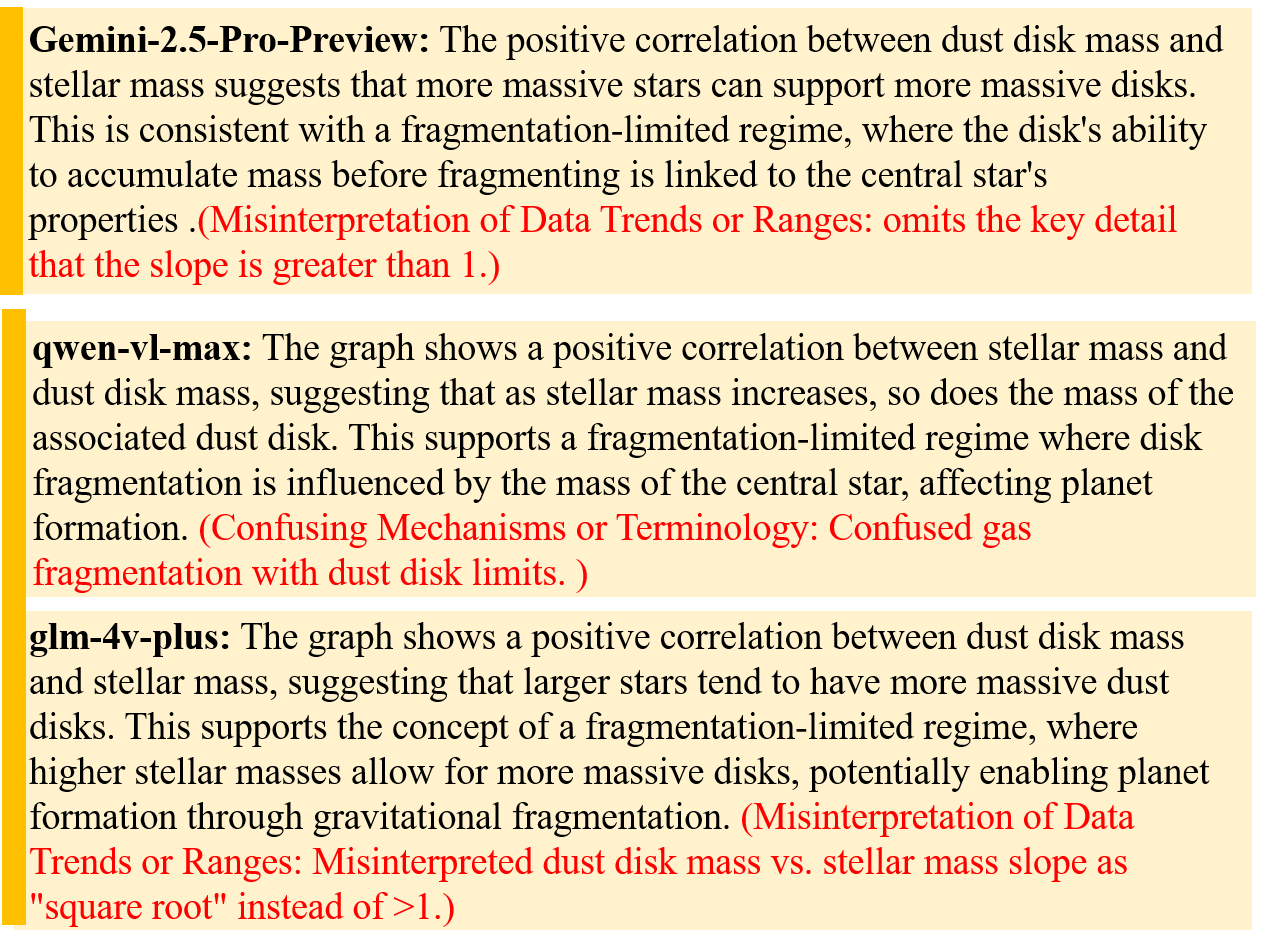}
    \caption{Failure case for KB-Inference question-answer pair generation. (Continued)}

\end{figure}

\begin{figure}[htbp]
    \centering
    
    \includegraphics[width=\linewidth]{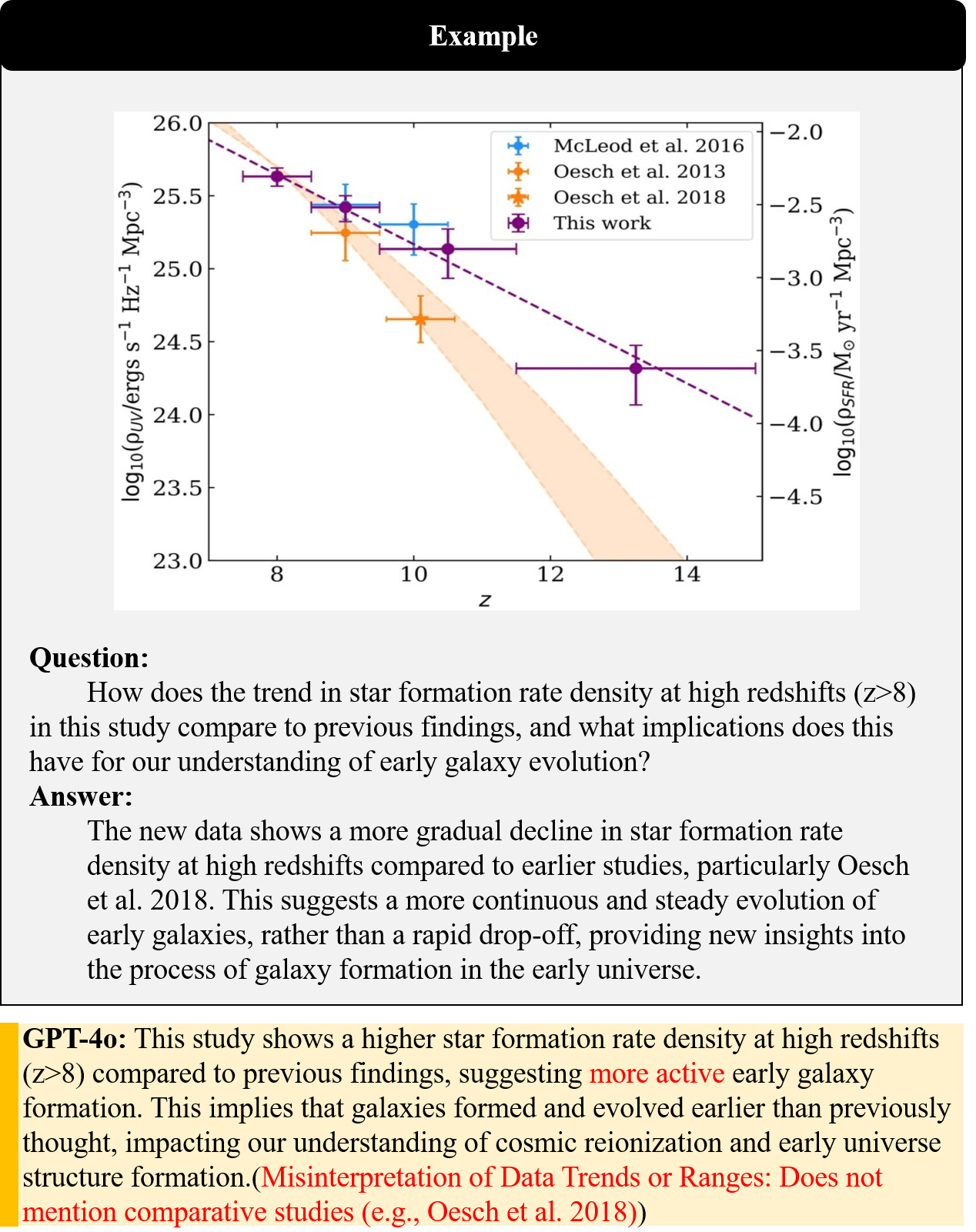}
    \caption{Failure case for KB-Inference question-answer pair generation.}
    \label{fig:kb-in_bad_3}
\end{figure}
\clearpage
\begin{figure}[htbp]\ContinuedFloat
    \centering
    \vspace*{-\baselineskip}
    \includegraphics[width=\linewidth]{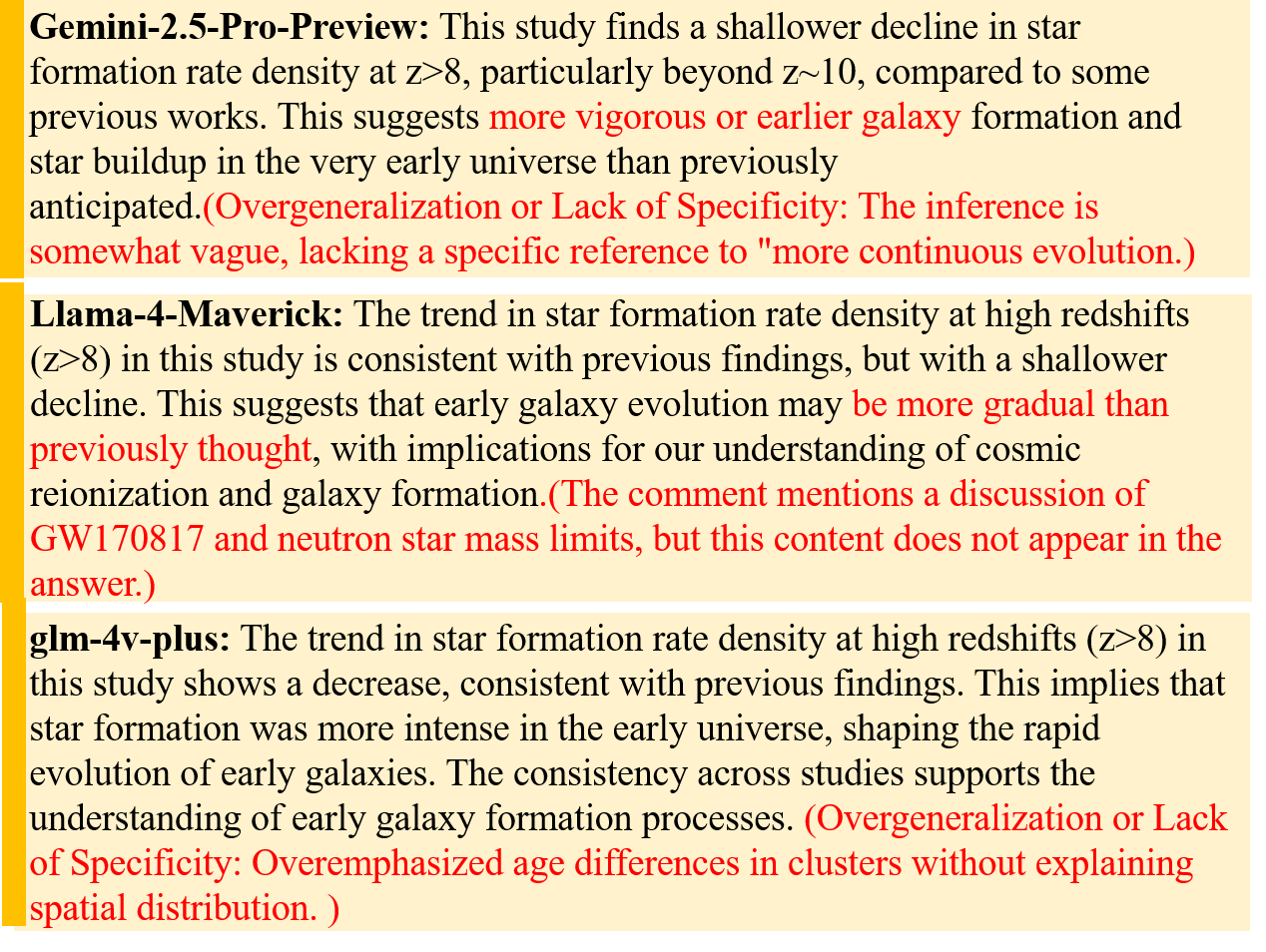}
    \caption{Failure case for KB-Inference question-answer pair generation. (Continued)}

\end{figure}


\clearpage
\section{O. Expert Scoring Guidelines}
\label{app:scoring_guidelines}

Experts are provided with a chart and a corresponding QA (Question-Answer) pair, along with the domain label (e.g., Astronomy, Biochemistry). Their task is to evaluate the QA pair along two dimensions:

\subsection{Domain Relevance}

Assess whether the QA pair meaningfully incorporates domain-specific knowledge beyond what is explicitly presented in the chart.

\begin{itemize}
    \item The QA should not merely restate chart labels, numbers, or trends. It should demonstrate a reasonable application of domain knowledge—such as scientific principles, expert terminology, or technical context—that complements and extends the chart's information.
    \item The use of domain knowledge must be appropriate and relevant. Introducing unrelated or incorrect domain knowledge should result in a lower score.
\end{itemize}

\noindent \textbf{Scoring Rubric (1–5):}
\begin{itemize}
    \item \textbf{1} – No domain knowledge, or domain knowledge is clearly incorrect.
    \item \textbf{2} – Slight or superficial domain knowledge; weakly relevant or only marginally extends the chart.
    \item \textbf{3} – Moderately appropriate domain knowledge with some depth; basic insights or moderate integration with chart content.
    \item \textbf{4} – Deep and precise domain knowledge, tightly connected to chart content; reflects strong understanding and expert-level reasoning.
    \item \textbf{5} – Domain knowledge is overly advanced or unnecessarily complex, reducing clarity or interpretability.
\end{itemize}

\subsection{QA Correctness}

Assess whether the question is clearly stated and whether the answer is factually correct based on the chart and relevant domain knowledge.

\begin{itemize}
    \item The question should be unambiguous, well-formed, and directly related to the chart.
    \item The answer should be logically and factually grounded in the chart content, possibly incorporating appropriate domain knowledge.
    \item Penalize hallucinated answers, vague questions, or any factual inconsistencies with the visualized data.
\end{itemize}

\end{document}